\PassOptionsToPackage{table}{xcolor}
\documentclass[a4paper,fleqn,screen]{cas-dc}

\usepackage[authoryear]{natbib}
\usepackage{caption}
\usepackage{subcaption}
\usepackage{booktabs}
\usepackage{multirow}
\usepackage{makecell}
\def\tsc#1{\csdef{#1}{\textsc{\lowercase{#1}}\xspace}}
\tsc{WGM}
\tsc{QE}

\begin{document}
\let\WriteBookmarks\relax

\shorttitle{CoRE-UIR: Prior-guided common and residual experts for efficient all-in-one remote sensing image restoration}

\shortauthors{Zhang et~al.}

\title[mode=title]{CoRE-UIR: Prior-guided common and residual experts for efficient all-in-one remote sensing image restoration}

\author[1]{Zaiyan Zhang}[
  style=chinese,
  auid=000,
  bioid=0,
  prefix=,
  orcid=0009-0001-7376-7101]
\ead{zzaiyan@whu.edu.cn}
\credit{Writing - Original draft, Methodology, Software, Data curation, Visualization}

\author[1]{Qiangqiang Yuan}[
  style=chinese,
  auid=000,
  bioid=0,
  prefix=,
  orcid=0000-0001-7140-2224]
\ead{qqyuan@sgg.whu.edu.cn}
\cormark[1]
\credit{Conceptualization, Supervision, Funding acquisition, Writing - Review \& Editing}

\author[1]{Jie Li}[
  style=chinese,
  auid=000,
  bioid=0,
  prefix=,
  orcid=0000-0002-4063-9381]
\ead{jli89@sgg.whu.edu.cn}
\credit{Conceptualization, Supervision, Funding acquisition, Writing - Review \& Editing}

\author[1]{Ziyang Lihe}[
  style=chinese,
  auid=000,
  bioid=0,
  prefix=,
  orcid=0009-0006-6912-648X]
\ead{Ziyang_Lihe@whu.edu.cn}
\credit{Investigation, Software, Validation}

\author[1]{Yu Wan}[
  style=chinese,
  auid=000,
  bioid=0,
  prefix=,
  orcid=0009-0000-8930-2336]
\ead{2024282140091@whu.edu.cn}
\credit{Investigation, Software, Validation}

\author[1]{Yuzeng Chen}[
  style=chinese,
  auid=000,
  bioid=0,
  prefix=,
  orcid=0000-0001-9533-8917]
\ead{yuzeng_chen@whu.edu.cn}
\credit{Investigation, Software, Validation}

\author[2]{Xin Su}[
  style=chinese,
  auid=000,
  bioid=0,
  prefix=,
  orcid=0000-0003-0957-4628]
\ead{xinsu.rs@whu.edu.cn}
\credit{Investigation, Software, Validation}

\author[3]{Liangpei Zhang}[
  style=chinese,
  auid=000,
  bioid=0,
  prefix=,
  orcid=0000-0001-6890-3650]
\ead{zlp62@whu.edu.cn}
\credit{Conceptualization, Supervision, Funding acquisition, Writing - Review \& Editing}

\affiliation[1]{organization={School of Geodesy and Geomatics, Wuhan University},
  city={Wuhan},
  citysep={}, %
  postcode={430079},
  state={Hubei},
  country={China}}

\affiliation[2]{organization={School of Artificial Intelligence, Wuhan University},
  city={Wuhan},
  citysep={}, %
  postcode={430072},
  state={Hubei},
  country={China}}

\affiliation[3]{organization={State Key Laboratory of Information Engineering in Surveying, Mapping and Remote Sensing, Wuhan University},
  city={Wuhan},
  citysep={}, %
  postcode={430079},
  state={Hubei},
  country={China}}

\cortext[1]{Corresponding author}

\begin{abstract}
  Remote sensing images acquired by unmanned aerial vehicles (UAVs) and satellites are often degraded by adverse weather, illumination variation, and imaging artifacts, which may co-occur and jointly induce global distribution shifts and local structural corruption. Although All-in-One image restoration offers an appealing unified alternative to task-specific pipelines, existing methods still suffer from weak or implicit degradation cues and parameter redundancy caused by full-rank multi-expert designs with overlapping restoration behaviors. We propose CoRE-UIR (Common and Residual Experts for Universal Image Restoration), a prior-guided global-local framework centered on the Common-and-Residual Expert Block (CoRE). CoRE explicitly decomposes restoration capacity into a common dense expert for degradation-invariant restoration and low-rank residual experts for degradation-specific compensation, enabling adaptive specialization without redundant expert replication. Built on this design, Degradation Prior Embedding (DPE) adapts frozen CLIP features into an explicit restoration-oriented prior, while Global Feature Modulation (GFM) aligns global feature statistics before local residual compensation. We also construct MDVD-108K (Multi-Degradation VisDrone), a large-scale UAV restoration dataset covering both single and compound degradations, together with a real-world test set. Extensive experiments on multiple datasets show that CoRE-UIR improves the overall average PSNR by 1.05~dB while running 11.83$\times$ faster and reducing peak memory by 85.3\% relative to the strongest baseline, BaryIR, thereby maintaining a favorable quality-efficiency trade-off. Evaluations on downstream tasks and unseen degradation also validate the generalizability of CoRE-UIR. The code and dataset will be released at \url{https://github.com/zzaiyan/CoRE-UIR}.
\end{abstract}

%
%
%


\maketitle

\section{Introduction} \label{sec:introduction}

Remote sensing imagery acquired from unmanned aerial vehicles (UAVs) and satellites plays a pivotal role in applications such as disaster response \citep{wang2025disasterm3}, environmental monitoring, urban planning, and intelligent target recognition \citep{liu2024crossmatch,chen2025hyperspectral,ZHANG2026650,wang2026region}. In practice, however, these images are frequently degraded by adverse weather, illumination variation, and imaging artifacts during acquisition \citep{shen2015missing}. Such degradations are diverse, spanning fog, dust, rain, low-light, and blur{. They also often appear in compound forms}, jointly reducing visibility, shifting appearance statistics, corrupting local structures, and undermining downstream perception and interpretation tasks \citep{yuan2020deep}.

Task-specific restoration models remain the dominant solution for individual degradations, such as dehazing \citep{song2023vision,li2019benchmarking}, deraining \citep{zamir2021mprnet}, and low-light enhancement \citep{cai2023retinexformer}. While effective for single tasks, maintaining one model per degradation is poorly suited to remote sensing pipelines that must process heterogeneous or overlapping degradations within a unified workflow. These limitations motivate All-in-One image restoration (AiOIR), which seeks a single restoration model for multiple degradation types \citep{jiang2025survey}.

Despite recent progress, existing AiOIR methods still face three coupled limitations in remote sensing scenarios. First, degradation cues are often weak, implicit, or insufficiently coupled to the restoration backbone, making it difficult to organize degradation-aware processing in a stable manner. Even when pre-trained vision-language models such as CLIP \citep{radford2021clip} are introduced, their representations are often used only as auxiliary hints rather than restoration-oriented priors. Since CLIP is trained on billion-scale vision-language data, parameter-efficient adaptation provides a practical way to preserve its strong generalization while translating semantic features into restoration-oriented degradation representations. Second, many adaptive restorers expand capacity through Mixture-of-Experts (MoE) \citep{shazeer2017moe} or multi-branch designs that replicate full-rank experts, leading to parameter redundancy and repeated learning of similar restoration behaviors. Third, remote sensing degradations often mix global statistical shifts with local structural corruption. Fog and dust may alter scene-level color, contrast, and visibility, yet their intensity can still vary with depth and layout, while rain introduces sparse local streaks. Likewise, most degradations require shared restoration abilities such as contrast enhancement, texture refinement, and brightness normalization, but some also need specialized operations such as rain-streak removal, motion compensation, or low-light denoising.

These observations motivate two complementary decompositions for universal restoration. We decouple degradation adaptation into global modulation and local compensation: the former handles scene-wide shifts in visibility, contrast, and color statistics, whereas the latter addresses spatially varying degradation patterns and structural corruption. We further decouple restoration capacity into a common dense expert and specialized residual experts: the common expert captures restoration behavior shared across degradations, while the residual experts model degradation-specific corrections without replicating full-rank branches.

Based on this view, we propose \textbf{CoRE-UIR} (\textit{Common and Residual Experts for Universal Image Restoration}), a prior-guided global-local framework for efficient AiOIR. CoRE-UIR first uses Degradation Prior Embedding (DPE) to adapt frozen CLIP features from global and local image views into a restoration-oriented degradation prior. Conditioned on this prior, Global Feature Modulation (GFM) performs prior-state global modulation at each stage entrance to organize intermediate features into a compact, prior-consistent, and more routable space. We further introduce the \textbf{Common-and-Residual Expert Block (CoRE)}, which couples the original dense backbone block, treated as a common dense expert, with low-rank residual experts for degradation-specific local compensation. The residual experts are sparsely selected by a lightweight Top-$k$ router, allowing multiple residual experts to be activated for compound degradations. These components are instantiated in a U-shaped global-local restoration backbone, with CoRE serving as the primary structural innovation.

To evaluate these design choices under remote sensing settings, we construct MDVD-108K (\textit{Multi-Degradation VisDrone Dataset}), a large-scale UAV multi-degradation dataset built on VisDrone \citep{zhu2021detection}, containing six single degradations, six compound degradations, and real-world degraded UAV images for qualitative evaluation. We further perform satellite-domain validation on MDRS-Landsat.

Our main contributions are summarized as follows:
\begin{itemize}
  \item A prior-guided global-local restoration framework, CoRE-UIR, is introduced to adapt frozen CLIP features into restoration-oriented degradation priors and inject them into a unified backbone through global modulation and local expert compensation.

  \item We propose the Common-and-Residual Expert Block (CoRE), which decomposes restoration capacity into a common dense expert and low-rank residual experts, reducing the redundancy of full-rank MoE branches while retaining degradation-adaptive compensation.

  \item We construct MDVD-108K, a large-scale UAV restoration benchmark comprising 108K paired synthetic samples with object detection annotations, covering six single degradations, six compound degradations, and real-world degraded UAV images.

  \item Extensive experiments on UAV and satellite benchmarks show that CoRE-UIR consistently improves restoration quality and achieves a stronger quality-efficiency trade-off than strong universal restoration baselines across single and compound degradations.
\end{itemize}

The remaining sections of the paper are organized as follows. Section~\ref{sec:related_work} presents related work. Section~\ref{sec:methodology} describes the proposed method. Section~\ref{sec:experiments} reports the experiments and analysis. Finally, Section~\ref{sec:conclusion} concludes the paper.

\section{Related Work} \label{sec:related_work}

\subsection{Single-Task Image Restoration}

Single-task image restoration is still dominated by task-specific or fixed-degradation models. In natural images, representative progress spans early CNN-based restoration and super-resolution methods such as SRCNN \citep{dong2015image} and RCAN \citep{zhang2018image}, generic transformer restorers such as IPT \citep{chen2021pre}, SwinIR \citep{liang2021swinir}, CAT \citep{chen2022cross}, and HAT \citep{chen2023activating}, and degradation-specific models for dehazing \citep{song2023vision,liu2026ihdcp}, low-light enhancement \citep{cai2023retinexformer}, dust removal \citep{wei2025robust}, together with generic backbones such as MPRNet \citep{zamir2021mprnet}, Restormer \citep{zamir2022restormer}, DGUNet \citep{mou2022dgunet}, NAFNet \citep{chen2022nafnet}, and MambaIR \citep{guo2024mambair}. More recent progress also includes newer transformer variants, task-specific weather restoration models, and deblurring benchmarks \citep{jin2025mb,zhang2022enhanced,zhang2021deep,wen2026structure,zhang2023mc}. These methods are effective when the degradation type is known and fixed, but they do not address the unified setting in which one model must adapt to heterogeneous degradations.

The same paradigm persists in remote sensing restoration. Earlier studies often relied on handcrafted low-rank and total-variation priors for hyperspectral denoising and image completion \citep{Cheng2014Patch,he2015total}. More recent deep models focus on task-specific cloud removal, missing-data reconstruction, and multimodal restoration, including multitemporal sequence modeling \citep{stucker2023u,zhang2025multi,shu2025restore}, transformer or diffusion-based cloud removal \citep{li2020thin,li2025cloudruler,jing2023denoising,sui2024diffusion}, remote-sensing dehazing \citep{wen2023encoder,wen2025cross}, panshapening \citep{cui2025pansharpening,cui2025enpowering} and SAR-optical or multimodal fusion strategies \citep{LuojiaSET,MTGAN,zhang2026ecrformer,zhang2026task,chen2026any2any}. Although these methods leverage temporal, sensor, or generative priors, they are still specialized to isolated restoration families such as cloud removal or missing-data recovery. As a result, they provide limited guidance for remote sensing scenarios where multiple degradations may co-exist within a unified deployment pipeline.

\subsection{All-in-One Image Restoration}

AiOIR relaxes the fixed-degradation assumption by training a single model for multiple degradations. Early representative methods include AirNet \citep{li2022airnet}, TransWeather \citep{valanarasu2022transweather}, WeatherDiff \citep{ozdenizci2023weatherdiff}, ProRes \citep{ma2023prores}, PromptIR \citep{potlapalli2024promptir}, and IDR \citep{zhang2023idr}, which explore degradation encoding, weather-oriented modeling, diffusion-based reconstruction, prompt conditioning, and ingredient-oriented learning. Later methods strengthen universal restoration through improved prompts, stronger degradation representations, or modular adaptation, such as sequential/prompt learning \citep{kong2024towards}, EvoIR \citep{ma2025evoir}, DACLIP-UIR \citep{luo2023daclip}, and BaryIR \citep{tang2026learning}. Related extensions further explore weather-oriented \citep{gao2023frequency,wen2025all} and perception-oriented formulations \citep{Perceive-IR,hu2025clusir,ClearAIR,wang2026residual}, as well as low-rank and domain-specialized variants \citep{ai2024lora,zhang2024uir,UniUIR}.

This universal-restoration paradigm has only recently been extended to remote sensing, partly enabled by multi-degradation benchmarks such as MDRS-Landsat introduced in Ada4DIR \citep{lihe2025ada4dir}. Related prompt-based universal restoration has also been explored for hyperspectral imagery \citep{wu2025mp} and multi-modal imagery \citep{cui2026unified}. Ada4DIR \citep{lihe2025ada4dir} enhances multi-level degradation extraction and adaptive degradation recognition through prompt-injection-fusion and model-driven prompt blocks, while PhyDAE \citep{dong2026phydae} introduces physics-guided degradation-adaptive experts with progressive degradation mining and sparse activation. These methods highlight the value of explicit degradation modeling and remote-sensing-specific priors, yet they still do not explicitly decompose restoration capacity into a common dense path and lightweight low-rank residual experts, nor do they fully address the redundancy of heavy expert branches under compound degradations.

\subsection{Mixture of Experts}

Mixture of Experts (MoE) offers an effective mechanism for scaling model capacity while keeping per-sample computation sparse \citep{shazeer2017moe,riquelme2021scaling}. The key idea is to maintain multiple experts and use a gating function to activate only a subset of them for each input, thereby improving specialization without evaluating the entire expert pool. This design has been widely explored in large-scale representation learning and provides a natural tool for modeling heterogeneous degradations in image restoration.

In AiOIR, MoE-style designs have been used to allocate degradation-dependent capacity through routed expert branches or adaptive expert modules. Representative examples include multi-expert adaptive selection \citep{yu2024multi}, feature-modulated experts \citep{zhang2024efficient}, CLIP-guided MoE gating in M2Restore \citep{wang2025m2restore}, complexity-aware experts in MoCE-IR \citep{zamfir2025complexity}, and degradation-adaptive experts in remote sensing restoration \citep{dong2026phydae}. These studies show that routed experts are effective for handling heterogeneous degradations and compound corruptions.
However, most existing MoE-based restorers allocate adaptation capacity through full-rank or heavily overlapping experts. In universal restoration, many operations such as visibility enhancement, denoising, and structure recovery are broadly shared across degradations, so expert replication can introduce redundancy in parameters, computation, and learned behaviors. This limitation is especially noticeable under compound degradations, where multiple routed branches may still repeat common restoration operations.

These observations motivate our common-residual expert design, where the dense backbone preserves common restoration capacity and the routed low-rank residual branch focuses on degradation-specific residual compensation. CoRE can also be viewed as an asymmetric sparse MoE, combining an always-active common expert with selectively activated low-rank residual experts to reduce parameter redundancy relative to full-rank expert replication. In this way, expert routing remains adaptive, while expert allocation becomes more redundancy-aware for universal remote sensing restoration.

\section{Methodology} \label{sec:methodology}

In this section, we first introduce the problem formulation and the overall framework, then describe the key components of the proposed CoRE-UIR, and finally present the training strategy.

\begin{figure*}[t!]
  \centering
  \includegraphics[width=\linewidth]{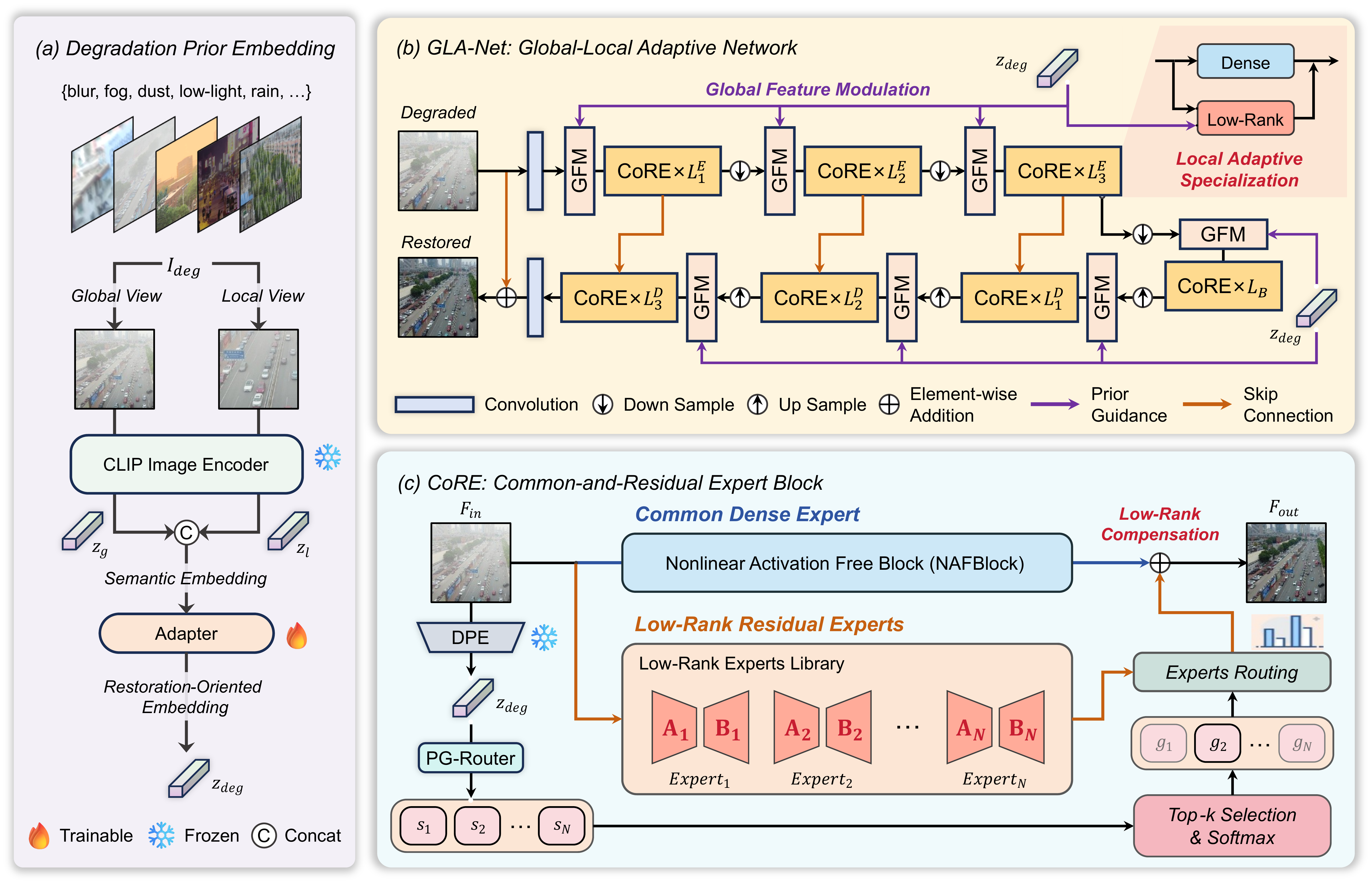}
  \caption{The overall framework of the proposed CoRE-UIR. (a) Degradation Prior Embedding (DPE): a frozen CLIP image encoder processes multi-scale views of the input, and a lightweight adapter maps the features into restoration-oriented degradation embeddings. (b) Global-Local Adaptive Network: the unified restoration backbone integrates Global Feature Modulation (GFM) and Common-and-Residual Expert Blocks (CoRE). (c) The CoRE block decomposes restoration into a common dense expert for degradation-invariant restoration and low-rank residual experts for degradation-specific compensation.}
  \label{fig:framework}
  \vspace{-1em}
\end{figure*}

\subsection{Overview}
\label{sec:overview}

\subsubsection{Problem Formulation}
\label{sec:problem_formulation}

Let $\mathcal{D} = \{d_m\}_{m=1}^{M}$ denote the set of $M$ base degradation types in a given restoration setting, and let $\mathcal{C} \subseteq \{\mathcal{S} \mid \emptyset \neq \mathcal{S} \subseteq \mathcal{D}\}$ denote the degradation configurations, where each $\mathcal{S} \in \mathcal{C}$ specifies the active degradation subset. For a clean remote sensing image $I_\text{gt} \in \mathbb{R}^{H \times W \times 3}$, the degraded observation is modeled as:
\begin{equation}
  I_\text{deg} = \mathcal{G}_{\mathcal{S}}\left(I_\text{gt}\right), \quad \mathcal{S} \in \mathcal{C},
  \label{eq:degradation_observation}
\end{equation}
where $\mathcal{G}_{\mathcal{S}}(\cdot)$ denotes the structured degradation operator induced by the active subset. Single degradation corresponds to $|\mathcal{S}| = 1$, where the observation is generated by one elemental operator, while compound degradation corresponds to $|\mathcal{S}| > 1$, where multiple operators are composed in sequence.

The goal is to learn one universal restoration model for all degradation configurations in the same setting:
\begin{equation}
  \hat{I} = f_{\theta}\left(I_\text{deg}\right), \quad \forall \, \mathcal{S} \in \mathcal{C},
\end{equation}
where $f_{\theta}$ shares one set of parameters across all base and compound degradations.

\subsubsection{Framework Overview}
\label{sec:framework_overview}

As summarized in Eqs.~\ref{eq:dpe} and~\ref{eq:restoration_framework}, CoRE-UIR consists of Degradation Prior Embedding (DPE) and the Global-Local Adaptive Network (GLA-Net). DPE converts one global view and one local view of the degraded image into a restoration-oriented prior embedding $z_\text{deg}$ using a frozen CLIP image encoder and a lightweight adapter, so that the prior space remains generalizable while becoming more suitable for restoration.
\begin{align}
  z_\text{deg} = & \text{DPE}\left(I_\text{deg}\right) \in \mathbb{R}^{d_z}, \label{eq:dpe}                                                     \\
  \hat{I} =      & \text{GLA-Net}\left(I_\text{deg}, z_\text{deg}\right) \in \mathbb{R}^{H \times W \times 3}. \label{eq:restoration_framework}
\end{align}

The Global-Local Adaptive Network (GLA-Net) restores the image conditioned on $z_\text{deg}$. It adopts a shared U-shaped backbone with two complementary modules: Global Feature Modulation (GFM), which injects stage-level global conditioning, and the Common-and-Residual Expert block (CoRE), which performs block-level local restoration. Within CoRE, the original dense block serves as the common expert, while routed low-rank residual experts provide degradation-specific compensation.

\subsection{Degradation Prior Embedding}
\label{sec:prior_extraction}

Frozen CLIP features provide a strong starting point for degradation reasoning, but they are optimized for semantic alignment rather than restoration control and therefore cannot be used directly as restoration priors. Benefiting from CLIP pretraining on billion-scale vision-language data, parameter-efficient adaptation can preserve its strong generalization while translating semantic representations into restoration-oriented degradation embeddings. We design Degradation Prior Embedding (DPE) to adapt CLIP features into a compact degradation space with minimal trainable overhead. By combining a global view and a local crop, DPE preserves both scene-level degradation context and local degradation textures before lightweight adaptation.

\subsubsection{Multi-Scale Prior Extraction}

Raw UAV aerial images typically exhibit high resolution and a spatially uniform scale, making direct full-resolution encoding computationally expensive and insensitive to both global degradation patterns (\emph{e.g.}, overall haze opacity) and local degradation details (\emph{e.g.}, fine rain streaks). We therefore construct two complementary views from $I_\text{deg}$:
\begin{equation}
  I_\text{g} = \text{Resize}\left(I_\text{deg}\right),\quad I_\text{l} = \text{Resize}\left(\text{Crop}\left(I_\text{deg}\right)\right),
  \label{eq:views}
\end{equation}
where both views are scaled to the CLIP encoder's native input resolution. The global view $I_\text{g}$ preserves overall degradation context, while the local view $I_\text{l}$ captures regional degradation textures. Their encoder outputs are fused by concatenation:
\begin{equation}
  z_\text{clip} = \text{Concat}\left(\text{CLIP}\left(I_\text{g}\right), \text{CLIP}\left(I_\text{l}\right)\right) \in \mathbb{R}^{2d_\text{clip}},
  \label{eq:ms_fusion}
\end{equation}
where $d_\text{clip}$ is the CLIP feature dimension of each individual view. This multi-scale fusion combines complementary degradation cues from global context and local textures before projection.

\subsubsection{Parameter-Efficient Adaptation}

Although the frozen CLIP encoder offers strong generalization, its pretraining objective targets high-level semantic alignment rather than low-level restoration guidance. We therefore introduce a lightweight learnable adapter to bridge this gap:
\begin{equation}
  z_\text{deg} = \text{Adapter}\left(z_\text{clip}\right) \in \mathbb{R}^{d_z},
  \label{eq:projector}
\end{equation}
where $\text{Adapter}(\cdot)$ consists of two linear layers with a GELU activation, mapping $z_\text{clip}$ to a compact restoration-oriented embedding $z_\text{deg}$ of dimension $d_z$. To make this embedding discriminative for mixed degradation configurations, we attach a linear classification head to $z_\text{deg}$ and optimize a multi-label degradation classification objective, while keeping the CLIP encoder frozen and updating only the adapter and classifier.

This parameter-efficient adaptation rapidly aligns the semantic feature space with the remote-sensing restoration domain. After convergence, the classification head is discarded and the resulting DPE is used to condition both GFM and CoRE during restoration learning.

\begin{figure}[t]
  \centering
  \includegraphics[width=\linewidth]{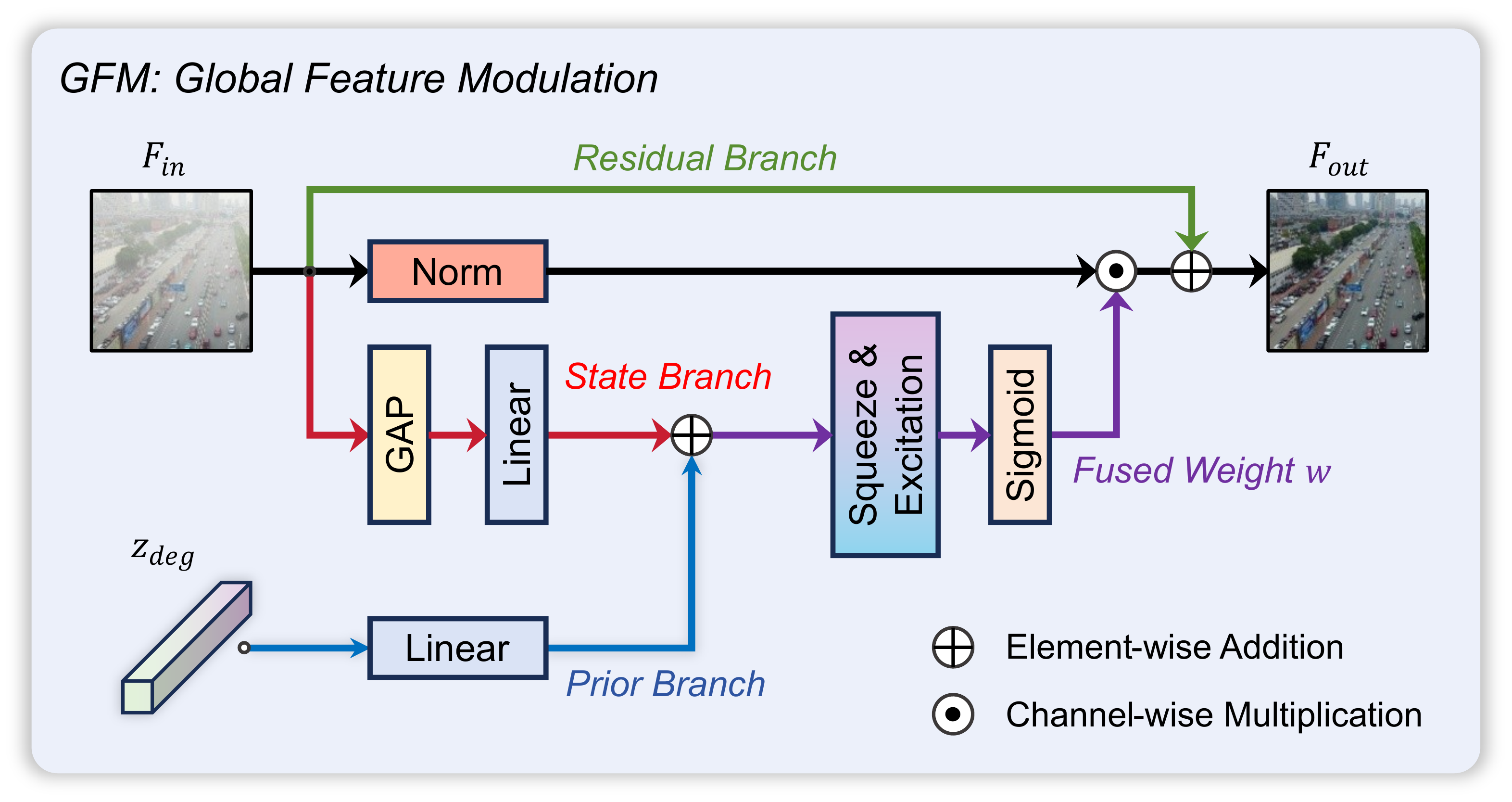}
  \caption{Architecture of the proposed \textbf{Global Feature Modulation (GFM)} module. GFM combines the degradation prior with the current feature state to perform lightweight degradation-aware channel recalibration before CoRE processing.}
  \label{fig:gfm_module}
  \vspace{-1em}
\end{figure}

\subsection{Global Feature Modulation (GFM)}
\label{sec:gfm}

An explicit degradation prior alone is insufficient for feature modulation, because the same degradation may appear with different severity and intermediate feature responses across images and network stages. GFM therefore combines the degradation prior with the current feature state to perform lightweight degradation-aware channel recalibration, as illustrated in Fig.~\ref{fig:gfm_module}.

Given a feature map $F \in \mathbb{R}^{H' \times W' \times C}$ from the backbone and the degradation prior $z_\text{deg} \in \mathbb{R}^{d_z}$, GFM extracts a state descriptor from the current feature response and a prior descriptor from the degradation embedding:
\begin{align}
  f_\text{state} = & \text{Linear}\left(\text{GAP}\left(F\right)\right) \in \mathbb{R}^{C},
  \label{eq:gfm_state}                                                                      \\
  f_\text{prior} = & \text{Linear}\left(z_\text{deg}\right) \in \mathbb{R}^{C},
  \label{eq:gfm_prior}
\end{align}
where $\text{GAP}(\cdot)$ denotes global average pooling across spatial dimensions. The two descriptors are then fused and passed through a bottleneck excitation head:
\begin{align}
  f_\text{fused} & = f_\text{state} + f_\text{prior}, \label{eq:gfm_fusion}                                                                \\
  w              & = \sigma\!\left(\text{MLP}_\text{SE}\left(f_\text{fused}\right)\right) \in \mathbb{R}^{C}, \, \label{eq:gfm_excitation}
\end{align}
where $\text{MLP}_\text{SE}(\cdot)$ denotes a two-layer bottleneck network with GELU activation and reduction ratio $\rho$, and $\sigma(\cdot)$ denotes the Sigmoid function. The resulting gate modulates the feature map as:
\begin{equation}
  \text{GFM}\left(F\right) = F + \text{LN}\left(F\right) \odot w,
  \label{eq:gfm_output}
\end{equation}
where $\text{LN}(\cdot)$ denotes LayerNorm, and $\odot$ denotes broadcast channel-wise multiplication over spatial dimensions.

GFM thus extends the squeeze-and-excitation paradigm \citep{hu2018squeeze} with an explicit degradation prior branch, providing compact global modulation before downstream CoRE processing.

\subsection{Common-and-Residual Expert Block (CoRE)}
\label{sec:core}

After GFM organizes global feature statistics, the restoration network still requires degradation-specific residual transformations to correct locally varying structural corruption. Since much of the restoration process is shared across degradations, replicating multiple full-rank experts is unnecessarily costly. We therefore introduce the Common-and-Residual Expert Block (CoRE), which is inserted in parallel with each backbone block and decomposes restoration capacity into a \textbf{common dense expert} and \textbf{low-rank residual experts}. Borrowing the low-rank bottleneck from LoRA-style adaptation, CoRE instead couples an always-active common expert with routed low-rank residual experts, yielding an asymmetric sparse-MoE form rather than fixed low-rank updates to a frozen backbone.

Specifically, the dense backbone block captures restoration knowledge shared across degradations, such as visibility enhancement, texture recovery, and structure reconstruction, while the routed low-rank branch focuses on degradation-specific residual compensation. The low-rank constraint limits the capacity of each residual expert, discouraging repeated learning of common operations and instead driving the experts to model degradation-dependent differences. This common-residual decomposition improves parameter efficiency while preserving adaptive capacity.

\subsubsection{Low-Rank Residual Expert Library}
Each CoRE block maintains a library of $N$ low-rank expert pairs $\{(A_n, B_n)\}_{n=1}^{N}$, where $A_n \in \mathbb{R}^{r \times C}$ and $B_n \in \mathbb{R}^{C \times r}$ are learnable projection matrices with bottleneck rank $r \ll C$, applying to the channel dimension at each spatial location. Each pair, therefore, forms a compact low-rank residual expert. In implementation, $A_n$ and $B_n$ are implemented by two successive $3\times3$ convolutions that map $C\rightarrow r$ and $r\rightarrow C$, respectively. This preserves a low-rank bottleneck along the channel dimension while enlarging the local receptive field.

\subsubsection{Prototype-Guided Router (PG-Router)}

\begin{figure}[t]
  \centering
  \includegraphics[width=\linewidth]{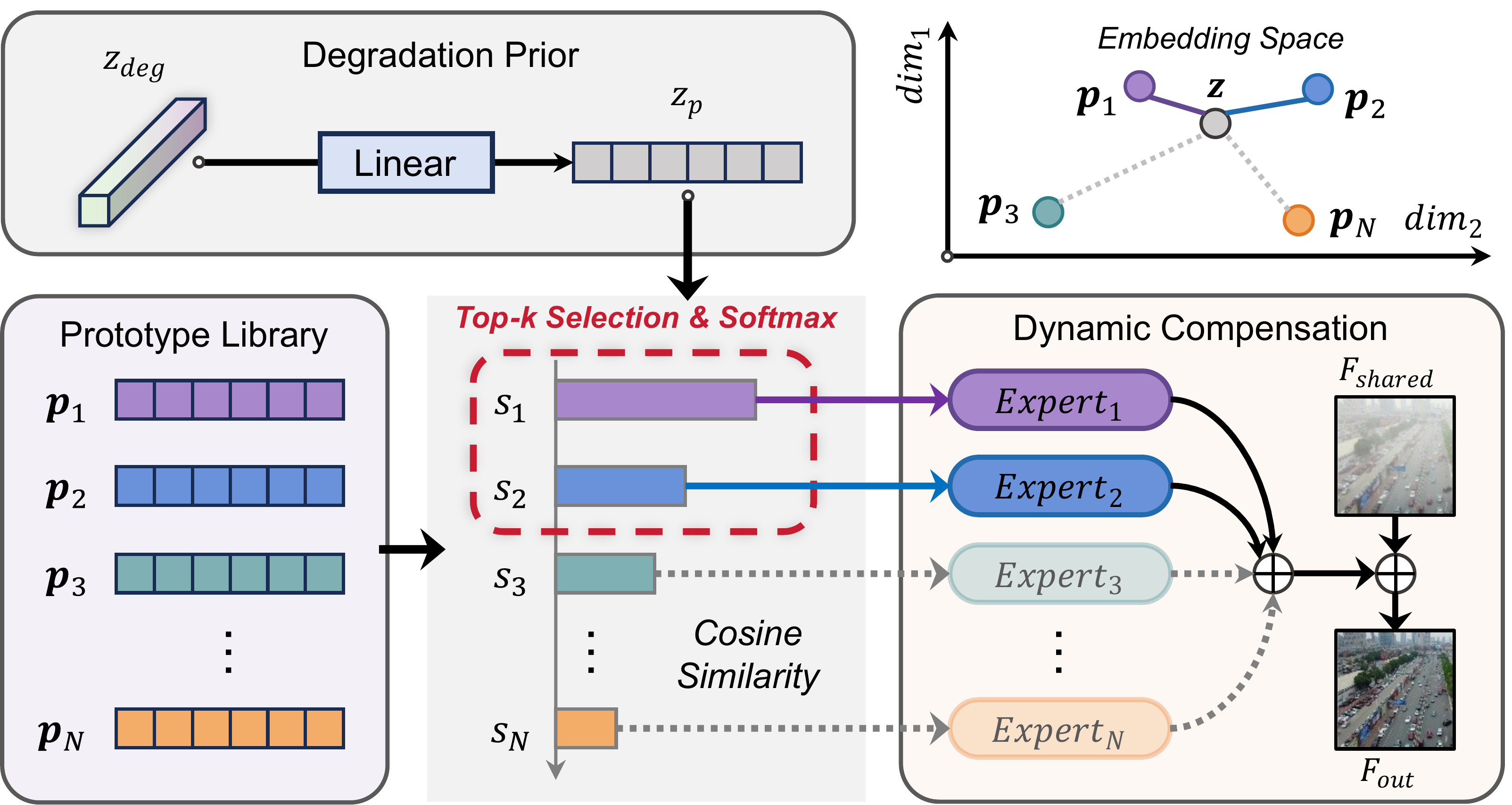}
  \caption{Architecture of the proposed \textbf{Prototype-Guided Router (PG-Router)}, which selects the most relevant low-rank residual experts for each input sample based on degradation archetypes.}
  \label{fig:pg_router}
  \vspace{-1em}
\end{figure}

We adopt a PG-Router to select low-rank residual experts, as shown in Fig.~\ref{fig:pg_router}. We maintain $N$ prototype vectors $\{p_n\}_{n=1}^{N}$, where each $p_n \in \mathbb{R}^{d_z}$ represents a learnable degradation archetype in the prior embedding space. Given the original degradation prior embedding $z_\text{deg}$, the router first maps it to a routing query:
\begin{equation}
  z_\text{p} = \text{Linear}\left(z_\text{deg}\right) \in \mathbb{R}^{d_z},
  \label{eq:routing_query}
\end{equation}
The routing logit for the $n$-th expert is then computed as the temperature-scaled cosine similarity between the normalized routing query and the corresponding prototype:
\begin{equation}
  s_n = \tau \cdot \hat{z}_\text{p}^\top \hat{p}_n, \quad n = 1, \ldots, N,
  \label{eq:routing_logits}
\end{equation}
where $\hat{z}_\text{p} = z_\text{p}/\|z_\text{p}\|_2$ and $\hat{p}_n = p_n/\|p_n\|_2$ are the $\ell_2$-normalized embeddings, and $\tau = \exp(t)$ is a learnable temperature controlling the sharpness of the routing distribution. The routing weights are then obtained by retaining the $k$ highest logits and normalizing:
\begin{equation}
  g = \text{Softmax}\big(\text{TopK}\left(s,\, k\right)\big) \in \mathbb{R}^{N},
  \label{eq:routing}
\end{equation}
where $\text{TopK}(\cdot, k)$ masks the remaining $N-k$ values to $-\infty$ before Softmax. This cosine-similarity-based design offers a natural interpretation: each prototype encodes a degradation archetype, and the routing score reflects how closely the current sample's degradation style aligns with that archetype. Sparse Top-$k$ activation is especially suitable for compound degradations because multiple low-rank experts can be activated simultaneously, while irrelevant experts remain suppressed.

\subsubsection{Low-Rank Residual Compensation}
The activated experts are aggregated to form a sample-adaptive low-rank residual branch. Meanwhile, the original backbone block serves as the common dense expert:
\begin{equation}
  E_\text{com}\left(F_\text{in}\right) = \text{BasicBlock}\left(F_\text{in}\right).
  \label{eq:core_dense}
\end{equation}
This dense expert preserves the full channel capacity of the backbone and is responsible for degradation-invariant restoration behavior. For an input feature $F \in \mathbb{R}^{H' \times W' \times C}$, the $n$-th low-rank expert first produces an individual residual response:
\begin{align}
  E_{\text{res},n}^\text{LR}\left(F\right)           & = B_n\left(A_n\left(F\right)\right), \label{eq:core_expert}                                           \\
  E_\text{res}^\text{LR}\left(F; z_\text{deg}\right) & = \sum_{n \in \mathcal{K}} g_n \cdot E_{\text{res},n}^\text{LR}\left(F\right), \, \label{eq:core_agg}
\end{align}
where $\mathcal{K}$ denotes the set of Top-$k$ activated experts and $g_n$ is the corresponding routing weight produced by the PG-Router conditioned on $z_\text{deg}$.

The final CoRE output combines the common dense expert and the routed low-rank residual experts through residual addition:
\begin{equation}
  F_\text{out} = \underbrace{E_\text{com}\left(F_\text{in}\right)}_{\text{common dense expert}} + \underbrace{E_\text{res}^\text{LR}\left(F_\text{in}; z_\text{deg}\right)}_{\text{low-rank residual experts}},
  \label{eq:core_residual}
\end{equation}
This formulation makes the common-residual division explicit: the common dense expert provides shared restoration capability, while the routed low-rank residual experts provide degradation-specific residual compensation. The efficiency gain comes from avoiding full-rank expert replication rather than sacrificing adaptive capacity.

\subsection{Global-Local Adaptive Network (GLA-Net)}
\label{sec:global_local_network}

We instantiate the universal restoration framework as a shared U-shaped encoder-decoder backbone with skip connections. The backbone contains several encoder stages, a bottleneck, and several decoder stages connected by strided downsampling and PixelShuffle upsampling \citep{shi2016real}. Each basic block is implemented by NAFBlock \citep{chen2022nafnet}. In implementation, GFM is applied once at the entrance of each stage, while CoRE is applied inside the stage for block-wise refinement.

Concretely, let $F_\text{in}^{i}$ and $F_\text{out}^{i}$ denote the input and output features of the $i$-th stage, respectively, and let $L_i$ be the number of blocks in that stage. The backbone first applies GFM once to the stage input and then recursively updates the feature through the stacked blocks:
\begin{align}
  F_{0}^{i}        & = \text{GFM}\left(F_\text{in}^{i}; z_\text{deg}\right), \label{eq:gla_gfm}       \\
  F_{j}^{i}        & = \text{CoRE}_{j}^{i}\left(F_{j-1}^{i}; z_\text{deg}\right), \label{eq:gla_core} \\
  F_\text{out}^{i} & = F_{L_i}^{i}. \label{eq:gla_output}
\end{align}

Thus, each stage performs one global modulation at its entrance, followed by recursive CoRE-based block updates. Repeating this pattern across the encoder, bottleneck, and decoder stages yields a compact implementation of GLA-Net, where GFM handles stage-level conditioning and CoRE handles block-level residual refinement.

\subsection{Training Strategy}
\label{sec:training_strategy}

CoRE-UIR adopts a two-phase training strategy.

\textbf{Phase I: Degradation Prior Adaptation.} We first optimize DPE with a multi-label degradation classification task over the $M$ base degradation types. The CLIP image encoder is frozen, and only the lightweight adapter and a linear classification head are trainable. The Phase-I objective is the binary cross-entropy loss:
\begin{equation}
  \mathcal{L}_\text{cls} = \text{BCE}\left(\text{Head}\left(z_\text{deg}\right), y\right),
  \label{eq:cls_loss}
\end{equation}
where $\text{Head}(\cdot)$ denotes the linear classification head, $y \in \{0,1\}^{M}$ is the multi-hot degradation label vector, and $\text{BCE}(\cdot)$ denotes the binary cross-entropy loss. Compound degradations are naturally represented by activating multiple entries in $y$. Since this phase involves very few trainable parameters, it converges quickly while adapting the pre-trained semantic representation model toward restoration-oriented embeddings for remote-sensing images.

\textbf{Phase II: Restoration Network Training.} After Phase I, the classification head is discarded and the entire DPE, including the CLIP encoder and adapter, is frozen. We then train only the instantiated restoration backbone for image restoration. The overall training objective combines pixel, structural, and perceptual losses:
\begin{equation}
  \mathcal{L}_\text{total} = \lambda_\text{pix} \mathcal{L}_\text{pix} + \lambda_\text{str} \mathcal{L}_\text{str} + \lambda_\text{per} \mathcal{L}_\text{per},
  \label{eq:total_loss}
\end{equation}
where $\mathcal{L}_\text{pix} = \|\hat{I} - I_\text{gt}\|_1$ denotes the pixel loss, $\mathcal{L}_\text{str} = 1 - \text{SSIM}(\hat{I}, I_\text{gt})$ denotes the structural loss \citep{SSIM}, and $\mathcal{L}_\text{per}$ denotes the perceptual loss implemented by LPIPS \citep{zhang2018unreasonable}. The concrete training hyperparameters are deferred to the implementation details in Section~\ref{subsubsec:implementation_details}.

Although the overall framework is trained in two phases, the additional cost of Phase I is small because only the lightweight adapter and classification head are optimized. More importantly, the pretrained DPE provides a stable and discriminative degradation embedding space, which makes the subsequent optimization of the CoRE-UIR restoration network more stable in Phase II.

\section{Experiments} \label{sec:experiments}

\subsection{Settings}
\label{sec:settings}

\begin{figure*}[t]
  \centering
  \begin{minipage}[t]{0.36\linewidth}
    \centering
    \includegraphics[width=\linewidth]{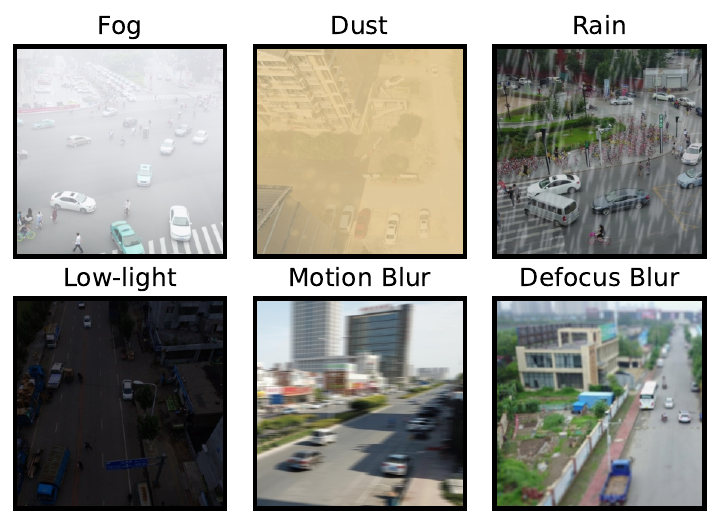}
    \small (a) MDVD-108K Single
  \end{minipage}\hfill
  \begin{minipage}[t]{0.36\linewidth}
    \centering
    \includegraphics[width=\linewidth]{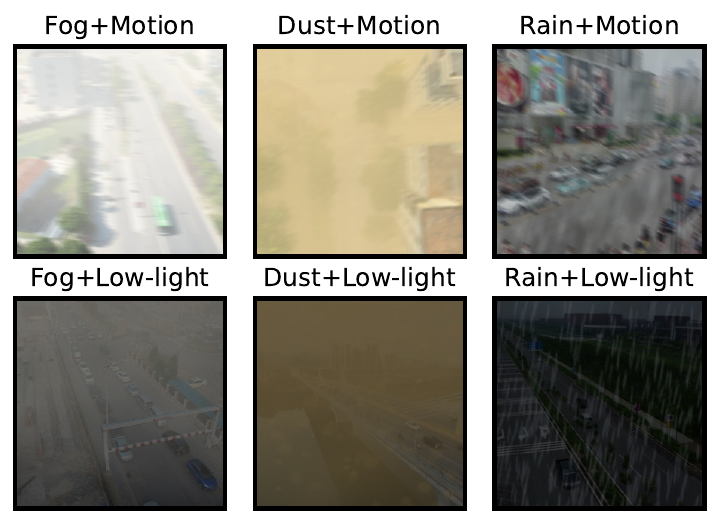}
    \small (b) MDVD-108K Compound
  \end{minipage}\hfill
  \begin{minipage}[t]{0.242\linewidth}
    \centering
    \includegraphics[width=\linewidth]{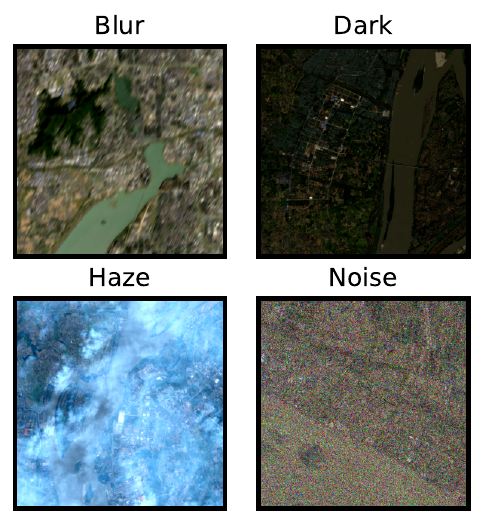}
    \small (c) MDRS-Landsat
  \end{minipage}

  \caption{Representative samples of the MDVD-108K and MDRS-Landsat datasets, showing (a) single degradations UAV samples, (b) compound degradations UAV samples, and (c) satellite samples from MDRS-Landsat.}
  \label{fig:dataset_samples}
  \vspace{-1em}
\end{figure*}

\begin{figure*}[t]
  \centering
  \includegraphics[width=0.98\linewidth]{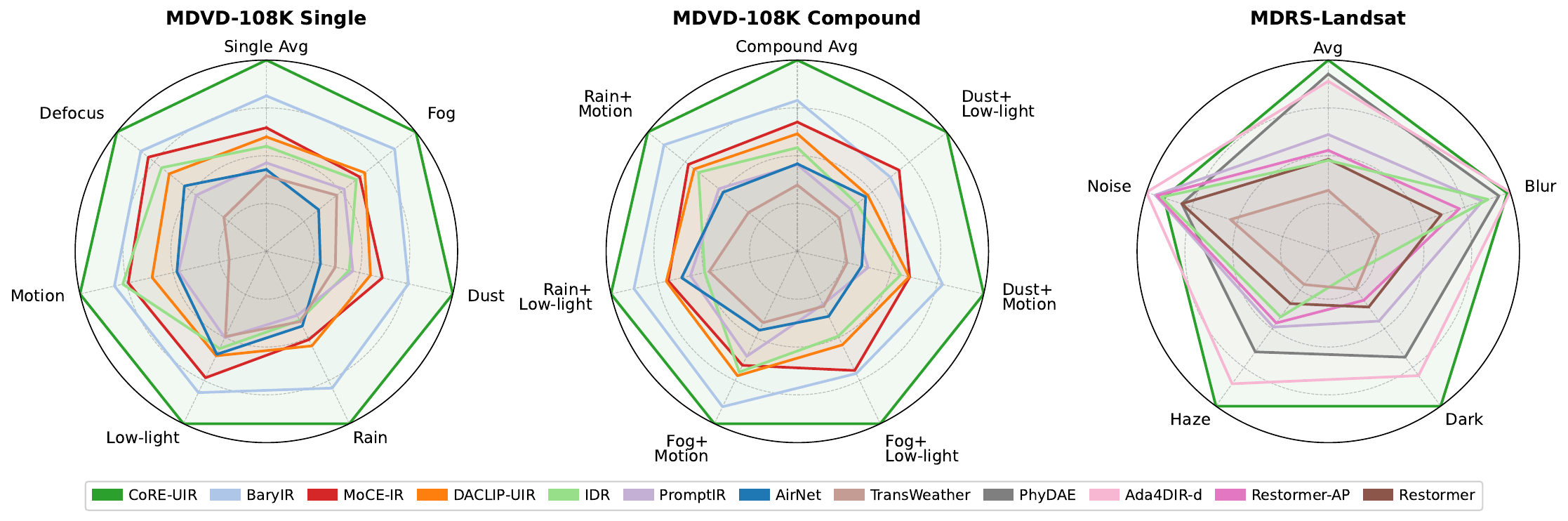}
  \caption{Radar-chart comparison of PSNR for representative algorithms. From left to right, the three panels summarize MDVD-108K single degradations, MDVD-108K compound degradations, and MDRS-Landsat. Larger radius indicates higher PSNR.}
  \label{fig:radar_psnr}
  \vspace{-1em}
\end{figure*}

\begin{table*}[t]
  \caption{Quantitative comparison on \textbf{weather-induced degradation} types (fog, dust, rain) from the MDVD-108K dataset. Best results are highlighted in \textbf{bold} and second-best are \underline{underlined}. $\uparrow$ indicates higher is better and $\downarrow$ indicates lower is better.}
  \label{tab:weather_degradation}
  \renewcommand{\arraystretch}{1.2}
  \centering
  \resizebox{\linewidth}{!}{%
    \begin{tabular}{ll|ccc|ccc|ccc|ccc}
      \toprule
      \multirow{2}{*}{\textbf{Method}}     & \multirow{2}{*}{\textbf{Venue}}
                                           & \multicolumn{3}{c|}{\textbf{Fog}}
                                           & \multicolumn{3}{c|}{\textbf{Dust}}
                                           & \multicolumn{3}{c|}{\textbf{Rain}}
                                           & \multicolumn{3}{c}{\textbf{Weather Average}}                                                                                                                                                                                                                                                         \\
                                           &                                              & PSNR$\uparrow$    & SSIM$\uparrow$     & LPIPS$\downarrow$
                                           & PSNR$\uparrow$                               & SSIM$\uparrow$    & LPIPS$\downarrow$
                                           & PSNR$\uparrow$                               & SSIM$\uparrow$    & LPIPS$\downarrow$
                                           & PSNR$\uparrow$                               & SSIM$\uparrow$    & LPIPS$\downarrow$                                                                                                                                                                                                                 \\
      \midrule
      \rowcolor{gray!20} \multicolumn{14}{l}{\textit{Single-task methods}}                                                                                                                                                                                                                                                                        \\
      \addlinespace[0.2em]
      MPRNet                               & CVPR'21                                      & 24.72             & 0.9277             & 0.0622             & 25.91             & 0.9043             & 0.1185             & 29.32             & 0.9590             & 0.0450             & 26.65             & 0.9303             & 0.0752             \\
      Restormer                            & CVPR'22                                      & 28.33             & 0.9553             & 0.0404             & 27.96             & 0.9208             & 0.0868             & 31.27             & 0.9718             & 0.0276             & 29.19             & 0.9493             & 0.0516             \\
      NAFNet                               & ECCV'22                                      & 26.77             & 0.9476             & 0.0382             & 26.54             & 0.9084             & 0.0696             & 30.95             & 0.9687             & 0.0281             & 28.09             & 0.9416             & 0.0453             \\
      DGUNet                               & CVPR'22                                      & 27.63             & 0.9452             & 0.0439             & 27.30             & 0.9028             & 0.1010             & 30.84             & 0.9601             & 0.0402             & 28.59             & 0.9360             & 0.0617             \\
      \midrule
      \rowcolor{gray!20} \multicolumn{14}{l}{\textit{All-in-One methods}}                                                                                                                                                                                                                                                                         \\
      \addlinespace[0.2em]
      TransWeather                         & CVPR'22                                      & 28.77             & 0.9533             & 0.0359             & 28.02             & 0.9128             & 0.0888             & 31.78             & 0.9656             & 0.0317             & 29.52             & 0.9439             & 0.0521             \\
      AirNet                               & CVPR'22                                      & 27.73             & 0.9537             & 0.0324             & 27.57             & 0.9179             & 0.0561             & 31.94             & 0.9728             & 0.0222             & 29.08             & 0.9482             & 0.0369             \\
      PromptIR                             & NeurIPS'23                                   & 29.20             & 0.9575             & 0.0266             & 28.56             & 0.9208             & 0.0494             & 31.56             & 0.9718             & 0.0215             & 29.78             & 0.9500             & 0.0325             \\
      IDR                                  & CVPR'23                                      & 29.90             & 0.9610             & 0.0329             & 28.45             & 0.9254             & 0.0762             & 31.74             & 0.9746             & 0.0221             & 30.03             & 0.9537             & 0.0438             \\
      DACLIP-UIR                           & ICLR'24                                      & 30.40             & 0.9619             & 0.0224             & 29.10             & 0.9255             & 0.0446             & 32.63             & 0.9755             & 0.0174             & 30.71             & 0.9543             & 0.0281             \\
      MoCE-IR                              & CVPR'25                                      & 30.10             & 0.9611             & 0.0234             & 29.46             & 0.9271             & 0.0440             & 32.41             & 0.9772             & 0.0162             & 30.66             & 0.9551             & 0.0279             \\
      BaryIR                               & TPAMI'26                                     & \underline{32.11} & \underline{0.9657} & \underline{0.0184} & \underline{30.27} & \underline{0.9304} & \underline{0.0372} & \underline{34.11} & \underline{0.9795} & \underline{0.0137} & \underline{32.16} & \underline{0.9585} & \underline{0.0231} \\
      \midrule
      \rowcolor{cyan!10} \textbf{CoRE-UIR} & \textit{Ours}                                & \textbf{33.33}    & \textbf{0.9673}    & \textbf{0.0161}    & \textbf{31.62}    & \textbf{0.9336}    & \textbf{0.0330}    & \textbf{35.35}    & \textbf{0.9817}    & \textbf{0.0114}    & \textbf{33.44}    & \textbf{0.9609}    & \textbf{0.0201}    \\
      \bottomrule
    \end{tabular}%
  }
  \vspace{-0.5em}
\end{table*}

\begin{table*}[t]
  \caption{Quantitative comparison on \textbf{imaging-induced degradation} types (low-light, motion blur, defocus blur) from the MDVD-108K dataset. Best results are highlighted in \textbf{bold} and second-best are \underline{underlined}. $\uparrow$ indicates higher is better and $\downarrow$ indicates lower is better.}
  \label{tab:imaging_degradation}
  \renewcommand{\arraystretch}{1.2}
  \centering
  \resizebox{\linewidth}{!}{%
    \begin{tabular}{ll|ccc|ccc|ccc|ccc}
      \toprule
      \multirow{2}{*}{\textbf{Method}}     & \multirow{2}{*}{\textbf{Venue}}
                                           & \multicolumn{3}{c|}{\textbf{Low-light}}
                                           & \multicolumn{3}{c|}{\textbf{Motion Blur}}
                                           & \multicolumn{3}{c|}{\textbf{Defocus Blur}}
                                           & \multicolumn{3}{c}{\textbf{Imaging Average}}                                                                                                                                                                                                                                                         \\
                                           &                                              & PSNR$\uparrow$    & SSIM$\uparrow$     & LPIPS$\downarrow$
                                           & PSNR$\uparrow$                               & SSIM$\uparrow$    & LPIPS$\downarrow$
                                           & PSNR$\uparrow$                               & SSIM$\uparrow$    & LPIPS$\downarrow$
                                           & PSNR$\uparrow$                               & SSIM$\uparrow$    & LPIPS$\downarrow$                                                                                                                                                                                                                 \\
      \midrule
      \rowcolor{gray!20} \multicolumn{14}{l}{\textit{Single-task methods}}                                                                                                                                                                                                                                                                        \\
      \addlinespace[0.2em]
      MPRNet                               & CVPR'21                                      & 22.15             & 0.8195             & 0.2042             & 25.64             & 0.7400             & 0.3351             & 28.43             & 0.8346             & 0.2190             & 25.41             & 0.7980             & 0.2528             \\
      Restormer                            & CVPR'22                                      & 24.41             & 0.8465             & 0.1642             & 27.43             & 0.8053             & 0.2278             & 29.66             & 0.8653             & 0.1495             & 27.16             & 0.8390             & 0.1805             \\
      NAFNet                               & ECCV'22                                      & 24.12             & 0.8394             & 0.1035             & 26.42             & 0.7602             & 0.1237             & 29.38             & 0.8495             & 0.0747             & 26.64             & 0.8164             & 0.1006             \\
      DGUNet                               & CVPR'22                                      & 24.57             & 0.8103             & 0.1847             & 25.09             & 0.6831             & 0.3793             & 28.51             & 0.8239             & 0.1899             & 26.06             & 0.7724             & 0.2513             \\
      \midrule
      \rowcolor{gray!20} \multicolumn{14}{l}{\textit{All-in-One methods}}                                                                                                                                                                                                                                                                         \\
      \addlinespace[0.2em]
      TransWeather                         & CVPR'22                                      & 25.82             & 0.8255             & 0.1682             & 25.83             & 0.7216             & 0.3281             & 29.09             & 0.8385             & 0.1755             & 26.91             & 0.7952             & 0.2239             \\
      AirNet                               & CVPR'22                                      & 26.59             & 0.8517             & 0.0929             & 26.86             & 0.7860             & 0.1003             & 29.71             & 0.8612             & 0.0594             & 27.72             & 0.8330             & 0.0842             \\
      PromptIR                             & NeurIPS'23                                   & 25.88             & 0.8496             & 0.0892             & 26.83             & 0.7785             & 0.0972             & 29.52             & 0.8575             & 0.0602             & 27.41             & 0.8285             & 0.0822             \\
      IDR                                  & CVPR'23                                      & 26.35             & 0.8592             & 0.1528             & 27.94             & \underline{0.8233} & 0.1886             & 30.06             & 0.8738             & 0.1329             & 28.11             & 0.8521             & 0.1581             \\
      DACLIP-UIR                           & ICLR'24                                      & 26.66             & 0.8560             & 0.0817             & 27.35             & 0.7994             & 0.0839             & 29.94             & 0.8660             & 0.0516             & 27.98             & 0.8405             & 0.0724             \\
      MoCE-IR                              & CVPR'25                                      & 27.60             & 0.8589             & 0.0782             & 27.83             & 0.8137             & 0.0789             & 30.27             & 0.8727             & 0.0510             & 28.57             & 0.8484             & 0.0694             \\
      BaryIR                               & TPAMI'26                                     & \underline{28.24} & \underline{0.8653} & \underline{0.0706} & \underline{28.10} & 0.8180             & \underline{0.0710} & \underline{30.38} & \underline{0.8747} & \underline{0.0451} & \underline{28.91} & \underline{0.8527} & \underline{0.0622} \\
      \midrule
      \rowcolor{cyan!10} \textbf{CoRE-UIR} & \textit{Ours}                                & \textbf{29.58}    & \textbf{0.8702}    & \textbf{0.0645}    & \textbf{28.78}    & \textbf{0.8376}    & \textbf{0.0600}    & \textbf{30.75}    & \textbf{0.8812}    & \textbf{0.0411}    & \textbf{29.71}    & \textbf{0.8630}    & \textbf{0.0552}    \\
      \bottomrule
    \end{tabular}%
  }
  \vspace{-0.5em}
\end{table*}

\begin{figure*}[t]
  \centering
  \begin{minipage}[c]{0.01\linewidth}
    \rotatebox{90}{Fog}
  \end{minipage}
  \begin{minipage}[c]{0.95\linewidth}
    \includegraphics[width=\linewidth]{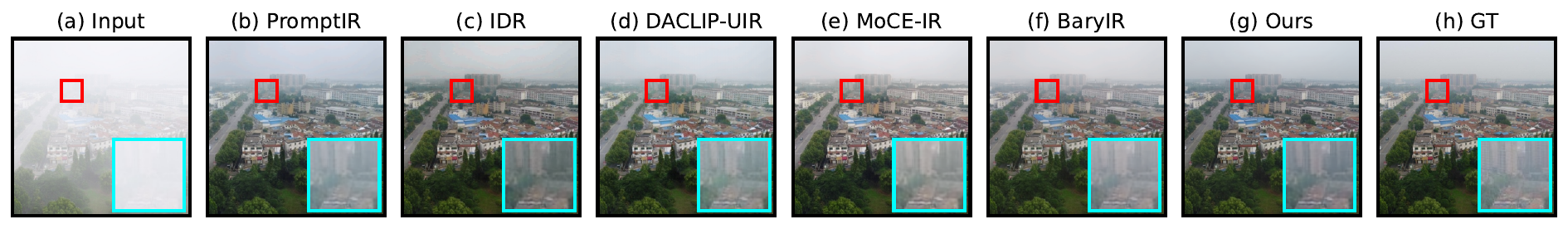}
    \includegraphics[width=\linewidth]{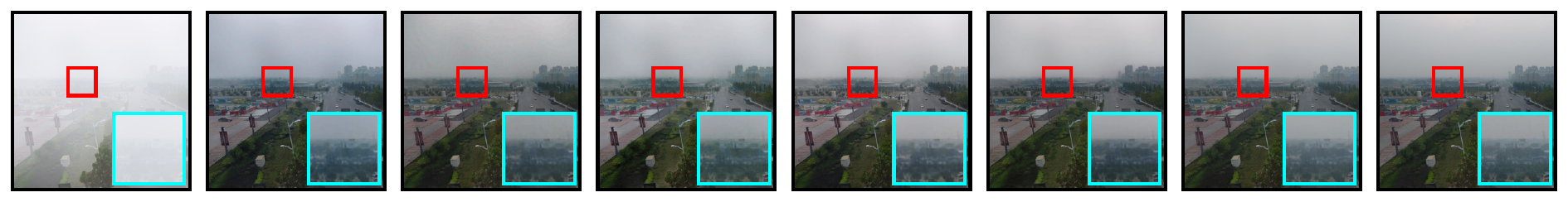}
  \end{minipage}

  \begin{minipage}[c]{0.01\linewidth}
    \rotatebox{90}{Dust}
  \end{minipage}
  \begin{minipage}[c]{0.95\linewidth}
    \includegraphics[width=\linewidth]{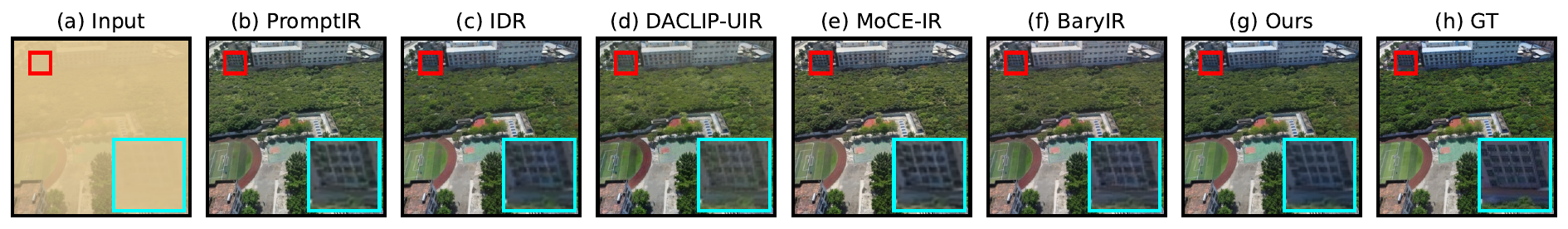}
    \includegraphics[width=\linewidth]{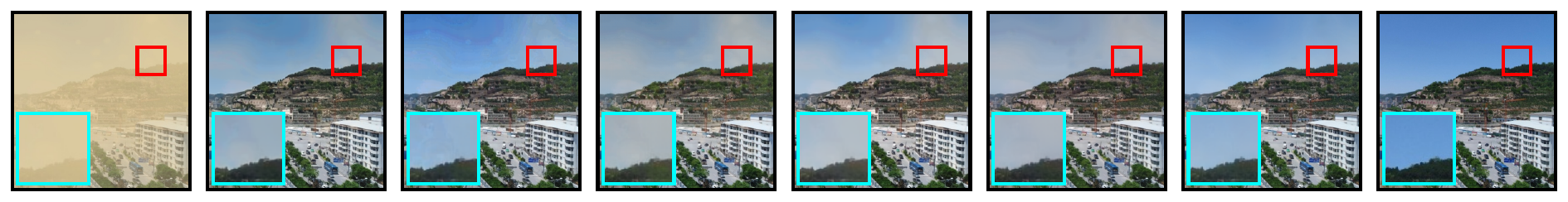}
  \end{minipage}

  \begin{minipage}[c]{0.01\linewidth}
    \rotatebox{90}{Rain}
  \end{minipage}
  \begin{minipage}[c]{0.95\linewidth}
    \includegraphics[width=\linewidth]{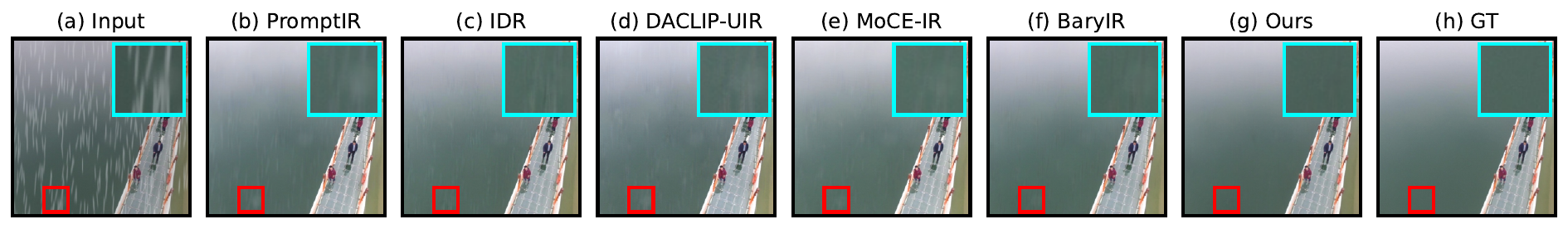}
    \includegraphics[width=\linewidth]{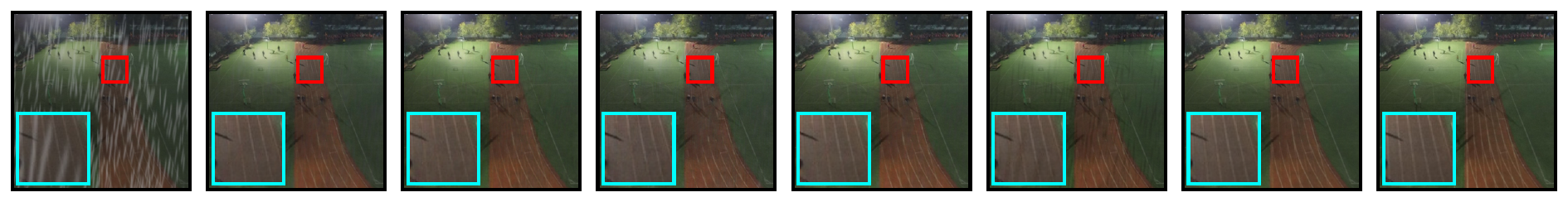}
  \end{minipage}

  \caption{Visual comparison on \textbf{weather-induced degradation} cases from MDVD-108K. CoRE-UIR produces more natural visibility, contrast, and color than prior universal restoration baselines, which is consistent with the role of DPE and GFM in organizing global appearance modulation under weather-induced degradations.}
  \label{fig:visual_single_1}
  \vspace{-1em}
\end{figure*}

\begin{figure*}[t]
  \centering
  \begin{minipage}[c]{0.01\linewidth}
    \rotatebox{90}{Low-light}
  \end{minipage}
  \begin{minipage}[c]{0.95\linewidth}
    \includegraphics[width=\linewidth]{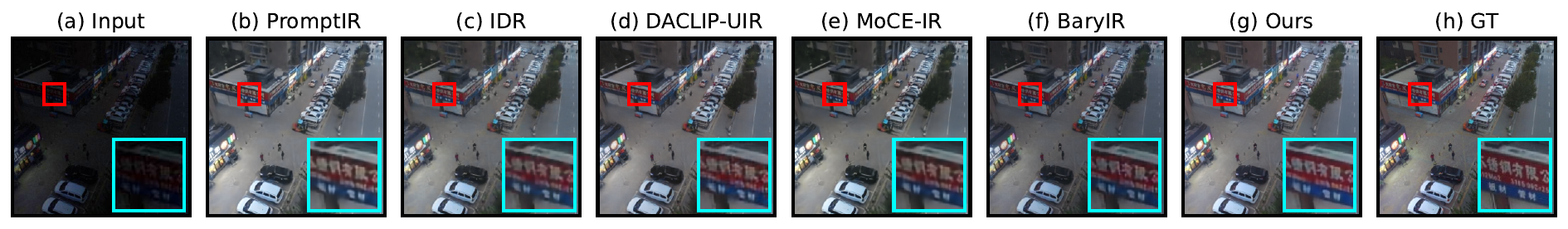}
    \includegraphics[width=\linewidth]{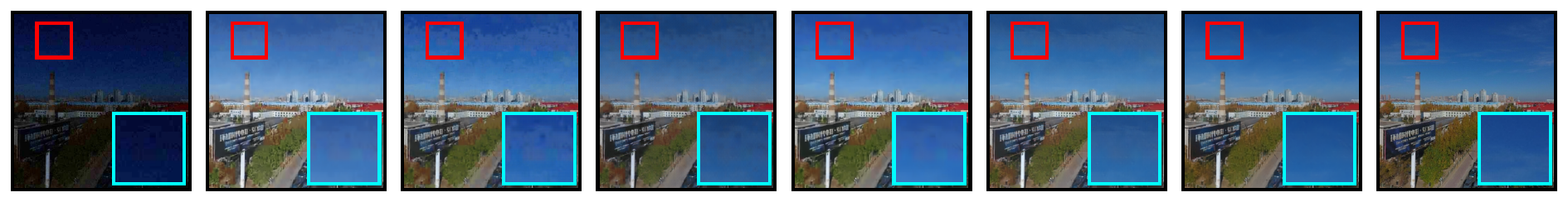}
  \end{minipage}

  \begin{minipage}[c]{0.01\linewidth}
    \rotatebox{90}{Motion Blur}
  \end{minipage}
  \begin{minipage}[c]{0.95\linewidth}
    \includegraphics[width=\linewidth]{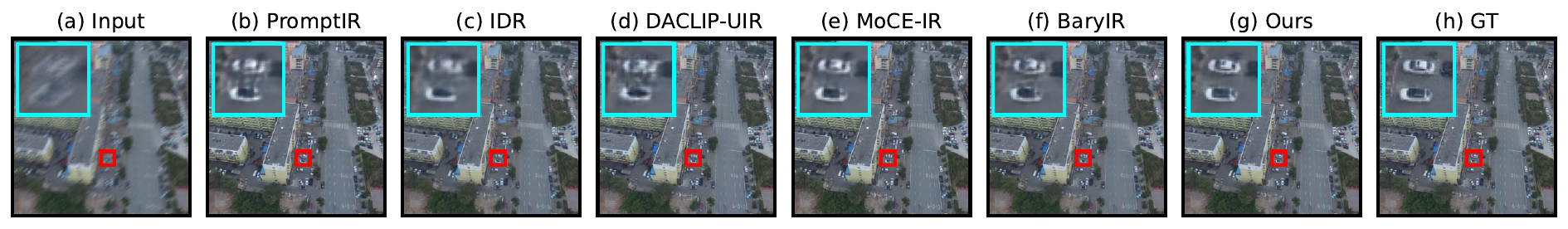}
    \includegraphics[width=\linewidth]{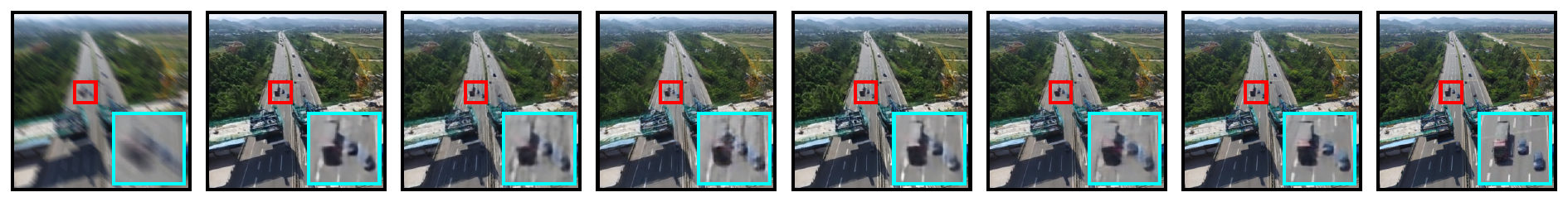}
  \end{minipage}

  \begin{minipage}[c]{0.01\linewidth}
    \rotatebox{90}{Defocus Blur}
  \end{minipage}
  \begin{minipage}[c]{0.95\linewidth}
    \includegraphics[width=\linewidth]{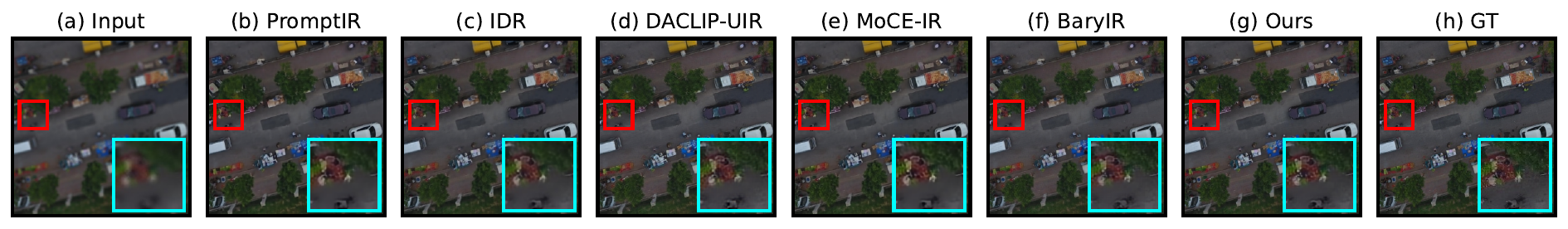}
    \includegraphics[width=\linewidth]{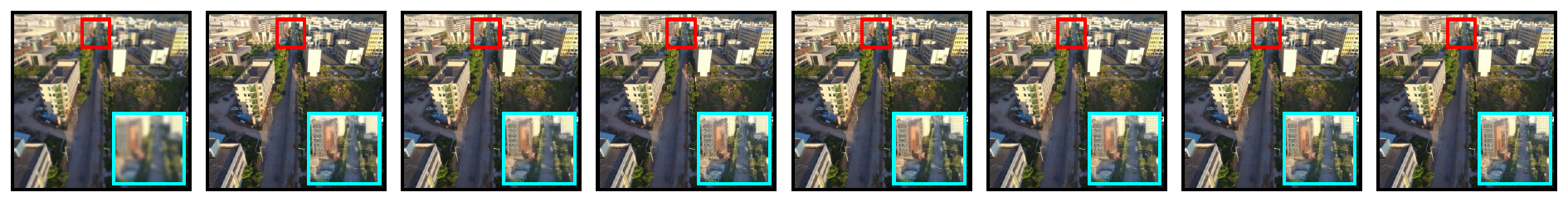}
  \end{minipage}

  \caption{Visual comparison on \textbf{imaging-induced degradation} cases from MDVD-108K. CoRE-UIR produces more uniform brightness correction and sharper structural recovery than prior universal restoration baselines, reflecting the benefit of low-rank residual specialization for local degradation compensation.}
  \label{fig:visual_single_2}
  \vspace{-1em}
\end{figure*}

\begin{figure*}[ht]
  \centering
  \begin{minipage}[c]{0.015\linewidth}
    \rotatebox{90}{Fog + Motion Blur}
  \end{minipage}
  \begin{minipage}[c]{0.95\linewidth}
    \includegraphics[width=\linewidth]{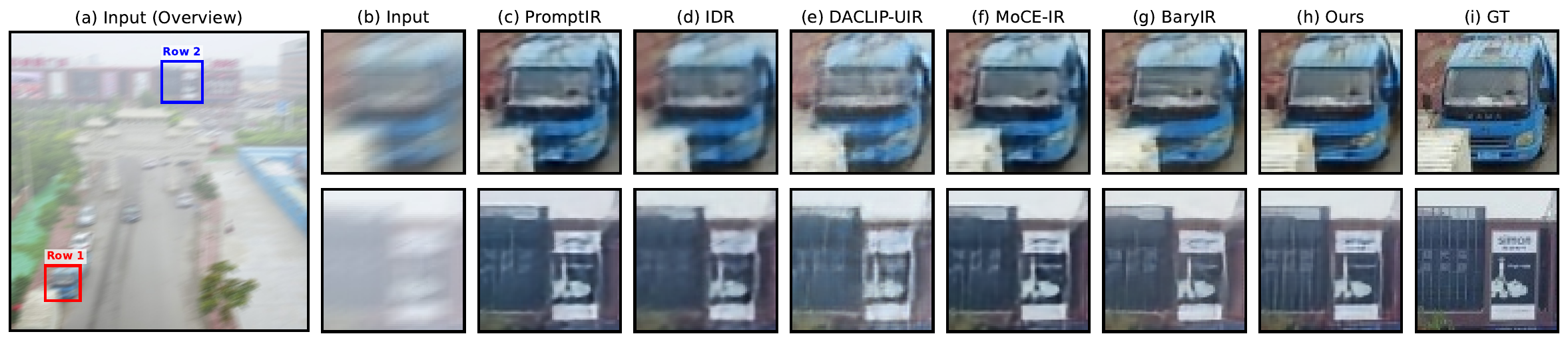}
  \end{minipage}

  \begin{minipage}[c]{0.015\linewidth}
    \rotatebox{90}{Dust + Motion Blur}
  \end{minipage}
  \begin{minipage}[c]{0.95\linewidth}
    \includegraphics[width=\linewidth]{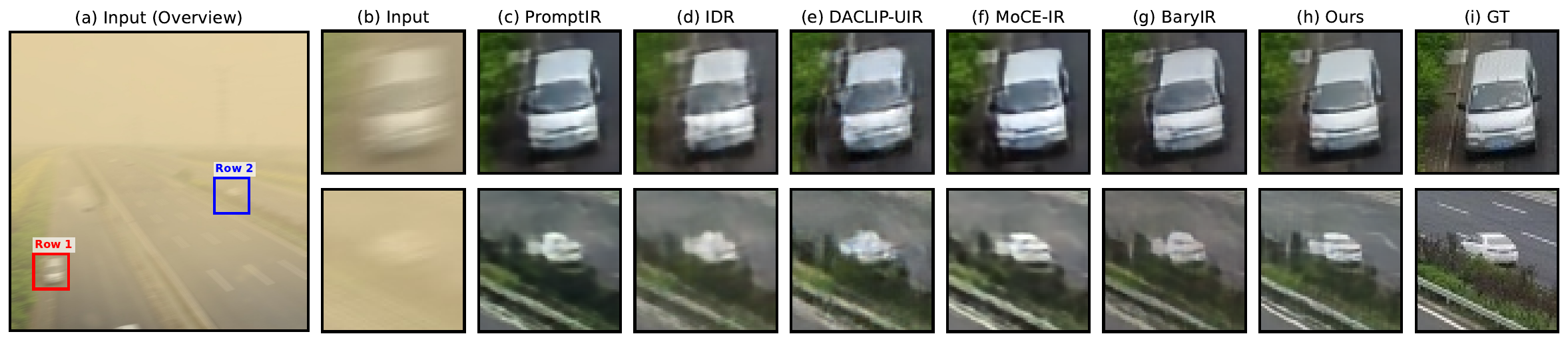}
  \end{minipage}

  \begin{minipage}[c]{0.015\linewidth}
    \rotatebox{90}{Rain + Low-light}
  \end{minipage}
  \begin{minipage}[c]{0.95\linewidth}
    \includegraphics[width=\linewidth]{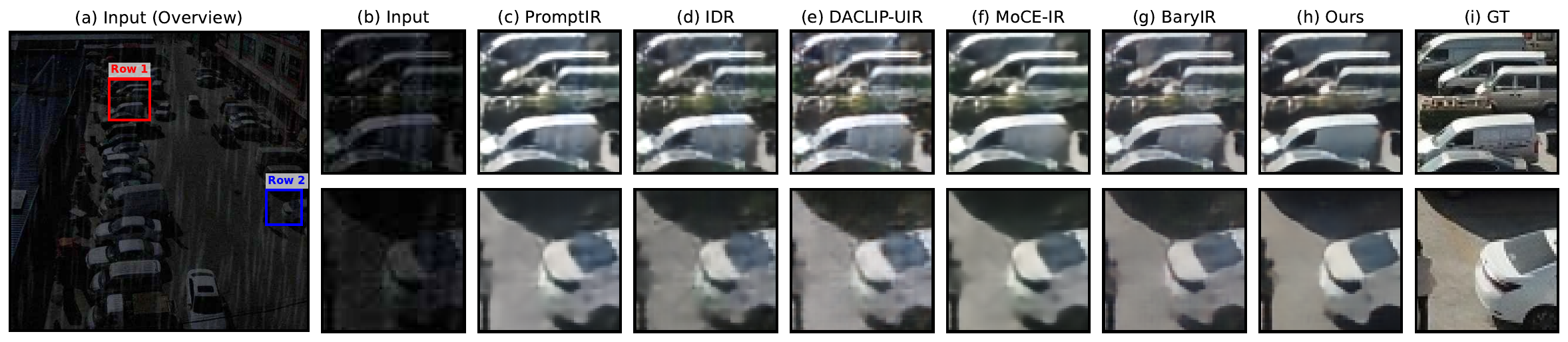}
  \end{minipage}

  \begin{minipage}[c]{0.015\linewidth}
    \rotatebox{90}{Fog + Low-light}
  \end{minipage}
  \begin{minipage}[c]{0.95\linewidth}
    \includegraphics[width=\linewidth]{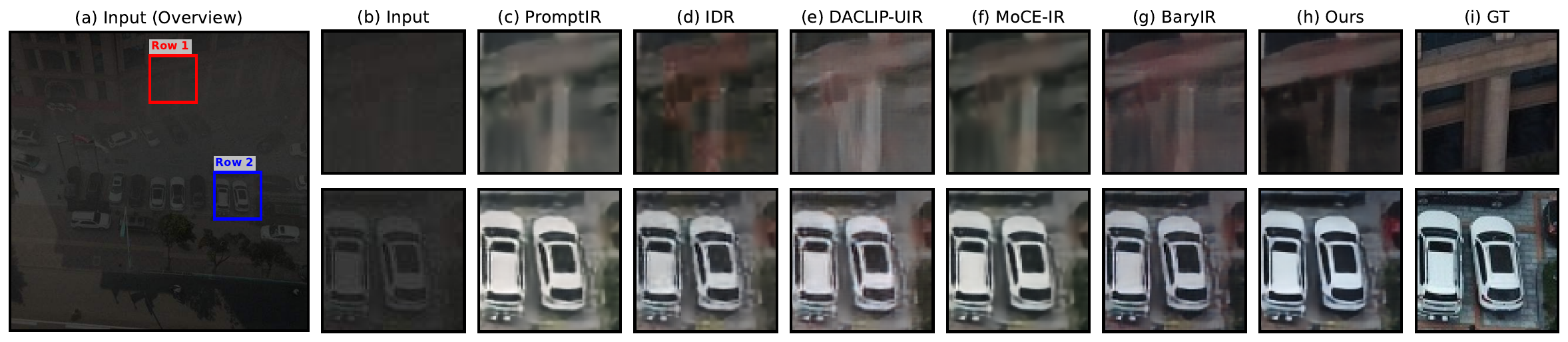}
  \end{minipage}

  \caption{Qualitative comparison for \textbf{compound degradation} cases. CoRE-UIR produces cleaner structures and fewer residual compound artifacts than prior universal restoration baselines under overlapping degradations, where the common dense path handles common restoration while low-rank residual experts compensate degradation-specific residuals.}
  \label{fig:visual_compound}
  \vspace{-1em}
\end{figure*}

\subsubsection{Datasets}

We evaluate CoRE-UIR on two public remote sensing datasets covering UAV and satellite imagery, respectively. Representative samples from these datasets are shown in Fig.~\ref{fig:dataset_samples}.

\textbf{MDVD-108K.} We construct a large-scale multi-degra\-dation UAV aerial dataset based on the VisDrone \citep{zhu2021detection} static detection imagery\footnote{The original VisDrone labels are mapped into three coarse categories: person (pedestrian/people), micro\_vehicle (bicycle/tricycle/awning-tricycle/motor), and vehicle (car/van/truck/bus).}. Degraded images are synthesized using Depth Anything V2 \citep{yang2024depth} for scene depth estimation and physically-grounded degradation models. The dataset contains six single degradations grouped into two categories: \textbf{weather-induced} degradations (fog, dust, rain), and \textbf{imaging-induced} degradations (low-light, motion blur, defocus blur). It further contains six selected cross-category compound degradations constructed by pairing the three weather-induced degradations with two imaging-induced degradations, namely motion blur and low-light. To avoid data leakage, the clean source images strictly follow the original VisDrone train/val/test partition before degradation synthesis, and all synthetic pairs are generated within their respective splits. In total, MDVD-108K comprises 108{,}500 images ($512\times512$): 108{,}000 synthetic paired samples (86{,}400/10{,}800/10{,}800 for training/validation/testing) and 500 real-world degraded UAV images collected via manual screening, which serve as a \textbf{real-world test set} for qualitative evaluation. Detailed synthesis protocols are provided in Appendix~\ref{appendix:dataset_synthesis}.

\textbf{MDRS-Landsat.} For satellite-domain evaluation, we employ MDRS-Landsat introduced by \citet{lihe2025ada4dir}. This dataset is designed for four restoration tasks, including deblurring, denoising, dehazing, and dedarkening. It is constructed from 5,400 clean Landsat-8 images ($512\times512$) from RSHaze \citep{song2023vision}, and each degradation subset is split into 5,130/270 samples for training/testing. Following \citet{lihe2025ada4dir}, degradations are synthesized with task-specific protocols, including anisotropic Gaussian blur, Gaussian noise, band-varied non-uniform haze based on Landsat-8 B9 (cirrus) information and atmospheric scattering, and power-law darkening transformation. In summary, MDRS-Landsat contains 20,520 training pairs and 1,080 testing pairs.

\subsubsection{Comparison Methods}
We compare CoRE-UIR against two categories of methods:

\textit{Single-task methods}: MPRNet \citep{zamir2021mprnet}, Restormer \citep{zamir2022restormer}, NAFNet \citep{chen2022nafnet}, and DGUNet \citep{mou2022dgunet}. For fair comparison, each single-task method is trained separately for every restoration subtask, and the reported averages aggregate the corresponding dedicated checkpoints.

\textit{All-in-One methods}: TransWeather \citep{valanarasu2022transweather}, AirNet \citep{li2022airnet}, PromptIR \citep{potlapalli2024promptir}, IDR \citep{zhang2023idr}, DACLIP-UIR\footnote{DACLIP-UIR follows its official 100-step inference setting for both restoration evaluation and efficiency measurement.} \citep{luo2023daclip}, Restormer-AP \citep{kong2024towards}, MoCE-IR \citep{zamfir2025complexity}, and BaryIR \citep{tang2026learning}.

In addition, we compared two physics-based models developed specifically for the four-degradation scenario in MDRS-Landsat: Ada4DIR \citep{lihe2025ada4dir} and PhyDAE \citep{dong2026phydae}.

\subsubsection{Evaluation Metrics}
For restoration quality, PSNR measures pixel-level reconstruction fidelity in decibels, SSIM \citep{SSIM}\footnote{PSNR and SSIM are computed on full RGB images without border cropping, per channel and then averaged.} evaluates structural consistency in terms of luminance, contrast, and local pattern similarity, and LPIPS \citep{zhang2018unreasonable} measures perceptual distance in a deep feature space and better reflects visual realism. Higher PSNR and SSIM indicate better restoration quality, while lower LPIPS indicates better perceptual quality.

For the Phase-I proxy task in the DPE ablations, we use multi-label degradation classification over the $M=6$ base degradation types, so compound samples can activate multiple labels. Precision, Recall, and F1 are macro-averaged.

\begin{table*}[t]
  \caption{Quantitative comparison on \textbf{compound degradation} types from the MDVD-108K dataset. Three representative pairwise combinations are shown, together with the average over all six compound types. Best results are highlighted in \textbf{bold} and second-best are \underline{underlined}. $\uparrow$ indicates higher is better and $\downarrow$ indicates lower is better.}
  \label{tab:compound_degradation}
  \renewcommand{\arraystretch}{1.2}
  \centering
  \resizebox{\linewidth}{!}{%
    \begin{tabular}{ll|ccc|ccc|ccc|ccc}
      \toprule
      \multirow{2}{*}{\textbf{Method}}     & \multirow{2}{*}{\textbf{Venue}}
                                           & \multicolumn{3}{c|}{\textbf{Rain+Low-light}}
                                           & \multicolumn{3}{c|}{\textbf{Dust+Motion}}
                                           & \multicolumn{3}{c|}{\textbf{Fog+Motion}}
                                           & \multicolumn{3}{c}{\textbf{Compound Average}}                                                                                                                                                                                                                                                         \\
                                           &
                                           & PSNR$\uparrow$                                & SSIM$\uparrow$    & LPIPS$\downarrow$
                                           & PSNR$\uparrow$                                & SSIM$\uparrow$    & LPIPS$\downarrow$
                                           & PSNR$\uparrow$                                & SSIM$\uparrow$    & LPIPS$\downarrow$
                                           & PSNR$\uparrow$                                & SSIM$\uparrow$    & LPIPS$\downarrow$                                                                                                                                                                                                                 \\
      \midrule
      \rowcolor{gray!20} \multicolumn{14}{l}{\textit{Single-task methods}}                                                                                                                                                                                                                                                                         \\
      \addlinespace[0.2em]
      MPRNet                               & CVPR'21                                       & 21.31             & 0.7219             & 0.3336             & 22.09             & 0.6897             & 0.4364             & 20.26             & 0.6756             & 0.4229             & 20.86             & 0.7067             & 0.3846             \\
      Restormer                            & CVPR'22                                       & 23.47             & 0.7644             & 0.2820             & 23.46             & 0.7498             & 0.3289             & 22.97             & 0.7568             & 0.3007             & 22.55             & 0.7591             & 0.2905             \\
      NAFNet                               & ECCV'22                                       & 21.15             & 0.7384             & 0.2126             & 22.53             & 0.7042             & 0.2104             & 22.13             & 0.7118             & 0.1810             & 21.78             & 0.7269             & 0.1974             \\
      DGUNet                               & CVPR'22                                       & 22.93             & 0.6979             & 0.3272             & 22.63             & 0.6308             & 0.4585             & 21.20             & 0.6240             & 0.4563             & 21.65             & 0.6640             & 0.3966             \\
      \midrule
      \rowcolor{gray!20} \multicolumn{14}{l}{\textit{All-in-One methods}}                                                                                                                                                                                                                                                                          \\
      \addlinespace[0.2em]
      TransWeather                         & CVPR'22                                       & 23.72             & 0.7180             & 0.3034             & 23.18             & 0.6708             & 0.4153             & 22.72             & 0.6762             & 0.4060             & 22.69             & 0.6977             & 0.3583             \\
      AirNet                               & CVPR'22                                       & 24.53             & 0.7646             & 0.1871             & 23.50             & 0.7271             & 0.1738             & 22.98             & 0.7375             & 0.1516             & 23.27             & 0.7493             & 0.1712             \\
      PromptIR                             & NeurIPS'23                                    & 24.27             & 0.7654             & 0.1788             & 23.64             & 0.7306             & 0.1616             & 23.88             & 0.7415             & 0.1376             & 23.25             & 0.7523             & 0.1607             \\
      IDR                                  & CVPR'23                                       & 23.86             & 0.7764             & 0.2604             & 24.35             & \underline{0.7689} & 0.2929             & 24.42             & 0.7826             & 0.2571             & 23.72             & 0.7790             & 0.2603             \\
      DACLIP-UIR                           & ICLR'24                                       & 24.98             & 0.7745             & 0.1638             & 24.53             & 0.7509             & 0.1465             & 24.55             & 0.7623             & 0.1218             & 24.10             & 0.7686             & 0.1433             \\
      MoCE-IR                              & CVPR'25                                       & 24.92             & 0.7743             & 0.1628             & 24.54             & 0.7552             & 0.1461             & 24.19             & 0.7663             & 0.1247             & 24.43             & 0.7729             & 0.1420             \\
      BaryIR                               & TPAMI'26                                      & \underline{25.92} & \underline{0.7866} & \underline{0.1482} & \underline{25.26} & 0.7677             & \underline{0.1269} & \underline{25.62} & \underline{0.7828} & \underline{0.1037} & \underline{25.02} & \underline{0.7840} & \underline{0.1262} \\
      \midrule
      \rowcolor{cyan!10} \textbf{CoRE-UIR} & \textit{Ours}                                 & \textbf{26.65}    & \textbf{0.7937}    & \textbf{0.1362}    & \textbf{26.05}    & \textbf{0.7847}    & \textbf{0.1148}    & \textbf{25.92}    & \textbf{0.8003}    & \textbf{0.0937}    & \textbf{26.17}    & \textbf{0.7989}    & \textbf{0.1136}    \\
      \bottomrule
    \end{tabular}%
  }
  \vspace{-0.5em}
\end{table*}

\begin{figure*}[t!]
  \centering
  \begin{minipage}[c]{0.015\linewidth}
    \rotatebox{90}{Real Fog}
  \end{minipage}
  \begin{minipage}[c]{0.95\linewidth}
    \includegraphics[width=\linewidth]{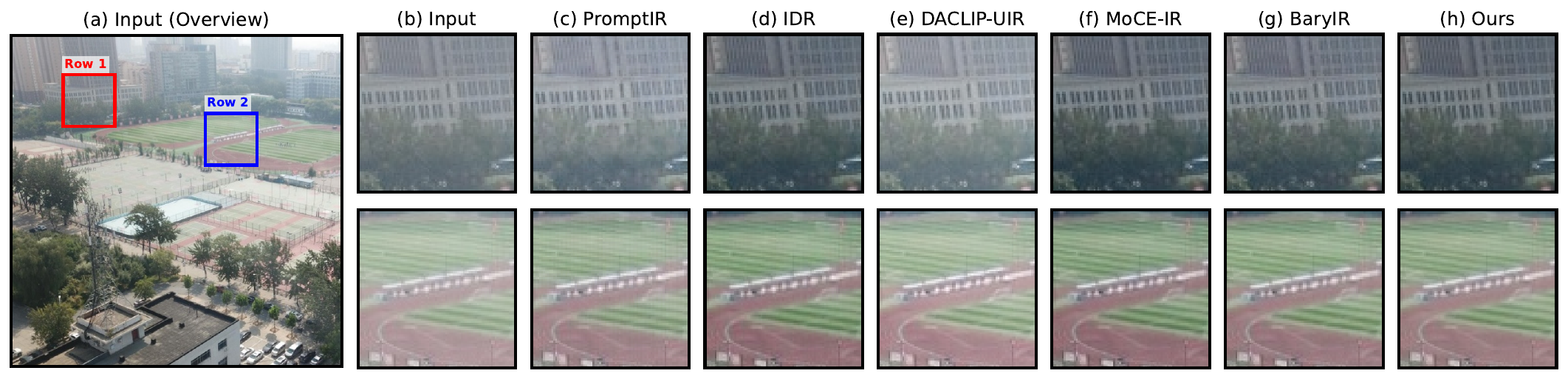}
  \end{minipage}

  \begin{minipage}[c]{0.015\linewidth}
    \rotatebox{90}{Real Motion Blur}
  \end{minipage}
  \begin{minipage}[c]{0.95\linewidth}
    \includegraphics[width=\linewidth]{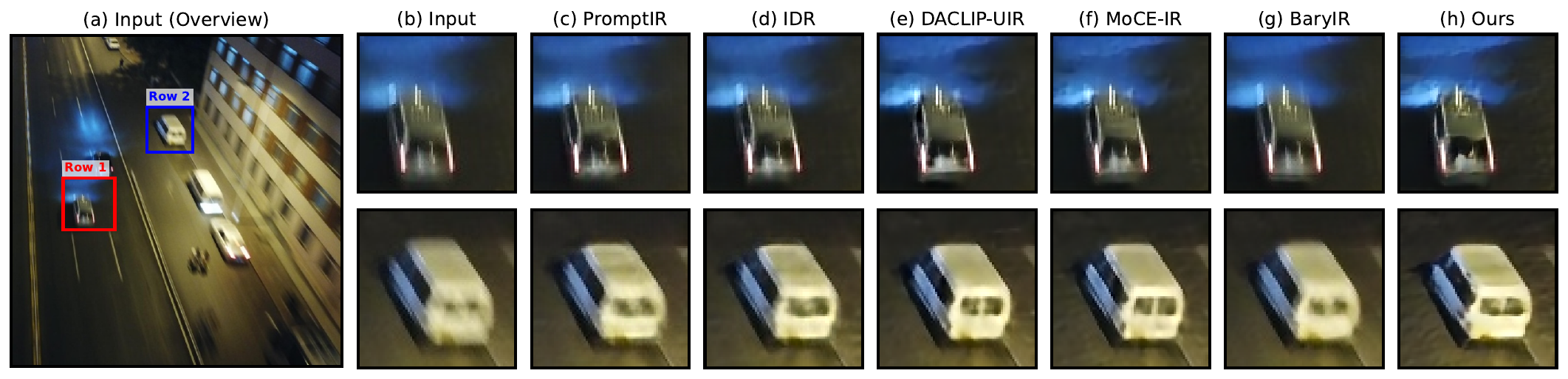}
  \end{minipage}

  \begin{minipage}[c]{0.015\linewidth}
    \rotatebox{90}{Real Low-light}
  \end{minipage}
  \begin{minipage}[c]{0.95\linewidth}
    \includegraphics[width=\linewidth]{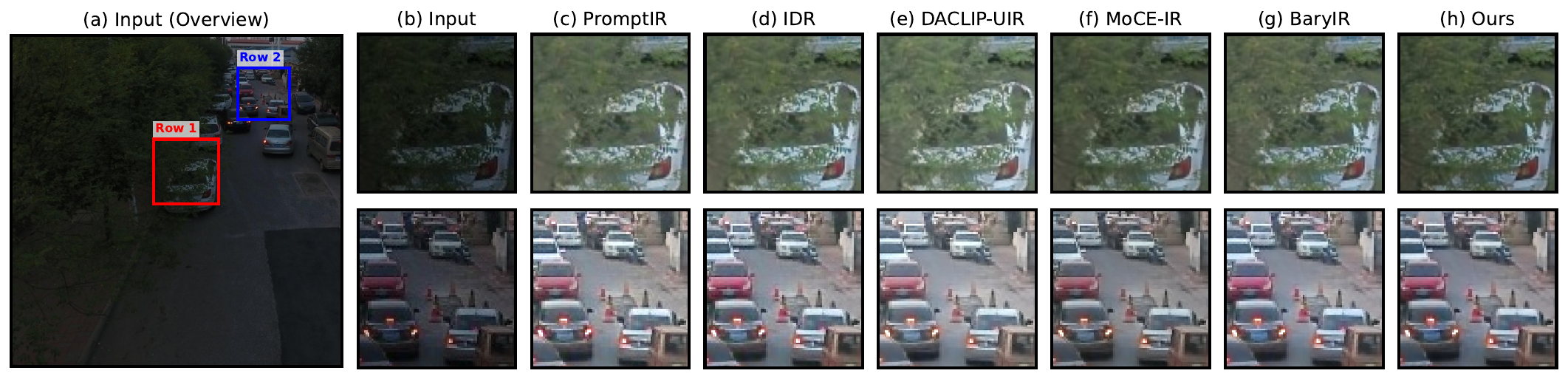}
  \end{minipage}

  \begin{minipage}[c]{0.015\linewidth}
    \rotatebox{90}{Real Defocus Blur}
  \end{minipage}
  \begin{minipage}[c]{0.95\linewidth}
    \includegraphics[width=\linewidth]{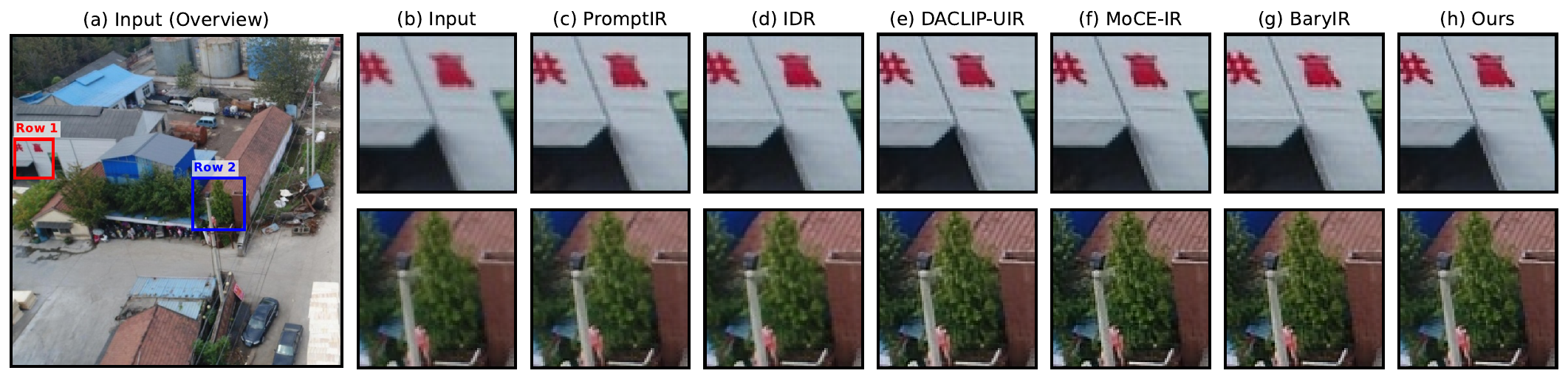}
  \end{minipage}
  \caption{Qualitative comparison for \textbf{real-world degradation} results. CoRE-UIR preserves land-cover structures while suppressing authentic degradation artifacts more reliably than competing universal restoration methods.}
  \label{fig:visual_real}
  \vspace{-1em}
\end{figure*}

\begin{figure*}[ht]
  \centering
  \begin{minipage}[c]{0.015\linewidth}
    \rotatebox{90}{Blur}
  \end{minipage}
  \begin{minipage}[c]{0.95\linewidth}
    \includegraphics[width=\linewidth]{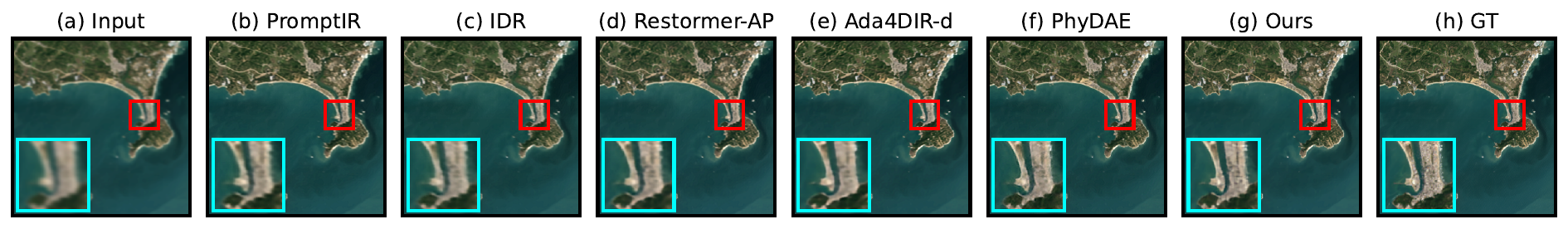}
    \includegraphics[width=\linewidth]{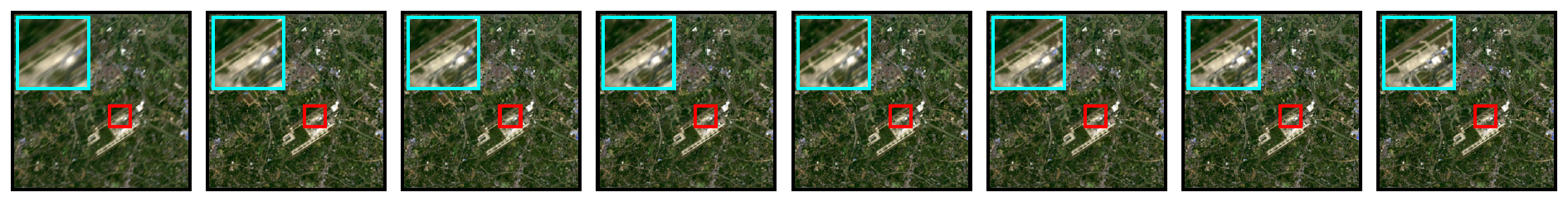}
  \end{minipage}

  \begin{minipage}[c]{0.015\linewidth}
    \rotatebox{90}{Dark}
  \end{minipage}
  \begin{minipage}[c]{0.95\linewidth}
    \includegraphics[width=\linewidth]{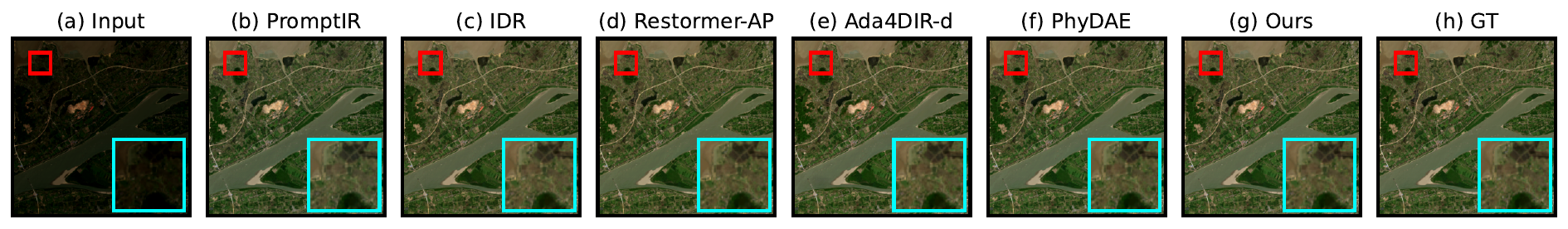}
    \includegraphics[width=\linewidth]{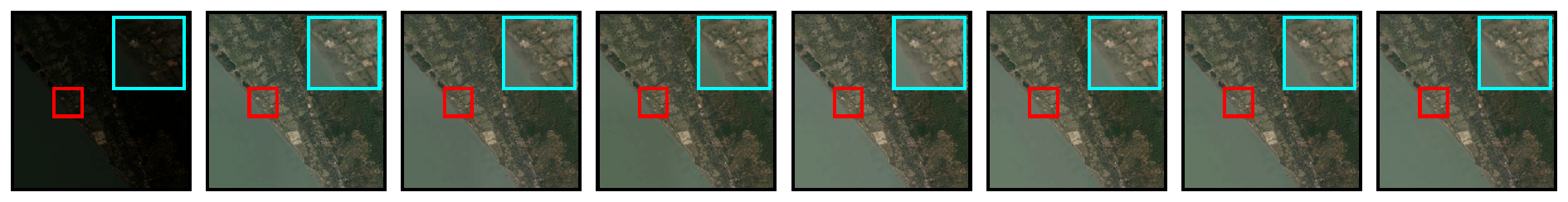}
  \end{minipage}

  \begin{minipage}[c]{0.015\linewidth}
    \rotatebox{90}{Haze}
  \end{minipage}
  \begin{minipage}[c]{0.95\linewidth}
    \includegraphics[width=\linewidth]{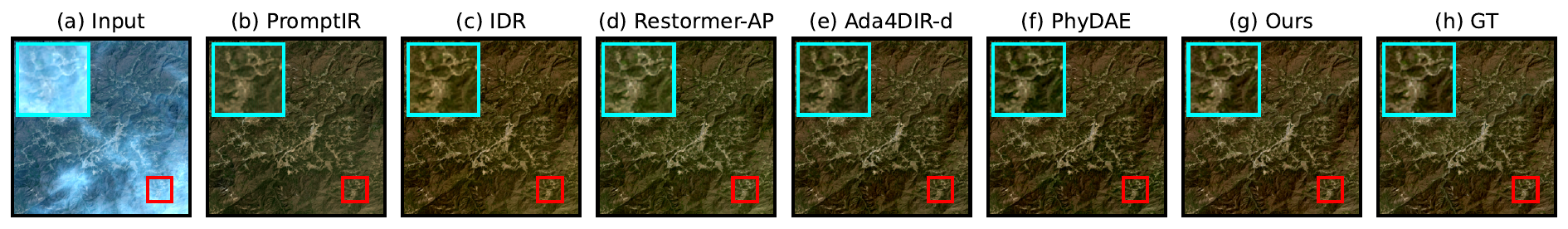}
    \includegraphics[width=\linewidth]{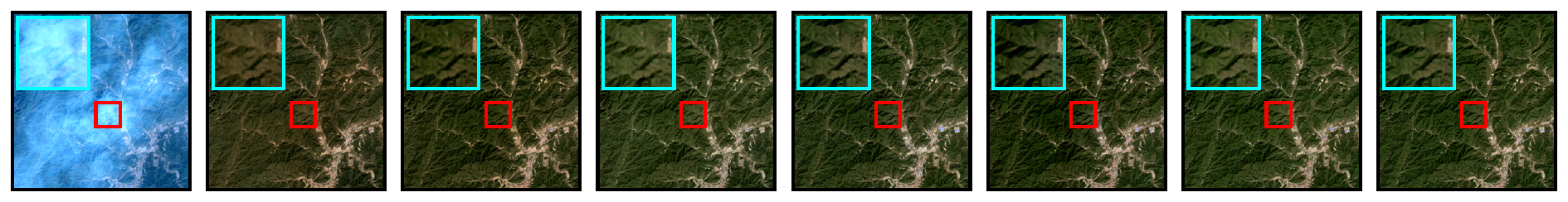}
  \end{minipage}

  \begin{minipage}[c]{0.015\linewidth}
    \rotatebox{90}{Noise}
  \end{minipage}
  \begin{minipage}[c]{0.95\linewidth}
    \includegraphics[width=\linewidth]{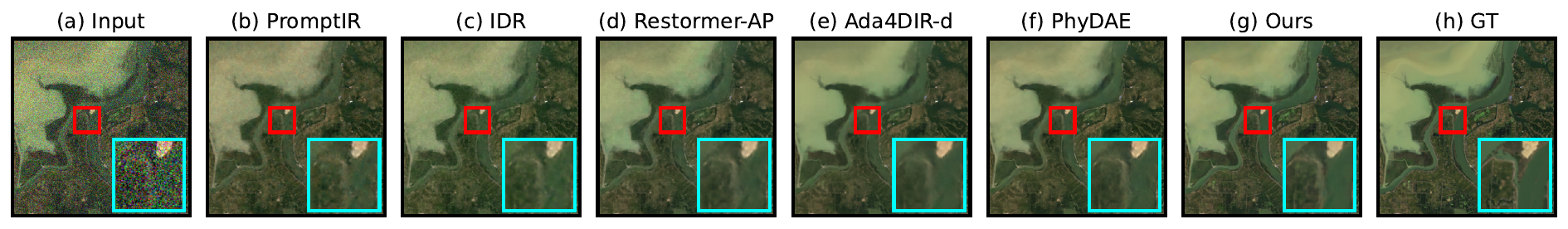}
    \includegraphics[width=\linewidth]{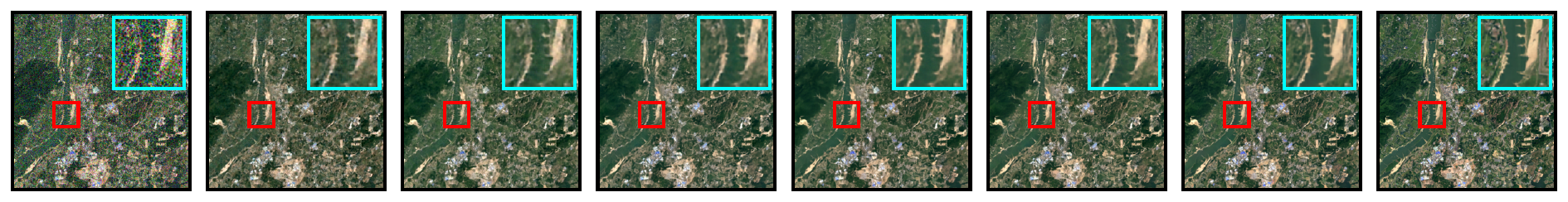}
  \end{minipage}

  \caption{Qualitative comparison on MDRS-Landsat. The panel illustrates that CoRE-UIR preserves large-scale structures while suppressing domain-shifted degradation artifacts more reliably than competing universal restoration methods.}
  \label{fig:visual_generalization}
  \vspace{-1em}
\end{figure*}

\subsubsection{Implementation Details}
\label{subsubsec:implementation_details}

The NAFNet backbone is configured with encoder stages of $[1, 1, 1, 28]$ NAFBlocks and a base channel width of $C = 32$. The degradation prior is extracted by a frozen CLIP ViT-B/32 ($224 \times 224$ input) image encoder and mapped to a degradation embedding of dimension $d_z = 384$ via a lightweight two-layer adapter ($d_\text{hidden} = 384$). For CoRE, we set the number of low-rank expert pairs to $N = 6$, matching the $M = 6$ base degradation types, the bottleneck rank to $r = 4$, and $k = 3$. The GFM uses a bottleneck reduction ratio of $\rho = 16$. The code and dataset will be released at \url{https://github.com/zzaiyan/CoRE-UIR}.

The framework is trained in two phases. In Phase I (Degradation Prior Adaptation), DPE is optimized with a multi-label degradation classification task over the $M=6$ base degradation types by freezing the CLIP encoder and updating only the lightweight adapter together with a linear classification head. We use the binary cross-entropy loss, the AdamW optimizer \citep{loshchilov2017decoupled} ($\beta_1=0.9$, $\beta_2=0.9$, and weight decay $10^{-3}$), a batch size of 16, and an initial learning rate of $2 \times 10^{-4}$ decayed to $1 \times 10^{-5}$ by cosine annealing. Phase I is trained for 30 epochs on both single and compound training samples, and the latest checkpoint is used as the DPE checkpoint for Phase II. For DPE, the local view is randomly cropped during training and replaced by a fixed center crop at inference. In Phase II (Restoration Network Training), the classification head is discarded, the entire DPE is frozen, and the instantiated restoration backbone (GLA-Net) is trained for 700,000 iters with the AdamW optimizer using the same settings, a batch size of 8, and cosine annealing \citep{loshchilov2016sgdr} that decays the learning rate from $1 \times 10^{-3}$ to $1 \times 10^{-6}$. During this phase, we jointly optimize a pixel loss (implemented as L1), a structural loss ($1-\text{SSIM}$), and a perceptual loss (implemented by LPIPS\footnote{https://github.com/richzhang/PerceptualSimilarity}), with weights 0.6, 0.2, and 0.2, respectively, in Eq.~\ref{eq:total_loss}. Training images are randomly cropped to $256 \times 256$ patches. Data augmentation includes random horizontal/vertical flips and rotations. For CoRE-UIR, the latest Phase-II checkpoint is used for evaluation. Baseline methods follow the authors' official training and inference settings. All experiments are conducted on a single NVIDIA RTX 4090 24GB GPU using PyTorch 2.5 framework.

\subsection{Comparison with State-of-the-Art}
\label{sec:comparison}

Fig.~\ref{fig:radar_psnr} first provides a compact PSNR overview of representative algorithms on MDVD-108K single degradations, MDVD-108K compound degradations, and MDRS-Landsat. The three radar panels show that CoRE-UIR forms the strongest overall envelope on the two MDVD-108K settings and preserves the best average with a competitive outer frontier on MDRS-Landsat, which is consistent with the detailed quantitative comparisons below.

\begin{table*}[t]
  \caption{Satellite-domain evaluation on the \textbf{MDRS-Landsat} dataset. Best results are highlighted in \textbf{bold} and second-best are \underline{underlined}. $\uparrow$ indicates higher is better and $\downarrow$ indicates lower is better. $^{*}$ denotes remote-sensing-specific methods.}
  \label{tab:cross_domain}
  \renewcommand{\arraystretch}{1.2}
  \setlength{\tabcolsep}{0.25em}
  \centering
  \resizebox{\linewidth}{!}{%
    \begin{tabular}{ll|ccc|ccc|ccc|ccc|ccc}
      \toprule
      \multirow{2}{*}{\textbf{Method}}     & \multirow{2}{*}{\textbf{Venue}}
                                           & \multicolumn{3}{c|}{\textbf{Blur}}
                                           & \multicolumn{3}{c|}{\textbf{Dark}}
                                           & \multicolumn{3}{c|}{\textbf{Haze}}
                                           & \multicolumn{3}{c|}{\textbf{Noise}}
                                           & \multicolumn{3}{c}{\textbf{Average}}                                                                                                                                                                                                                                                                                                                    \\
                                           &                                      & PSNR$\uparrow$    & SSIM$\uparrow$     & LPIPS$\downarrow$
                                           & PSNR$\uparrow$                       & SSIM$\uparrow$    & LPIPS$\downarrow$
                                           & PSNR$\uparrow$                       & SSIM$\uparrow$    & LPIPS$\downarrow$
                                           & PSNR$\uparrow$                       & SSIM$\uparrow$    & LPIPS$\downarrow$
                                           & PSNR$\uparrow$                       & SSIM$\uparrow$    & LPIPS$\downarrow$                                                                                                                                                                                                                                                                            \\
      \midrule
      \rowcolor{gray!20} \multicolumn{17}{l}{\textit{Single-task methods}}                                                                                                                                                                                                                                                                                                                           \\
      \addlinespace[0.2em]
      MPRNet                               & CVPR'21                              & 32.54             & 0.8404             & 0.3514             & 31.52             & 0.9666             & 0.0523          & 29.15             & 0.9673             & 0.0575             & 33.02             & 0.8271             & 0.2515             & 31.56             & 0.9003             & 0.1782             \\
      Restormer                            & CVPR'22                              & 35.23             & 0.8559             & 0.2099             & 37.86             & 0.9872             & 0.0110          & 36.18             & 0.9867             & 0.0124             & 34.53             & 0.8589             & 0.1356             & 35.95             & 0.9222             & 0.0922             \\
      NAFNet                               & ECCV'22                              & 33.10             & 0.8120             & 0.3194             & 30.40             & 0.9516             & 0.0542          & 31.56             & 0.9642             & 0.0417             & 33.08             & 0.8263             & 0.1872             & 32.03             & 0.8885             & 0.1506             \\
      DGUNet                               & CVPR'22                              & 29.64             & 0.7822             & 0.3405             & 27.15             & 0.9010             & 0.1339          & 27.45             & 0.9338             & 0.0713             & 30.31             & 0.7314             & 0.2491             & 28.64             & 0.8371             & 0.1987             \\
      \midrule
      \rowcolor{gray!20} \multicolumn{17}{l}{\textit{All-in-One methods}}                                                                                                                                                                                                                                                                                                                            \\
      \addlinespace[0.2em]
      TransWeather                         & CVPR'22                              & 33.45             & 0.8159             & 0.2868             & 36.33             & 0.9705             & 0.0193          & 35.02             & 0.9689             & 0.0241             & 33.69             & 0.8428             & 0.1530             & 34.62             & 0.8995             & 0.1208             \\
      AirNet                               & CVPR'22                              & 28.27             & 0.7887             & 0.3244             & 28.38             & 0.9472             & 0.0569          & 24.39             & 0.9331             & 0.0641             & 30.30             & 0.7446             & 0.1918             & 27.84             & 0.8534             & 0.1593             \\
      PromptIR                             & NeurIPS'23                           & 36.41             & 0.8861             & 0.1557             & 39.09             & 0.9900             & 0.0084          & 37.61             & 0.9897             & 0.0084             & \underline{34.99} & \underline{0.8729} & 0.1029             & 37.02             & 0.9347             & 0.0689             \\
      IDR                                  & CVPR'23                              & 36.57             & 0.8902             & 0.1498             & 35.19             & 0.9865             & 0.0096          & 36.99             & 0.9892             & 0.0087             & 34.88             & 0.8681             & 0.1091             & 35.91             & 0.9335             & 0.0693             \\
      Restormer-AP                         & CVPR'24                              & 35.75             & 0.8732             & 0.1800             & 37.27             & 0.9885             & 0.0102          & 37.36             & 0.9888             & 0.0098             & 34.96             & 0.8697             & 0.1239             & 36.34             & 0.9301             & 0.0810             \\
      Ada4DIR-d$^{*}$              & INFFUS'25                            & \textbf{37.20}    & \underline{0.9004} & \underline{0.1308} & \underline{43.85} & \underline{0.9954} & \textbf{0.0023} & \underline{41.06} & \underline{0.9938} & \textbf{0.0038}    & \textbf{35.14}    & 0.8724             & 0.1268             & 39.31             & \underline{0.9405} & 0.0659             \\
      PhyDAE$^{*}$                 & TGRS'26                              & 36.88             & 0.8824             & 0.1487             & 42.24             & 0.9949             & 0.0121          & 39.12             & 0.9928             & 0.0227             & 34.53             & 0.8651             & \underline{0.0991} & \underline{39.62} & 0.9390             & \underline{0.0600} \\
      \midrule
      \rowcolor{cyan!10} \textbf{CoRE-UIR} & \textit{Ours}                        & \underline{37.12} & \textbf{0.9039}    & \textbf{0.0517}    & \textbf{46.51}    & \textbf{0.9964}    & \textbf{0.0023} & \textbf{42.42}    & \textbf{0.9947}    & \underline{0.0059} & 34.84             & \textbf{0.8754}    & \textbf{0.0910}    & \textbf{40.22}    & \textbf{0.9426}    & \textbf{0.0377}    \\
      \bottomrule
    \end{tabular}%
  }
  \vspace{-1em}
\end{table*}

\subsubsection{Single Degradation}

Tables~\ref{tab:weather_degradation} and~\ref{tab:imaging_degradation} show that CoRE-UIR attains the best average performance across both weather-induced and imaging-induced degradations, and ranks first across all six single-degradation subsets. Relative to BaryIR, the strongest recent universal restoration baseline by average score, CoRE-UIR gains 1.28~dB on the weather average and 0.80~dB on the imaging average, while also improving SSIM and LPIPS in both groups. Other recent universal restoration models remain competitive on individual cases: DACLIP-UIR is strong on weather-related perceptual quality, and MoCE-IR narrows the gap on several imaging subsets, but CoRE-UIR is more consistent across both groups. The visual comparisons are broadly aligned with these margins: CoRE-UIR recovers clearer distant facades in fog, removes the yellow cast more thoroughly in dust, and preserves bridge, pedestrian, shop-sign, vehicle, and roadside boundaries more cleanly in rain, low-light, motion-blur, and defocus scenes.

\subsubsection{Compound Degradation}

Table~\ref{tab:compound_degradation} shows that the advantage is retained under overlapping degradations. CoRE-UIR achieves the best result on all three representative combinations and improves the compound average by 1.15~dB over BaryIR and by 1.74~dB over MoCE-IR, while also delivering the best SSIM and LPIPS. Although visual differences can be subtle in some cases, the gains are consistent across PSNR, SSIM, and LPIPS in this harder compound-degradation setting. In the visual examples, CoRE-UIR restores clearer bus contours and sharper sign edges in fog+motion, and better balances streak suppression with dark-region recovery in rain+low-light. Competing methods remain visually reasonable, but they tend to leave slightly more residual artifacts or softer boundaries in these examples.

\subsubsection{Real-World Degradation}

We further examine the real-world degradation test set of MDVD-108K, where ground-truth references are unavailable. The qualitative comparisons in Fig.~\ref{fig:visual_real} suggest that CoRE-UIR generalizes stably beyond the synthetic pipeline: it recovers building windows and track boundaries more clearly in fog, preserves vehicle contours and highlight structure more faithfully under night motion blur, reveals dark-region objects without obvious over-brightening in low light, and restores character strokes and roof boundaries more cleanly in defocus scenes. Several competing baselines also produce appealing results on individual images, but CoRE-UIR appears comparatively balanced between detail recovery and artifact suppression across the four scenes.

\subsubsection{Satellite-Domain Evaluation}

We further use MDRS-Landsat for satellite-domain evaluation, where the comparison is stronger because Ada4DIR-d and PhyDAE are both tailored to remote sensing restoration with physical or domain-specific priors. Ada4DIR-d remains highly competitive and obtains the best blur PSNR (37.20) and noise PSNR (35.14), while PhyDAE provides the strongest non-ours overall average (39.62) together with low LPIPS. CoRE-UIR nevertheless achieves the best overall average of 40.22 and the best overall SSIM/LPIPS of 0.9426/0.0377, mainly due to clear gains on dark restoration (46.51 vs. 43.85) and haze removal (42.42 vs. 41.06). The qualitative comparisons show a similar tendency: CoRE-UIR preserves coastlines and sandbars more cleanly in blur, lifts dark regions with less loss of land-cover contrast, removes the haze veil while keeping ridge textures, and maintains cleaner shoreline boundaries under noise. These results suggest that the prior-guided global-local design adapts well to the satellite benchmark, demonstrating its versatility across different remote sensing scenarios.

\begin{table*}[t!]
  \centering
  \caption{Ablation study on \textbf{Degradation Prior Embedding (DPE)}. The table is organized into three groups: (a) training strategy, (b) prior encoder, and (c) input view type. Complexity is counted on the entire DPE branch, including the encoder, adapter, and classification head. “--” means the metric is not applicable for that variant.}
  \label{tab:ablation_dpe}
  \resizebox{\linewidth}{!}{%
    \begin{tabular}{l|ccc|ccc|ccc}
      \toprule
      \multirow{2}{*}{\textbf{Variant}} & \multicolumn{3}{c|}{\textbf{Classification}} & \multicolumn{3}{c|}{\textbf{Restoration}} & \multicolumn{3}{c}{\textbf{Complexity}}                                                                                                         \\
                                        & Pre. (\%)$\uparrow$                          & Rec. (\%)$\uparrow$                       & F1 (\%)$\uparrow$                       & PSNR$\uparrow$ & SSIM$\uparrow$  & LPIPS$\downarrow$ & Total Params   & Trainable     & FLOPs         \\
      \midrule
      \rowcolor{gray!10} \multicolumn{10}{l}{\textit{(a) Training strategy}}                                                                                                                                                                                                         \\
      \addlinespace[0.2em]
      Joint training (no Phase-I)       & --                                           & --                                        & --                                      & 29.87          & 0.8804          & 0.0647            & 88.57M         & 0.72M         & 8.83G         \\
      Two-phase training (ours)         & \textbf{99.4}                                & 99.2                                      & \textbf{99.3}                           & \textbf{30.67} & \textbf{0.8928} & \textbf{0.0503}   & 88.57M         & 0.72M         & 8.83G         \\
      \midrule
      \rowcolor{gray!10} \multicolumn{10}{l}{\textit{(b) Prior encoder}}                                                                                                                                                                                                             \\
      \addlinespace[0.2em]
      ResNet-50 (trainable)             & 98.1                                         & 98.1                                      & 98.1                                    & 29.51          & 0.8824          & 0.0612            & 26.95M         & 26.95M        & 8.22G         \\
      DACLIP (w/ adapter)               & 98.9                                         & 99.1                                      & 99.0                                    & 30.49          & 0.8911          & 0.0536            & 183.87M        & 1.09M         & 9.09G         \\
      RemoteCLIP (w/ adapter)   & 98.7                                 & 99.0                              & 98.8                            & 30.44  & 0.8913  & 0.0546    & 88.57M & 0.72M & 8.83G \\
      CLIP (w/ adapter, ours)           & \textbf{99.4}                                & \textbf{99.2}                             & \textbf{99.3}                           & \textbf{30.67} & \textbf{0.8928} & \textbf{0.0503}   & 88.57M         & 0.72M         & 8.83G         \\
      \midrule
      \rowcolor{gray!10} \multicolumn{10}{l}{\textit{(c) Input view type}}                                                                                                                                                                                                           \\
      \addlinespace[0.2em]
      Global view only                  & 99.3                                         & 98.8                                      & 99.0                                    & 30.45          & 0.8894          & 0.0533            & 88.31M         & 0.46M         & 4.41G         \\
      Local view only                   & 96.9                                         & 94.0                                      & 95.4                                    & 30.11          & 0.8806          & 0.0635            & 88.31M         & 0.46M         & 4.41G         \\
      Global + local views (ours)       & \textbf{99.4}                                & \textbf{99.2}                             & \textbf{99.3}                           & \textbf{30.67} & \textbf{0.8928} & \textbf{0.0503}   & 88.57M         & 0.72M         & 8.83G         \\
      \bottomrule
    \end{tabular}%
  }
  \vspace{-1em}
\end{table*}

\begin{figure*}[t!]
  \centering
  \includegraphics[width=\linewidth]{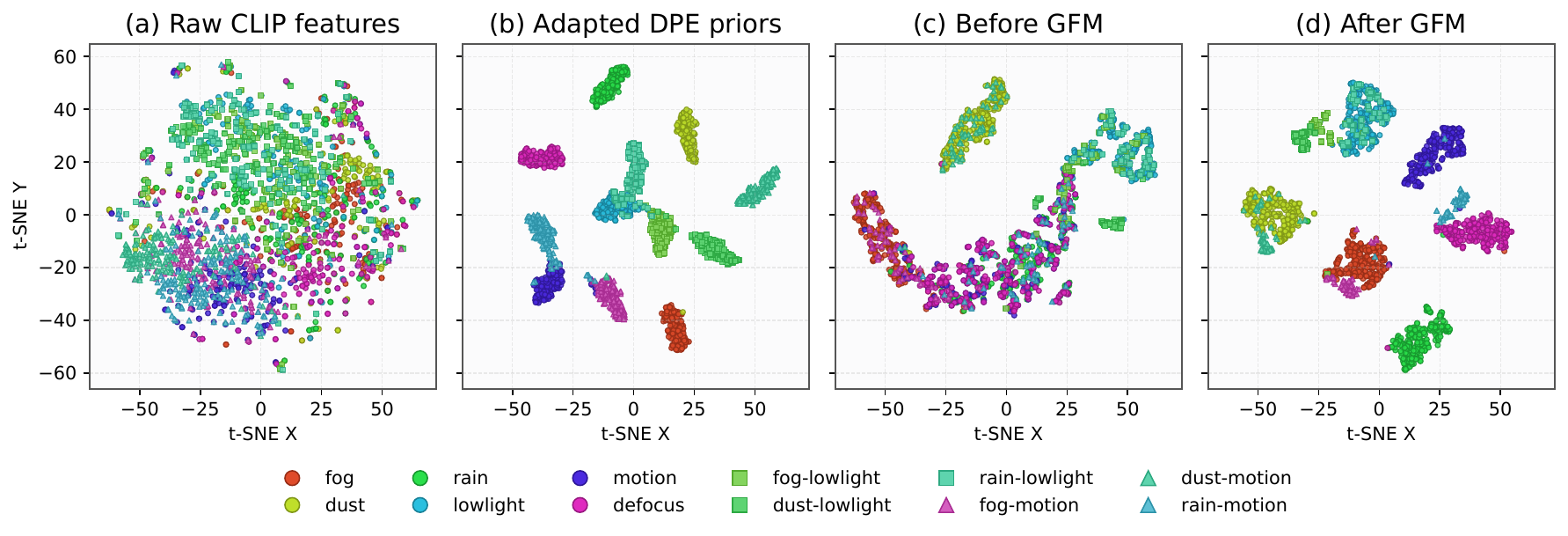}
  \caption{Feature visualization for \textbf{DPE} and \textbf{GFM}, showing (a) Raw CLIP features, (b) Adapted DPE priors, (c) Features before GFM, and (d) Features after GFM. (a)--(b) demonstrate that DPE improves the degradation discriminativeness of the raw CLIP feature, while (c)--(d) show that GFM reshapes the restoration feature space into a more compact and routable manifold.}
  \label{fig:dpe_gfm_tsne}
  \vspace{-1em}
\end{figure*}

\subsection{Ablation Studies}
\label{sec:ablation}

We conduct ablation experiments on MDVD-108K to evaluate DPE, GFM, and CoRE. Quantitative restoration results are reported by averaging the metrics over all single and compound degradation samples. Restoration and efficiency measurements use $512 \times 512$ inputs. Unless otherwise specified, Table~\ref{tab:ablation_dpe} reports only DPE-branch complexity, while Tables~\ref{tab:ablation_gfm_branches} and~\ref{tab:ablation_core} report only GLA-Net complexity. End-to-end efficiency is analyzed separately in Section~\ref{sec:further_analysis}.

\subsubsection{Degradation Prior Embedding}

We first examine whether DPE provides priors that are both discriminative and useful for restoration. Table~\ref{tab:ablation_dpe} studies this question from three aspects: training strategy, prior encoder, and input view type. For the Phase-I proxy task, we additionally report macro Precision, Recall, and F1.

\textbf{Training strategy.} Comparing the first two rows shows that the proposed two-phase optimization improves restoration under identical DPE complexity, improving PSNR by 0.80~dB, raising SSIM from 0.8804 to 0.8928, and lowering LPIPS from 0.0647 to 0.0503. This indicates that Phase I is not merely an auxiliary objective, but reshapes the degradation embedding into a more restoration-oriented prior before the backbone is trained. Fig.~\ref{fig:dpe_gfm_tsne}(a)--(b) provides a direct visualization of this effect: the raw frozen CLIP features are heavily entangled across single and compound degradations, whereas the adapted DPE priors form much tighter and cleaner clusters with larger inter-cluster margins\footnote{The t-SNE projection uses 2,000 randomly sampled features with perplexity 30 and seed 42.}.

\textbf{Prior encoder.} Table~\ref{tab:ablation_dpe}(b) shows a clear hierarchy among four representative prior encoders: the fully trainable ResNet-50 \citep{ResNet} is weakest, DACLIP \citep{luo2023daclip} and RemoteCLIP \citep{remoteclip} bring clear gains as degradation-aware and remote-sensing priors, and the proposed CLIP with adapter tuning achieves the best classification and restoration results. Notably, although the trainable ResNet-50 already attains respectable classification performance, the restoration model guided by its priors still trails the CLIP-based design by 1.16~dB PSNR. This indicates that the gain comes not simply from fitting the proxy classification task or enlarging the trainable encoder, but from adapting foundation priors efficiently, with the general CLIP prior performing best with only 0.72M trainable parameters. The feature distribution in Fig.~\ref{fig:dpe_gfm_tsne}(b) is also more structured than the raw CLIP space in Fig.~\ref{fig:dpe_gfm_tsne}(a), which is consistent with the quantitative improvement.

\textbf{Input view type.} Using only the global or local view weakens both classification and restoration, with the local-only variant degrading the most. The combined global-local representation achieves the best overall result, confirming that broad scene context and local degradation textures are complementary for prior learning.

\subsubsection{Global Feature Modulation (GFM)}

We next examine whether GFM indeed performs effective prior-state global feature modulation before CoRE specializes local residuals. Table~\ref{tab:ablation_gfm_branches} compares removing GFM entirely, simplifying it to a single branch, and using the full dual-branch design while keeping the rest of the pipeline unchanged. Removing GFM causes the largest performance drop, reducing PSNR from 30.67 to 28.87. Using only the prior branch or only the state branch recovers part of the gain, but both remain inferior to the full module, indicating that effective modulation requires both explicit degradation conditioning and the current feature-state response.

Fig.~\ref{fig:dpe_gfm_tsne}(c)--(d) visualizes pooled image features before and after the first GFM module. Before GFM, the restoration features largely lie on an elongated and partially entangled manifold, with several degradation types still overlapping around the central transition region. After modulation, the features reorganize into tighter and more separated groups, while related degradations still preserve meaningful neighborhood relations rather than collapsing into overly rigid class partitions. This behavior matches the role of GFM: it does not simply classify degradations, but reshapes intermediate features into a more compact, prior-consistent, and more routable space for downstream CoRE specialization.

\begin{table}[t!]
  \centering
  \caption{Ablation study on \textbf{GFM}. The table compares removing GFM, single-branch simplifications, and the full dual-branch module.}
  \label{tab:ablation_gfm_branches}
  \resizebox{\linewidth}{!}{%
    \setlength{\tabcolsep}{0.8em}%
    \begin{tabular}{l|cccc}
      \toprule
      Configuration     & Params          & FLOPs           & PSNR$\uparrow$ & SSIM$\uparrow$  \\
      \midrule
      \addlinespace[0.2em]
      No GFM            & 21.06M          & 76.90G          & 28.87          & 0.8729          \\
      Prior branch only & 21.69M          & 76.90G          & 30.54          & 0.8914          \\
      State branch only & 21.56M          & 76.93G          & 30.27          & 0.8896          \\
      Full GFM (Ours)   & \textbf{22.12M} & \textbf{76.93G} & \textbf{30.67} & \textbf{0.8928} \\
      \bottomrule
    \end{tabular}%
  }
  \vspace{-1em}
\end{table}

\begin{table}[ht]
  \centering
  \caption{Ablation study on \textbf{CoRE}. The table is organized into four groups: (a) common-residual expert design, (b) expert sparsity, (c) routing granularity, and (d) routing method.}
  \label{tab:ablation_core}
  \resizebox{\linewidth}{!}{%
    \begin{tabular}{l|cccc}
      \toprule
      Configuration              & Params & FLOPs   & PSNR$\uparrow$ & SSIM$\uparrow$   \\
      \midrule
      \rowcolor{gray!10} \multicolumn{5}{l}{\textit{(a) Common-residual expert design}} \\
      \addlinespace[0.2em]
      Common expert only         & 18.47M & 63.91G  & 30.36          & 0.8895           \\
      Full-rank FFN MoE          & 51.71M & 121.89G & 30.51          & 0.8918           \\
      CoRE (ours)                & 22.12M & 76.93G  & \textbf{30.67} & \textbf{0.8928}  \\
      \midrule
      \rowcolor{gray!10} \multicolumn{5}{l}{\textit{(b) Expert sparsity}}               \\
      \addlinespace[0.2em]
      $k=1$, $N=6$               & 22.12M & 68.25G  & 30.40          & 0.8901           \\
      $k=3$, $N=6$ (ours)        & 22.12M & 76.93G  & \textbf{30.67} & \textbf{0.8928}  \\
      $k=6$, $N=6$               & 22.12M & 89.96G  & 30.58          & 0.8924           \\
      \midrule
      \rowcolor{gray!10} \multicolumn{5}{l}{\textit{(c) Routing granularity}}           \\
      \addlinespace[0.2em]
      sample-level routing       & 22.11M & 76.93G  & 30.52          & 0.8913           \\
      stage-level routing (ours) & 22.12M & 76.93G  & \textbf{30.67} & \textbf{0.8928}  \\
      block-level routing        & 22.19M & 76.93G  & 30.47          & 0.8887           \\
      \midrule
      \rowcolor{gray!10} \multicolumn{5}{l}{\textit{(d) Routing method}}                \\
      \addlinespace[0.2em]
      MLP router                 & 23.45M & 76.94G  & 30.50          & 0.8916           \\
      PG-Router (ours)           & 22.12M & 76.93G  & \textbf{30.67} & \textbf{0.8928}  \\
      \bottomrule
    \end{tabular}%
  }
  \vspace{-1em}
\end{table}

\begin{figure*}[t!]
  \centering
  \begin{minipage}[c]{0.015\linewidth}
    \rotatebox{90}{Fog}
  \end{minipage}
  \begin{minipage}[c]{0.95\linewidth}
    \includegraphics[width=\linewidth]{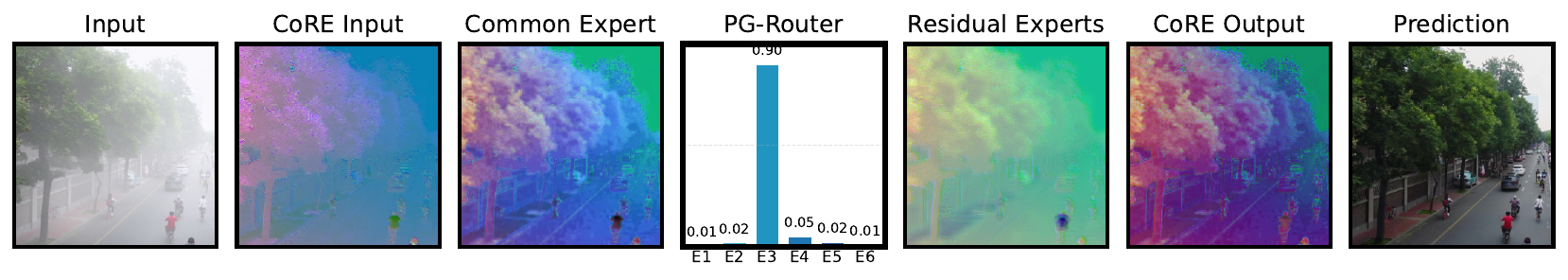}
  \end{minipage}

  \begin{minipage}[c]{0.015\linewidth}
    \rotatebox{90}{Rain}
  \end{minipage}
  \begin{minipage}[c]{0.95\linewidth}
    \includegraphics[width=\linewidth]{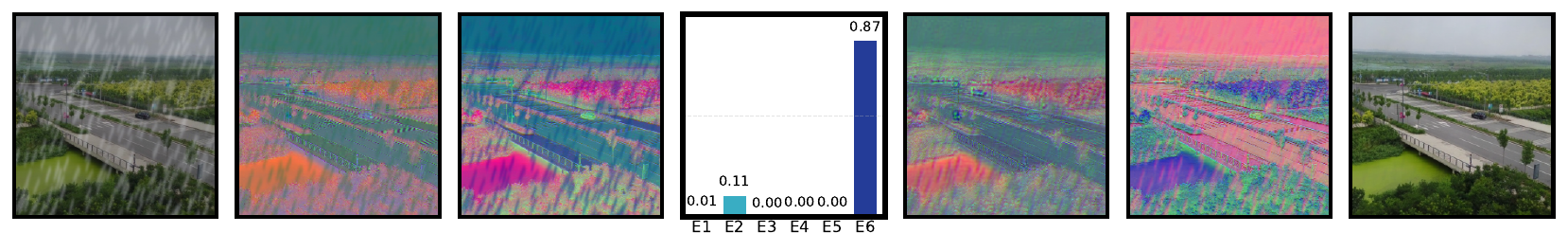}
  \end{minipage}

  \begin{minipage}[c]{0.015\linewidth}
    \rotatebox{90}{Dust+Lowlight}
  \end{minipage}
  \begin{minipage}[c]{0.95\linewidth}
    \includegraphics[width=\linewidth]{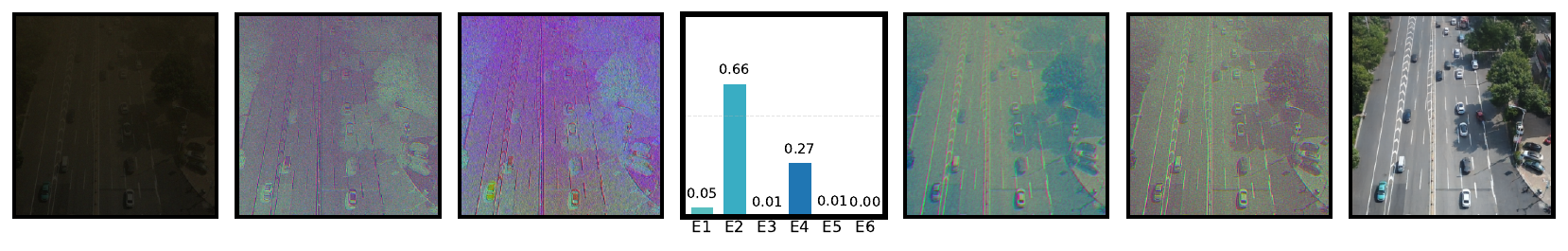}
  \end{minipage}

  \caption{Feature visualization for the first \textbf{CoRE} module at stage 2. From left to right, we show Input, CoRE Input Feature, Common Expert Feature, Router Weights, Residual Experts Feature, CoRE Output Feature, and Prediction. The common expert exhibits similar responses across degradations and builds a shared restoration basis, while the residual experts focus on degradation-specific compensation such as fog depth contrast, rain background brightening, and distant denoising in dust-dominated regions.}
  \label{fig:core_visualization}
  \vspace{-1em}
\end{figure*}

\subsubsection{Common-and-Residual Expert Block (CoRE)}

Finally, we examine whether CoRE provides efficient degradation-specific residual specialization. We study the common-residual expert design, expert sparsity, and routing strategy.

\textbf{Common-residual expert design.} Table~\ref{tab:ablation_core} consolidates the quantitative study of CoRE into four groups, where the first two focus on expert design and the last two focus on routing. Table~\ref{tab:ablation_core}(a) compares the common dense expert only variant, a full-rank FFN MoE branch, and the proposed CoRE branch to validate the common-residual expert design. The common dense expert only variant already reaches 30.36 PSNR and 0.8895 SSIM, indicating that a large portion of restoration behavior is shared across degradations. The full-rank MoE slightly improves adaptive capacity but expands the model to 51.71M parameters and 121.89G FLOPs. CoRE further improves the result to 30.67/0.8928 while remaining much lighter, validating the proposed common dense plus low-rank residual decomposition.

\textbf{Expert sparsity.} Table~\ref{tab:ablation_core}(b) fixes the expert-pair number to $N=6$, matching the $M=6$ base degradation types, and studies only the effect of $k$. When $k=1$, routing is overly restrictive and cannot combine complementary experts for compound degradations. When $k=6$, the combination becomes nearly dense, increases FLOPs from 76.93G to 89.96G, and yields no further gain. $k=3$ therefore provides the most balanced trade-off between flexibility and sparsity.

\textbf{Common dense \textit{vs.} low-rank residual experts.} Fig.~\ref{fig:core_visualization} visualizes the first CoRE module in Block 2. Across fog, rain, and dust+low-light scenes, the common dense expert exhibits highly similar responses and mainly strengthens the dominant scene layout, large structural boundaries, and overall contrast, thereby providing a stable restoration basis. The low-rank residual experts are more localized and degradation-dependent: in fog they emphasize the haze veil and distant depth transition, in rain they focus more strongly on streak-corrupted structures and background recovery, and in dust+low-light they concentrate on illumination lifting together with lane and vehicle details. The PG-Router activations are also sparse and input-dependent: different single degradations activate different dominant experts, whereas the compound degradation activates multiple experts jointly. Taken together, these responses visually support the intended common-residual decomposition and the sparse adaptive routing mechanism.

\textbf{Routing granularity and router type.} Table~\ref{tab:ablation_core}(c) and Table~\ref{tab:ablation_core}(d) show that stage-level routing achieves the best result and is therefore adopted in CoRE-UIR. The sample-level variant is more stable but too coarse to exploit local degradation diversity, whereas the block-level variant is more flexible but less stable to optimize. Stage-level routing therefore provides the best balance between adaptability and stability. The PG-Router also slightly improves accuracy while reducing parameters relative to the generic MLP alternative, so it is kept as the default router.

\subsection{Further Analysis}
\label{sec:further_analysis}

We further examine CoRE-UIR from four practical perspectives: model efficiency, prior quality, downstream perception utility, and robustness to unseen compound degradations. These analyses evaluate deployment cost, the effect of DPE prediction reliability, whether restoration benefits target perception, and whether the model remains stable under unseen degradation combinations.

\subsubsection{Model Efficiency}

\begin{figure}[t]
  \centering
  \includegraphics[width=\linewidth]{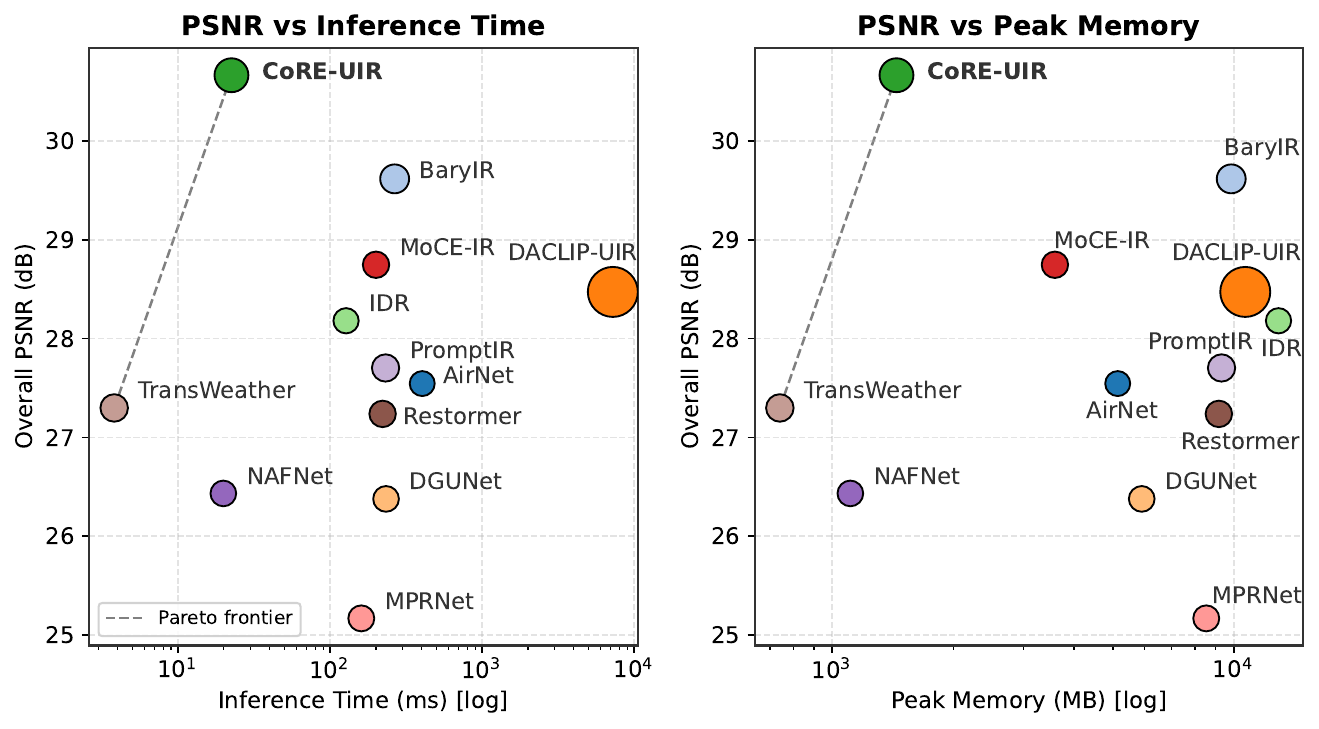}
  \caption{Quality-efficiency trade-off on MDVD-108K. Left: PSNR \textit{vs.} inference time. Right: PSNR \textit{vs.} peak memory. Both horizontal axes are logarithmic, and CoRE-UIR stays near the upper-left frontier in both views.}
  \label{fig:tradeoff}
  \vspace{-1em}
\end{figure}

Following the benchmark protocol, all efficiency measurements are collected on a single NVIDIA RTX 4090 GPU with batch size 4. DPE's CLIP encoder is run with 16-bit AMP, while other methods follow their official settings. We first warm up 5 batches, then time 20 consecutive batches, and finally divide the batch latency by 4 to report per-image mean and standard deviation. Table~\ref{tab:efficiency} summarizes complexity, measured cost, and restoration quality under this unified setting. Although CoRE-UIR includes a frozen CLIP ViT-B/32 prior encoder, its full inference path remains efficient: after including DPE, the total complexity is 85.76G FLOPs, with $22.44\pm1.06$ ms latency, 44.56 img/s throughput, and 1.45 GB peak memory. Relative to BaryIR, it still gains 1.05~dB PSNR, runs 11.83$\times$ faster, and uses 85.3\% less peak memory while preserving the best overall quality. By contrast, DACLIP-UIR is constrained by iterative sampling: one reverse step costs $72.00\pm0.33$ ms, but 100 denoising steps expand end-to-end latency to $7226.26\pm6.56$ ms. Fig.~\ref{fig:tradeoff} shows the same trend: TransWeather occupies the extreme low-latency corner at lower quality, whereas CoRE-UIR remains near the low-latency and low-memory side while exceeding 30 dB. Overall, CoRE-UIR preserves a favorable quality-efficiency trade-off for practical deployment.

\begin{table*}[t!]
  \centering
  \caption{Efficiency and restoration-quality comparison among \textbf{AiOIR methods} on MDVD-108K. Best results are highlighted in \textbf{bold} and second-best are \underline{underlined}. CoRE-UIR achieves the best restoration quality while remaining substantially efficient.}
  \label{tab:efficiency}
  \resizebox{\linewidth}{!}{%
    \setlength{\tabcolsep}{0.8em}%
    \begin{tabular}{ll|cc|ccc|ccc}
      \toprule
      \multirow{2}{*}{\textbf{Method}}     & \multirow{2}{*}{\textbf{Venue}} & \multicolumn{2}{c|}{\textbf{Complexity}} & \multicolumn{3}{c|}{\textbf{Measured Cost}} & \multicolumn{3}{c}{\textbf{Restoration}}                                                                                                               \\
                                           &                                 & Params$\downarrow$                       & FLOPs$\downarrow$                           & Latency (ms)$\downarrow$                 & Throughput$\uparrow$    & Memory$\downarrow$  & PSNR$\uparrow$    & SSIM$\uparrow$     & LPIPS$\downarrow$  \\
      \midrule
      TransWeather                         & CVPR'22                         & 38.1M                                    & \textbf{24.2G}                              & \textbf{3.80$\pm$0.99}                   & \textbf{263.04 img/s}   & \textbf{0.74 GB}    & 27.30             & 0.8409             & 0.1747             \\
      AirNet                               & CVPR'22                         & \textbf{8.93M}                           & 1174.8G                                     & 403.17$\pm$0.52                          & 2.48 img/s              & 5.14 GB             & 27.54             & 0.8670             & 0.0790             \\
      PromptIR                             & NeurIPS'23                      & 35.6M                                    & 690.9G                                      & 231.47$\pm$0.38                          & 4.32 img/s              & 9.32 GB             & 27.70             & 0.8665             & 0.0746             \\
      IDR                                  & CVPR'23                         & \underline{12.3M}                        & 363.1G                                      & 127.24$\pm$0.30                          & 7.86 img/s              & 12.92 GB            & 28.18             & 0.8823             & 0.1275             \\
      DACLIP-UIR                           & ICLR'24                         & 295.2M                                   & 56.5T                                       & 7226.26$\pm$6.56                         & 0.14 img/s              & 10.68 GB            & 28.47             & 0.8759             & 0.0658             \\
      MoCE-IR                              & CVPR'25                         & 25.4M                                    & 382.4G                                      & 200.33$\pm$0.93                          & 4.99 img/s              & 3.58 GB             & 28.75             & 0.8803             & 0.0642             \\
      BaryIR                               & TPAMI'26                        & 53.5M                                    & 851.1G                                      & 265.58$\pm$0.74                          & 3.77 img/s              & 9.85 GB             & \underline{29.62} & \underline{0.8853} & \underline{0.0566} \\
      \midrule
      \rowcolor{cyan!10} \textbf{CoRE-UIR} & \textit{Ours}                   & 110.7M                                   & \underline{85.76G}                          & \underline{22.44$\pm$1.06}               & \underline{44.56 img/s} & \underline{1.45 GB} & \textbf{30.67}    & \textbf{0.8928}    & \textbf{0.0503}    \\
      \bottomrule
    \end{tabular}%
  }
  \vspace{-1em}
\end{table*}

\subsubsection{DPE Prior Reliability}
\label{sec:prior_quality_analysis}

To examine how prior reliability affects restoration, we stratify the test samples according to the DPE multi-label prediction. Exact match means that the predicted degradation-label set is identical to the ground truth; partial match means that at least one degradation label is correct but the set is incomplete or contains extra labels; wrong means that no ground-truth degradation label is recovered. Table~\ref{tab:prior_quality_analysis} shows that 98.55\% of test samples fall into the exact-match bucket, where CoRE-UIR reaches 30.76~dB PSNR and 0.0493 LPIPS. The small partial and wrong buckets exhibit clearly lower restoration quality, especially in PSNR and LPIPS, indicating that inaccurate priors tend to coincide with difficult samples and residual artifacts. Nevertheless, these cases account for only 1.45\% of the test set, so the overall result remains close to the exact-match bucket.

\begin{table}[t!]
  \centering
  \caption{Prior-quality stratified restoration performance on MDVD-108K.}
  \label{tab:prior_quality_analysis}
  \resizebox{\linewidth}{!}{%
    \setlength{\tabcolsep}{0.55em}%
    \begin{tabular}{l|ccccc}
      \toprule
      DPE prediction & Samples & Ratio    & PSNR$\uparrow$ & SSIM$\uparrow$ & LPIPS$\downarrow$ \\
      \midrule
      Exact match    & 10,643  & 98.55\%  & 30.76          & 0.8939         & 0.0493            \\
      Partial match  & 143     & 1.32\%   & 25.17          & 0.8144         & 0.1066            \\
      Wrong          & 14      & 0.13\%   & 17.73          & 0.8262         & 0.1732            \\
      \midrule
      All            & 10,800  & 100.00\% & 30.67          & 0.8928         & 0.0503            \\
      \bottomrule
    \end{tabular}%
  }
\end{table}

\begin{table}[t!]
  \centering
  \caption{Downstream object detection with a fixed YOLO11s detector. Metrics are reported in percentage. Best and second-best restored-image results are highlighted in \textbf{bold} and \underline{underlined}.}
  \label{tab:downstream_detection}
  \resizebox{\linewidth}{!}{%
    \setlength{\tabcolsep}{0.72em}%
    \begin{tabular}{l|cccc}
      \toprule
      \textbf{Input}    & Precision$\uparrow$ & Recall$\uparrow$  & mAP$_{50}\uparrow$ & mAP$_{50:95}\uparrow$ \\
      \midrule
      \rowcolor{gray!20} \multicolumn{5}{l}{\textit{(a) Synthetic test set -- 10,800 samples}}                 \\
      \addlinespace[0.2em]
      Clean             & 72.40               & 52.81             & 57.40              & 31.49                 \\
      \cmidrule(lr){1-5}
      Degraded          & 54.47               & 28.27             & 28.38              & 15.01                 \\
      MoCE-IR           & 65.65               & 43.84             & 46.67              & 25.05                 \\
      BaryIR            & \underline{66.52}   & \underline{45.33} & \underline{48.66}  & \underline{25.65}     \\
      \textbf{CoRE-UIR} & \textbf{67.84}      & \textbf{47.08}    & \textbf{50.36}     & \textbf{27.10}        \\
      \midrule
      \rowcolor{gray!20} \multicolumn{5}{l}{\textit{(b) Real test set -- 500 samples}}                         \\
      \addlinespace[0.2em]
      Real degraded     & 55.06               & 35.91             & 37.16              & 20.65                 \\
      MoCE-IR           & 55.29               & 37.64             & 38.70              & 22.30                 \\
      BaryIR            & \underline{58.70}   & \underline{39.15} & \underline{40.86}  & \underline{23.89}     \\
      \textbf{CoRE-UIR} & \textbf{59.12}      & \textbf{40.73}    & \textbf{41.73}     & \textbf{25.20}        \\
      \bottomrule
    \end{tabular}%
  }
  \vspace{-1em}
\end{table}

\subsubsection{Downstream Object Detection}
\label{sec:downstream_detection}

\begin{figure}[t!]
  \centering
  \begin{minipage}[c]{0.035\linewidth}
    \rotatebox{90}{Motion Blur}
  \end{minipage}
  \begin{minipage}[c]{0.94\linewidth}
    \includegraphics[width=\linewidth]{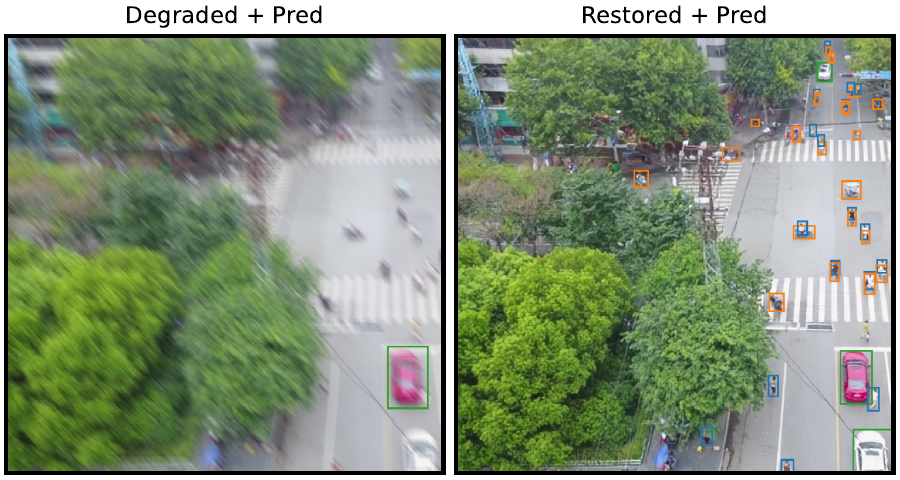}
  \end{minipage}

  \vspace{0.3em}
  \begin{minipage}[c]{0.035\linewidth}
    \rotatebox{90}{Real Low-light}
  \end{minipage}
  \begin{minipage}[c]{0.94\linewidth}
    \includegraphics[width=\linewidth]{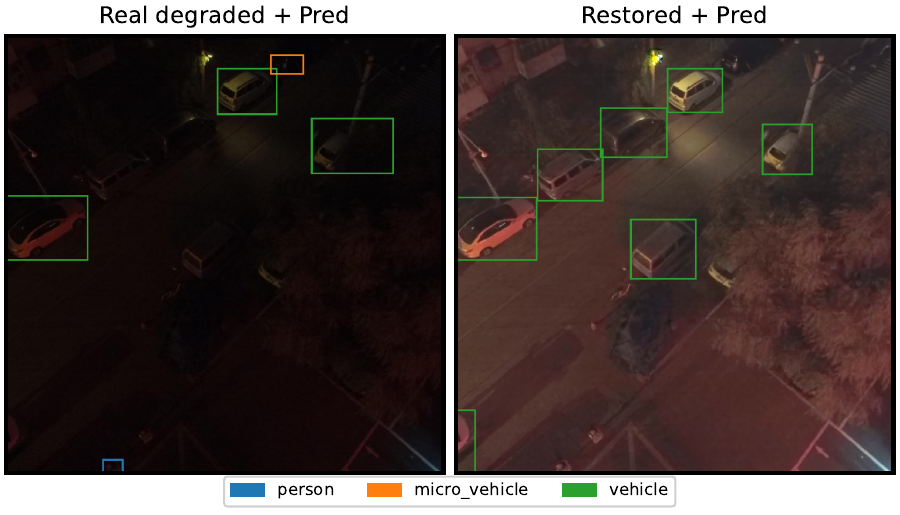}
  \end{minipage}
  \caption{Qualitative downstream detection examples on synthetic motion blur and real low-light scenes. Compared with degraded inputs, CoRE-UIR restores clearer object boundaries and illumination, yielding more complete detector responses.}
  \label{fig:downstream_detection_visual}
  \vspace{-1em}
\end{figure}

To evaluate whether restoration benefits downstream perception, we train a YOLO11s \citep{khanam2024yolov11} detector for 50 epochs\footnote{We use the official implementation of YOLO11s by Ultralytics at \url{https://github.com/ultralytics/ultralytics} and keep all default training and inference settings.} using only clean GT images from the MDVD-108K training split. The detector weights are then fixed and evaluated on clean GT, degraded input, and restored images from both the paired synthetic test set and the real degraded test set with detection labels. As shown in Table~\ref{tab:downstream_detection}, degraded inputs sharply weaken detection, whereas restored images recover a large portion of the lost accuracy. CoRE-UIR gives the best restored-image results on the synthetic test set, recovering 75.7\% and 73.4\% of the degradation-induced gaps in mAP$_{50}$ and mAP$_{50:95}$. The same trend holds on the real test set without clean GT images, where CoRE-UIR also outperforms BaryIR and the degraded input. Fig.~\ref{fig:downstream_detection_visual} further shows that restoration sharpens motion-blurred object boundaries and brightens real low-light scenes, leading to more complete detections of vehicles, micro-vehicles, and pedestrians. These results indicate that CoRE-UIR preserves object-level cues useful for a detector that is never trained on restored images.

\subsubsection{Unseen Compound-Degradation Robustness}
\label{sec:unseen_compound_robustness}

We further evaluate whether the learned restoration behavior transfers to unseen compound degradations. Here, seen compound types refer to the six compound degradations defined in the dataset, while unseen ones are previously unused combinations of a weather degradation with defocus blur. Table~\ref{tab:unseen_compound_robustness} compares three representative universal restoration baselines under this setting. Although unseen cases are more challenging for all methods, CoRE-UIR remains best across the three metrics. Relative to BaryIR, it gains 0.73~dB PSNR and also preserves clearer structural and perceptual quality. These results suggest that the common-residual expert design does not merely memorize observed compound pairs, but preserves useful degradation-adaptive behavior for new combinations of known degradation factors.

\begin{table}[t!]
  \centering
  \caption{Robustness to unseen compound degradations on MDVD-108K. Best and second-best results are highlighted in \textbf{bold} and \underline{underlined}.}
  \label{tab:unseen_compound_robustness}
  \resizebox{\linewidth}{!}{%
    \setlength{\tabcolsep}{0.5em}%
    \begin{tabular}{l|ccc|ccc}
      \toprule
      \multirow{2}{*}{\textbf{Method}}     & \multicolumn{3}{c|}{\textbf{Seen Compound Avg.}} & \multicolumn{3}{c}{\textbf{Unseen Compound Avg.}}                                                                                    \\
                                           & PSNR$\uparrow$                                   & SSIM$\uparrow$                                    & LPIPS$\downarrow$  & PSNR$\uparrow$    & SSIM$\uparrow$     & LPIPS$\downarrow$  \\
      \midrule
      MoCE-IR                              & 24.43                                            & 0.7729                                            & 0.1420             & 22.21             & \underline{0.7341} & 0.2397             \\
      BaryIR                               & \underline{25.02}                                & \underline{0.7840}                                & \underline{0.1262} & \underline{22.64} & 0.7312             & \underline{0.2011} \\
      \rowcolor{cyan!10} \textbf{CoRE-UIR} & \textbf{26.17}                                   & \textbf{0.7989}                                   & \textbf{0.1136}    & \textbf{23.37}    & \textbf{0.7742}    & \textbf{0.1502}    \\
      \bottomrule
    \end{tabular}%
  }
  \vspace{-1em}
\end{table}

\section{Conclusion} \label{sec:conclusion}

In this paper, we present CoRE-UIR, a prior-guided global-local framework for efficient all-in-one remote sensing image restoration. By coupling restoration-oriented degradation priors from DPE, prior-state global alignment from GFM, and common-residual specialization from CoRE, the model first establishes a reliable global restoration basis and then injects degradation-specific corrections through low-rank residual experts. This design reduces redundant expert replication, yields more interpretable specialization behavior, and remains efficient in deployment. We also construct MDVD-108K, a large-scale UAV multi-degradation restoration dataset covering both synthetic and real-world degraded images.

Extensive experiments on MDVD-108K and MDRS-Landsat validate the effectiveness of this design across single-degradation, compound-degradation, real-world qualitative, satellite-domain, downstream detection, and unseen compound-degradation evaluations. The results show that CoRE-UIR not only improves restoration quality, but also advances the Pareto frontier of universal restoration by achieving a better balance between accuracy, latency, and memory cost than existing baselines. More broadly, these findings suggest that universal restoration benefits from keeping shared restoration behaviors in a common dense expert and modeling degradation-specific differences as lightweight residual corrections.

Despite these advancements, CoRE-UIR still mainly addresses known degradation factors and evaluates downstream utility through object detection. Future work will focus on extending CoRE-UIR to broader unseen degradation categories, more complex real-world mixtures, and additional downstream remote sensing tasks.

\section{Acknowledgments}

This work was supported in part by the National Natural Science Foundation of China under Grants 42471414, 42471504 and 42230108.

\appendix
\section{Dataset Synthesis Protocol}
\label{appendix:dataset_synthesis}

This appendix instantiates the degradation operator $\mathcal{G}_{\mathcal{S}}$ and the selected configuration set $\mathcal{C}$ in Section~\ref{sec:problem_formulation} for the \textbf{MDVD-108K} benchmark. The goal is to generate controllable degradations while preserving a reasonable degree of physical realism for UAV imagery. For each base degradation $d \in \mathcal{D}$ and severity level $s$, we predefine a parameter space $\Omega_{d,s}$ and randomly sample
\begin{equation}
  \boldsymbol{\theta}_{d,s}\sim p(\boldsymbol{\theta}\mid d,s).
\end{equation}
Let $T_d(\cdot;\boldsymbol{\theta}_{d,s})$ denote the corresponding synthesis operator. For a singleton active set $\mathcal{S}=\{d\}$, the operator in Eq.~\ref{eq:degradation_observation} reduces to
\begin{equation}
  \mathcal{G}_{\{d\}}(\mathbf{I}) = T_d(\mathbf{I};\boldsymbol{\theta}_{d,s}).
\end{equation}
Starting from a clean image $\mathbf{I}\in[0,1]^{H\times W\times 3}$, we estimate a monocular depth map and convert it into a normalized distance-like representation $\mathbf{D}\in[0,1]^{H\times W}$, where distant regions are close to $1$ and near regions are close to $0$. All depth-aware degradation models are defined on this normalized map. Compound cases use the ordered cascade in Appendix~\ref{appendix:compound_degradation_protocol}. To avoid data leakage, train/val/test samples are synthesized independently from disjoint clean source images that follow the original VisDrone partition, and real degraded images are never reused as synthesis sources.

\subsection{Single-Degradation Models}
\label{appendix:single_degradation_models}

The current release instantiates six base operators, namely defocus blur, motion blur, fog, rain, dust, and low-light. Each operator corresponds to one singleton case $\mathcal{S}=\{d\}$ in Section~\ref{sec:problem_formulation}.

\paragraph{Defocus blur.}
Defocus blur is modeled in a scene-adaptive manner. Instead of directly amplifying normalized depth differences, we first estimate the global depth span of the scene and use it to modulate the effective separation strength between depth layers. The image is partitioned into several depth bins, and each bin is assigned a disk-blur radius
\begin{equation}
  r_b=r_g+r_{\max}\bigl(\lambda\,s_d\,|c_b-f|\bigr)^p,
\end{equation}
where $r_g$ is a global blur floor, $f$ is the sampled focal plane, and $\lambda$ is determined by the scene depth span. The final result is obtained by softly blending the responses of all depth bins. This design avoids unrealistically large foreground--background differences in scenes with limited depth variation.

\paragraph{Motion blur.}
Motion blur is modeled as a trajectory-induced point spread function (PSF) during exposure. Let $\boldsymbol{\Gamma}(t)$ denote the image-plane motion trajectory and $w(t)$ the exposure weighting function. The degraded image is generated by convolving the clean image with the corresponding motion PSF,
\begin{equation}
  \mathbf{Y}=h_{\mathrm{motion}}\ast\mathbf{I}.
\end{equation}
This formulation covers common motion patterns in UAV imagery, including camera shake and platform motion.

\paragraph{Fog.}
Fog is synthesized using a depth-aware atmospheric scattering model. We first map the normalized distance map to an effective depth with a positive lower bound and then compute a transmission map
\begin{equation}
  t(x)=\exp\bigl(-\beta(x)\,\tilde{\mathbf{D}}(x)^{\gamma}\bigr),
\end{equation}
where $\beta(x)$ is either constant or modulated by a low-frequency random field to produce non-uniform fog. The final foggy image is written as
\begin{equation}
  \mathbf{Y}(x)=t(x)\mathbf{I}(x)+\bigl(1-t(x)\bigr)\mathbf{A},
\end{equation}
with $\mathbf{A}$ denoting atmospheric light. This model reproduces depth-dependent contrast attenuation and spatially varying haze density.

\paragraph{Rain.}
Rain degradation is composed of three components: a streak layer, a low-frequency rain veil, and optional sparse lens droplets at severe levels. Rain streaks are synthesized by sampling line primitives with random length, orientation, and opacity, followed by anisotropic blur and contrast shaping. The streak layer is first blended with the clean image as
\begin{equation}
  \mathbf{I}_r=(1-\alpha_r R)\mathbf{I}+\alpha_r R\,\mathbf{C}_r+\eta_r R,
\end{equation}
where $R$ denotes the processed streak map, $\mathbf{C}_r$ is the rain color, and $\alpha_r$ and $\eta_r$ control opacity and brightness enhancement, respectively. A low-frequency veil field is then introduced to mimic the global curtain-like appearance of rainfall, yielding
\begin{equation}
  \mathbf{I}_v(x)=(1-m_v(x))\,t_v(x)\mathbf{I}_r(x)+m_v(x)\mathbf{A}_v,
\end{equation}
where $t_v(x)$ and $m_v(x)$ denote veil-related attenuation and mixing terms, and $\mathbf{A}_v$ is the corresponding airlight color. Heavy-rain cases may additionally include localized droplet distortions. This formulation allows the synthesized rain to exhibit both local streak structures and global visibility degradation.

\paragraph{Dust.}
Dust degradation combines warm-color atmospheric scattering, spatially non-uniform attenuation, and sparse particulate occlusion. The base layer follows a haze-like scattering process with yellow-brown airlight:
\begin{equation}
  \mathbf{I}_b(x)=t_d(x)\mathbf{I}(x)+\bigl(1-t_d(x)\bigr)\mathbf{A}_d,
\end{equation}
where $t_d(x)$ is the dust transmission map and $\mathbf{A}_d$ denotes warm atmospheric light. A sparse set of elliptical particles is then blended onto the base image by
\begin{equation}
  \mathbf{I}_p(x)=\bigl(1-M_p(x)\bigr)\mathbf{I}_b(x)+M_p(x)\mathbf{C}_p,
\end{equation}
where $M_p(x)$ is the aggregate particle mask and $\mathbf{C}_p$ is a slightly brighter particle color. Finally, saturation and contrast are moderately compressed to mimic the desaturated appearance of dusty environments. Compared with standard haze synthesis, this model better reflects the color cast and particulate interference frequently observed in UAV scenes.

\paragraph{Low-light.}
Low-light degradation primarily simulates under-exposure and sensor noise. We first attenuate exposure and apply a gamma transform,
\begin{equation}
  \mathbf{I}_0=(\alpha_e\mathbf{I})^{\gamma_\ell},
\end{equation}
and then inject both signal-dependent shot noise and signal-independent read noise. To approximate practical sensor artifacts, the noise field is lightly smoothed to form spatially correlated noise. Mild color shifts and compression-like distortions can also be introduced at stronger settings.

\subsection{Compound Degradation Protocol}
\label{appendix:compound_degradation_protocol}

The compound benchmark instantiates a selected subset of the $|\mathcal{S}|>1$ cases in Eq.~\ref{eq:degradation_observation} through cascaded processing rather than direct pixel-wise mixing. For an ordered active degradation tuple $\mathcal{S}=(d^{(1)},\ldots,d^{(K)})$, we define
\begin{equation}
  \mathcal{G}_{\mathcal{S}} = T_{d^{(K)}}\left(\cdot;\boldsymbol{\theta}_{d^{(K)},s_K}\right) \circ \cdots \circ T_{d^{(1)}}\left(\cdot;\boldsymbol{\theta}_{d^{(1)},s_1}\right),
\end{equation}
so that the degraded sample is generated by
\begin{align}
  \mathbf{Y} & = \mathcal{G}_{\mathcal{S}}(\mathbf{I})                                                                                                                                                     \\
             & = T_{d^{(K)}}\Bigl(\cdots T_{d^{(2)}}\bigl(T_{d^{(1)}}(\mathbf{I};\boldsymbol{\theta}_{d^{(1)},s_1});\boldsymbol{\theta}_{d^{(2)},s_2}\bigr)\cdots;\boldsymbol{\theta}_{d^{(K)},s_K}\Bigr).
\end{align}

In MDVD-108K we use $K=2$ and a fixed cross-category order: a weather operator $d^{(1)}\in\{\text{fog},\text{rain},\text{dust}\}$ is applied first, followed by an imaging operator $d^{(2)}\in\{\text{motion}, \text{low-light}\}$. This yields six compound settings: fog+motion, fog+low-light, dust+motion, dust+low-light, rain+motion, and rain+low-light. Sequential composition preserves the physical interpretation of each stage. For example, fog+motion first attenuates scene radiance through scattering and then applies directional blur, while rain+low-light couples visible rain structures with subsequent under-exposure and sensor noise. Compared with direct pixel-wise mixing, this cascade better matches the progressive accumulation of multiple degradations in practical imaging pipelines.

\subsection{Dataset Statistics and Split Policy}
\label{appendix:dataset_statistics}

\begin{table}[t]
  \centering
  \caption{Stratified sample counts of MDVD-108K.}
  \label{tab:mdvd108k_split}
  \small
  \setlength{\tabcolsep}{1em}
  \begin{tabular}{lccr}
    \toprule
    Split      & \makecell[c]{Single                                \\Degradation}  & \makecell[c]{Compound                                \\Degradation} & Total   \\
    \midrule
    Training   & $12{,}000 \times 6$ & $2{,}400 \times 6$ & 86,400  \\
    Validation & $1{,}500 \times 6$  & $300 \times 6$     & 10,800  \\
    Testing    & $1{,}500 \times 6$  & $300 \times 6$     & 10,800  \\
    Real       & --                  & --                 & 500     \\
    \midrule
    Total      & $15{,}000 \times 6$ & $3{,}000 \times 6$ & 108,500 \\
    \bottomrule
  \end{tabular}
\end{table}

\begin{figure}[t]
  \centering
  \includegraphics[width=\linewidth]{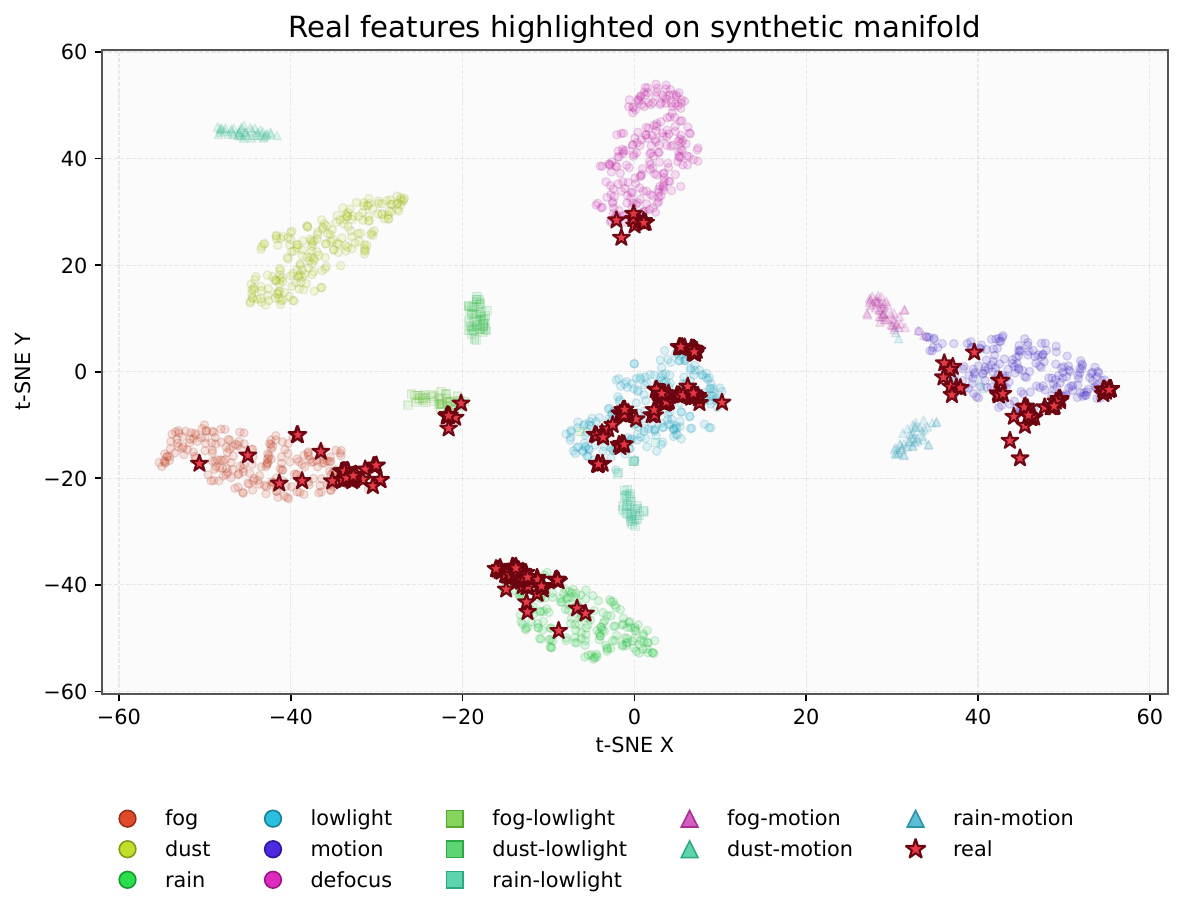}
  \caption{Synthetic-real comparison in the DPE feature space. Colored points are synthetic degradations and red stars are real UAV samples. The broader synthetic manifold covers the real samples.}
  \label{fig:real_vs_synthetic_dpe}
  \vspace{-1em}
\end{figure}

Table~\ref{tab:mdvd108k_split} shows that MDVD-108K contains $15{,}000$ samples for each single degradation type and $3{,}000$ samples for each compound setting, yielding $(15{,}000+3{,}000)\times6=108{,}000$ synthesized images in total. The split ratio is fixed to $8{:}1{:}1$ for training, validation, and testing. Each split is synthesized only from its own clean source images, so the same undegraded image never contributes samples across train/validation/test partitions. The real-world degraded subset is reserved for evaluation only and does not participate in degradation synthesis. Overall, MDVD-108K combines physically motivated operator design with strict split separation, making it controllable, reproducible, and suitable for systematic evaluation.

\subsection{Synthetic-Real Distribution Analysis}
\label{appendix:synthetic_real_distribution}

To examine the distribution gap between synthesized and real degradations, we compare synthetic degraded samples and real-world degraded UAV images in the DPE feature space. Fig.~\ref{fig:real_vs_synthetic_dpe} overlays real samples on the synthetic manifold. The synthetic samples span a broader set of degradation regions, while the real samples are distributed inside or near these synthetic clusters rather than forming an isolated out-of-distribution group. This distributional relationship indicates that MDVD-108K covers the main real-world degradation appearances observed in UAV imagery within the learned degradation-prior space.

\printcredits

\bibliographystyle{cas-model2-names}

\bibliography{main.bib}

@article{ai2024lora,
  title   = {LoRA-IR: Taming low-rank experts for efficient all-in-one image restoration},
  author  = {Ai, Yuang and Huang, Huaibo and He, Ran},
  journal = {arXiv preprint arXiv:2410.15385},
  year    = {2024}
}

@inproceedings{cai2023retinexformer,
  title     = {Retinexformer: One-stage retinex-based transformer for low-light image enhancement},
  author    = {Cai, Yuanhao and Bian, Hao and Lin, Jing and Wang, Haoqian and Timofte, Radu and Zhang, Yulun},
  booktitle = {Proceedings of the IEEE/CVF International Conference on Computer Vision},
  pages     = {12504--12513},
  year      = {2023}
}

@inproceedings{chen2021pre,
  title     = {Pre-trained image processing transformer},
  author    = {Chen, Hanting and Wang, Yunhe and Guo, Tianyu and Xu, Chang and Deng, Yiping and Liu, Zhenhua and Ma, Siwei and Xu, Chunjing and Xu, Chao and Gao, Wen},
  booktitle = {Proceedings of the IEEE/CVF conference on computer vision and pattern recognition},
  pages     = {12299--12310},
  year      = {2021}
}

@article{chen2026any2any,
  title   = {Any2any: Unified arbitrary modality translation for remote sensing},
  author  = {Chen, Haoyang and Zhang, Jing and Wang, Hebaixu and Wang, Shiqin and Huang, Pohsun and Li, Jiayuan and Guo, Haonan and Wang, Di and Wang, Zheng and Du, Bo},
  journal = {arXiv preprint arXiv:2603.04114},
  year    = {2026}
}

@inproceedings{chen2022nafnet,
  title        = {Simple baselines for image restoration},
  author       = {Chen, Liangyu and Chu, Xiaojie and Zhang, Xiangyu and Sun, Jian},
  booktitle    = {Proceedings of the European Conference on Computer Vision},
  pages        = {17--33},
  year         = {2022},
  organization = {Springer}
}

@inproceedings{chen2023activating,
  title     = {Activating more pixels in image super-resolution transformer},
  author    = {Chen, Xiangyu and Wang, Xintao and Zhou, Jiantao and Qiao, Yu and Dong, Chao},
  booktitle = {Proceedings of the IEEE/CVF conference on computer vision and pattern recognition},
  pages     = {22367--22377},
  year      = {2023}
}

@article{chen2025hyperspectral,
  title     = {Hyperspectral Video Tracking with Spectral-Spatial Fusion and Memory Enhancement},
  author    = {Chen, Yuzeng and Yuan, Qiangqiang and Xie, Hong and Tang, Yuqi and Xiao, Yi and He, Jiang and Guan, Renxiang and Liu, Xinwang and Zhang, Liangpei},
  journal   = {IEEE Transactions on Image Processing},
  year      = {2025},
  publisher = {IEEE}
}

@article{chen2022cross,
  title   = {Cross aggregation transformer for image restoration},
  author  = {Chen, Zheng and Zhang, Yulun and Gu, Jinjin and Kong, Linghe and Yuan, Xin and others},
  journal = {Advances in Neural Information Processing Systems},
  volume  = {35},
  pages   = {25478--25490},
  year    = {2022}
}

@article{Cheng2014Patch,
  author  = {Cheng, Qing and Shen, Huanfeng and Zhang, Liangpei and Yuan, Qiangqiang and Zeng, Chao},
  year    = {2014},
  month   = {06},
  pages   = {54-68},
  title   = {Cloud removal for remotely sensed images by similar pixel replacement guided with a spatio-temporal MRF model},
  volume  = {92},
  journal = {ISPRS J. Photogramm. Remote Sens.}
}

@article{cui2026unified,
  title   = {A Unified Foundation Model for All-in-One Multi-Modal Remote Sensing Image Restoration and Fusion with Language Prompting},
  author  = {Cui, Yongchuan and Liu, Peng},
  journal = {arXiv preprint arXiv:2604.05629},
  year    = {2026}
}

@article{cui2025pansharpening,
  title     = {Pansharpening via predictive filtering with element-wise feature mixing},
  author    = {Cui, Yongchuan and Liu, Peng and Ma, Yan and Chen, Lajiao and Xu, Mengzhen and Guo, Xingyan},
  journal   = {ISPRS Journal of Photogrammetry and Remote Sensing},
  volume    = {219},
  pages     = {22--37},
  year      = {2025},
  publisher = {Elsevier}
}

@inproceedings{cui2025enpowering,
  title     = {Enpowering Your Pansharpening Models with Generalizability: Unified Distribution is All You Need},
  author    = {Cui, Yongchuan and Liu, Peng and Zhang, Hui},
  booktitle = {Proceedings of the IEEE/CVF International Conference on Computer Vision},
  pages     = {11850--11860},
  year      = {2025}
}

@article{dong2015image,
  title     = {Image super-resolution using deep convolutional networks},
  author    = {Dong, Chao and Loy, Chen Change and He, Kaiming and Tang, Xiaoou},
  journal   = {IEEE Transactions on Pattern Analysis and Machine Intelligence},
  volume    = {38},
  number    = {2},
  pages     = {295--307},
  year      = {2015},
  publisher = {IEEE}
}

@article{dong2026phydae,
  title     = {PhyDAE: Physics-Guided Degradation-Adaptive Experts for All-in-One Remote Sensing Image Restoration},
  author    = {Dong, Zhe and Zhang, Zhengning and Sun, Yuzhe and Jiang, Haochen and Liu, Tianzhu and Gu, Yanfeng},
  journal   = {IEEE Transactions on Geoscience and Remote Sensing},
  year      = {2026},
  publisher = {IEEE}
}

@article{gao2023frequency,
  title     = {Frequency-oriented efficient transformer for all-in-one weather-degraded image restoration},
  author    = {Gao, Tao and Wen, Yuanbo and Zhang, Kaihao and Zhang, Jing and Chen, Ting and Liu, Lidong and Luo, Wenhan},
  journal   = {IEEE Transactions on Circuits and Systems for Video Technology},
  volume    = {34},
  number    = {3},
  pages     = {1886--1899},
  year      = {2023},
  publisher = {IEEE}
}

@inproceedings{guo2024mambair,
  title        = {Mambair: A simple baseline for image restoration with state-space model},
  author       = {Guo, Hang and Li, Jinmin and Dai, Tao and Ouyang, Zhihao and Ren, Xudong and Xia, Shu-Tao},
  booktitle    = {European conference on computer vision},
  pages        = {222--241},
  year         = {2024},
  organization = {Springer}
}

@inproceedings{ResNet,
  title     = {Deep residual learning for image recognition},
  author    = {He, Kaiming and Zhang, Xiangyu and Ren, Shaoqing and Sun, Jian},
  booktitle = {Proceedings of the IEEE Conference on Computer Vision and Pattern Recognition},
  pages     = {770--778},
  year      = {2016}
}

@article{he2015total,
  title     = {Total-variation-regularized low-rank matrix factorization for hyperspectral image restoration},
  author    = {He, Wei and Zhang, Hongyan and Zhang, Liangpei and Shen, Huanfeng},
  journal   = {IEEE transactions on geoscience and remote sensing},
  volume    = {54},
  number    = {1},
  pages     = {178--188},
  year      = {2015},
  publisher = {IEEE}
}

@inproceedings{hu2018squeeze,
  title     = {Squeeze-and-excitation networks},
  author    = {Hu, Jie and Shen, Li and Sun, Gang},
  booktitle = {Proceedings of the IEEE conference on computer vision and pattern recognition},
  pages     = {7132--7141},
  year      = {2018}
}

@article{hu2025clusir,
  title   = {ClusIR: Towards Cluster-Guided All-in-One Image Restoration},
  author  = {Hu, Shengkai and Ma, Jiaqi and Wan, Jun and Min, Wenwen and Jing, Yongcheng and Zhang, Lefei and Tao, Dacheng},
  journal = {arXiv preprint arXiv:2512.10948},
  year    = {2025}
}

@article{jiang2025survey,
  title   = {A survey on all-in-one image restoration: Taxonomy, evaluation and future trends},
  author  = {Jiang, Junjun and Zuo, Zengyuan and Wu, Gang and Jiang, Kui and Liu, Xianming},
  journal = {IEEE Transactions on Pattern Analysis and Machine Intelligence},
  volume  = {47},
  number  = {12},
  pages   = {11892--11911},
  year    = {2025}
}

@article{jin2025mb,
  title     = {MB-TaylorFormer V2: Improved multi-branch linear transformer expanded by Taylor formula for image restoration},
  author    = {Jin, Zhi and Qiu, Yuwei and Zhang, Kaihao and Li, Hongdong and Luo, Wenhan},
  journal   = {IEEE Transactions on Pattern Analysis and Machine Intelligence},
  year      = {2025},
  publisher = {IEEE}
}

@article{jing2023denoising,
  title     = {Denoising diffusion probabilistic feature-based network for cloud removal in Sentinel-2 imagery},
  author    = {Jing, Ran and Duan, Fuzhou and Lu, Fengxian and Zhang, Miao and Zhao, Wenji},
  journal   = {Remote Sensing},
  volume    = {15},
  number    = {9},
  pages     = {2217},
  year      = {2023},
  publisher = {MDPI}
}

@article{khanam2024yolov11,
  title   = {Yolov11: An overview of the key architectural enhancements},
  author  = {Khanam, Rahima and Hussain, Muhammad},
  journal = {arXiv preprint arXiv:2410.17725},
  year    = {2024}
}

@article{kong2024towards,
  title   = {Towards effective multiple-in-one image restoration: A sequential and prompt learning strategy},
  author  = {Kong, Xiangtao and Dong, Chao and Zhang, Lei},
  journal = {arXiv preprint arXiv:2401.03379},
  year    = {2024}
}

@inproceedings{li2022airnet,
  title     = {All-in-one image restoration for unknown corruption},
  author    = {Li, Boyun and Liu, Xiao and Hu, Peng and Wu, Zhongqin and Lv, Jiancheng and Peng, Xi},
  booktitle = {Proceedings of the IEEE/CVF Conference on Computer Vision and Pattern Recognition},
  pages     = {17452--17462},
  year      = {2022}
}

@article{li2019benchmarking,
  title     = {Benchmarking single-image dehazing and beyond},
  author    = {Li, Boyi and Ren, Wenqi and Fu, Dengpan and Tao, Dacheng and Feng, Dan and Zeng, Wenjun and Wang, Zhangyang},
  journal   = {IEEE Transactions on Image Processing},
  volume    = {28},
  number    = {1},
  pages     = {492--505},
  year      = {2019},
  publisher = {IEEE}
}

@article{li2025cloudruler,
  title     = {CloudRuler: Rule-based transformer for cloud removal in Landsat images},
  author    = {Li, Jun and Wang, Yihui and Sheng, Qinghong and Wu, Zhaocong and Wang, Bo and Ling, Xiao and Liu, Xiang and Du, Yang and Gao, Fan and Camps-Valls, Gustau and others},
  journal   = {Remote Sensing of Environment},
  volume    = {328},
  pages     = {114913},
  year      = {2025},
  publisher = {Elsevier}
}

@article{li2020thin,
  title     = {Thin cloud removal in optical remote sensing images based on generative adversarial networks and physical model of cloud distortion},
  author    = {Li, Jun and Wu, Zhaocong and Hu, Zhongwen and Zhang, Jiaqi and Li, Mingliang and Mo, Lu and Molinier, Matthieu},
  journal   = {ISPRS Journal of Photogrammetry and Remote Sensing},
  volume    = {166},
  pages     = {373--389},
  year      = {2020},
  publisher = {Elsevier}
}

@inproceedings{liang2021swinir,
  title     = {SwinIR: Image Restoration Using Swin Transformer},
  author    = {Liang, Jingyun and Cao, Jiezhang and Sun, Guolei and Zhang, Kai and Van Gool, Luc and Timofte, Radu},
  booktitle = {Proceedings of the IEEE/CVF International Conference on Computer Vision Workshops},
  pages     = {1833--1844},
  year      = {2021}
}

@article{lihe2025ada4dir,
  title     = {Ada4DIR: An adaptive model-driven all-in-one image restoration network for remote sensing images},
  author    = {Lihe, Ziyang and Yuan, Qiangqiang and He, Jiang and Jin, Xianyu and Xiao, Yi and Chen, Yuzeng and Shen, Huanfeng and Zhang, Liangpei},
  journal   = {Information Fusion},
  volume    = {118},
  pages     = {102930},
  year      = {2025},
  publisher = {Elsevier}
}

@article{remoteclip,
  author  = {Fan Liu and
             Delong Chen and
             Zhangqingyun Guan and
             Xiaocong Zhou and
             Jiale Zhu and
             Qiaolin Ye and
             Liyong Fu and
             Jun Zhou},
  title   = {RemoteCLIP: A Vision Language Foundation Model for Remote Sensing},
  journal = {IEEE Transactions on Geoscience and Remote Sensing},
  volume  = {62},
  pages   = {1--16},
  year    = {2024}
}

@article{liu2024crossmatch,
  author  = {Liu, Ruizhong and Luo, Tingzhang and Huang, Shaoguang and Wu, Yuwei and Jiang, Zhen and Zhang, Hongyan},
  journal = {IEEE Transactions on Geoscience and Remote Sensing},
  title   = {CrossMatch: Cross-View Matching for Semi-Supervised Remote Sensing Image Segmentation},
  year    = {2024},
  volume  = {62},
  pages   = {1-15},
  doi     = {10.1109/TGRS.2024.3507050}
}

@article{liu2026ihdcp,
  title     = {IHDCP: Single image dehazing using inverted haze density correction prior},
  author    = {Liu, Yun and Li, Tao and Tan, Chunping and Ren, Wenqi and Ancuti, Cosmin and Lin, Weisi},
  journal   = {IEEE Transactions on Image Processing},
  year      = {2026},
  publisher = {IEEE}
}

@inproceedings{loshchilov2017decoupled,
  title     = {Decoupled Weight Decay Regularization},
  author    = {Loshchilov, Ilya and Hutter, Frank},
  booktitle = {International Conference on Learning Representations},
  year      = {2017}
}

@inproceedings{loshchilov2016sgdr,
  title     = {SGDR: Stochastic gradient descent with warm restarts},
  author    = {Loshchilov, Ilya and Hutter, Frank},
  booktitle = {International Conference on Learning Representations},
  year      = {2017}
}

@inproceedings{luo2023daclip,
  title     = {Controlling vision-language models for universal image restoration},
  author    = {Luo, Ziwei and Gustafsson, Fredrik K and Zhao, Zheng and Sj{\"o}lund, Jens and Sch{\"o}n, Thomas B},
  booktitle = {International Conference on Learning Representations},
  pages     = {1--13},
  year      = {2024}
}

@article{ma2023prores,
  title   = {Prores: Exploring degradation-aware visual prompt for universal image restoration},
  author  = {Ma, Jiaqi and Cheng, Tianheng and Wang, Guoli and Zhang, Qian and Wang, Xinggang and Zhang, Lefei},
  journal = {arXiv preprint arXiv:2306.13653},
  year    = {2023}
}

@article{ma2025evoir,
  title   = {EvoIR: Towards All-in-One Image Restoration via Evolutionary Frequency Modulation},
  author  = {Ma, Jiaqi and Hu, Shengkai and Zhang, Xu and Wan, Jun and Huang, Jiaxing and Zhang, Lefei and Khan, Salman},
  journal = {arXiv preprint arXiv:2512.05104},
  year    = {2025}
}

@inproceedings{mou2022dgunet,
  title     = {Deep generalized unfolding networks for image restoration},
  author    = {Mou, Chong and Wang, Qian and Zhang, Jian},
  booktitle = {Proceedings of the IEEE/CVF Conference on Computer Vision and Pattern Recognition},
  pages     = {17399--17410},
  year      = {2022}
}

@article{ozdenizci2023weatherdiff,
  title     = {Restoring vision in adverse weather conditions with patch-based denoising diffusion models},
  author    = {{\"O}zdenizci, Ozan and Bhatt, Mukund and Legenstein, Robert},
  journal   = {IEEE Transactions on Pattern Analysis and Machine Intelligence},
  volume    = {45},
  number    = {8},
  pages     = {10346--10357},
  year      = {2023},
  publisher = {IEEE}
}

@article{LuojiaSET,
  title   = {HDRSA-Net: Hybrid dynamic residual self-attention network for SAR-assisted optical image cloud and shadow removal},
  journal = {ISPRS Journal of Photogrammetry and Remote Sensing},
  volume  = {218},
  pages   = {258-275},
  year    = {2024},
  issn    = {0924-2716},
  author  = {Jun Pan and Jiangong Xu and Xiaoyu Yu and Guo Ye and Mi Wang and Yumin Chen and Jianshen Ma}
}

@inproceedings{potlapalli2024promptir,
  title     = {PromptIR: Prompting for all-in-one image restoration},
  author    = {Potlapalli, Vaishnav and Zamir, Syed Waqas and Khan, Salman and Khan, Fahad Shahbaz},
  booktitle = {Advances in Neural Information Processing Systems},
  year      = {2024},
  volume    = {36},
  pages     = {71275--71293}
}

@inproceedings{radford2021clip,
  title     = {Learning transferable visual models from natural language supervision},
  author    = {Radford, Alec and Kim, Jong Wook and Hallacy, Chris and Ramesh, Aditya and Goh, Gabriel and Agarwal, Sandhini and Sastry, Girish and Askell, Amanda and Mishkin, Pamela and Clark, Jack and others},
  booktitle = {Proceedings of the International Conference on Machine Learning},
  pages     = {8748--8763},
  year      = {2021}
}

@article{riquelme2021scaling,
  title   = {Scaling vision with sparse mixture of experts},
  author  = {Riquelme, Carlos and Puigcerver, Joan and Mustafa, Basil and Neumann, Maxim and Jenatton, Rodolphe and Susano Pinto, Andr{\'e} and Keysers, Daniel and Houlsby, Neil},
  journal = {Advances in Neural Information Processing Systems},
  volume  = {34},
  pages   = {8583--8595},
  year    = {2021}
}

@article{shazeer2017moe,
  title   = {Outrageously large neural networks: The sparsely-gated mixture-of-experts layer},
  author  = {Shazeer, Noam and Mirhoseini, Azalia and Maziarz, Krzysztof and Davis, Andy and Le, Quoc and Hinton, Geoffrey and Dean, Jeff},
  journal = {International Conference on Learning Representations},
  year    = {2017}
}

@article{shen2015missing,
  title     = {Missing information reconstruction of remote sensing data: A technical review},
  author    = {Shen, Huanfeng and Li, Xinghua and Cheng, Qing and Zeng, Chao and Yang, Gang and Li, Huifang and Zhang, Liangpei},
  journal   = {IEEE Geoscience and Remote Sensing Magazine},
  volume    = {3},
  number    = {3},
  pages     = {61--85},
  year      = {2015},
  publisher = {IEEE}
}

@inproceedings{shi2016real,
  title     = {Real-time single image and video super-resolution using an efficient sub-pixel convolutional neural network},
  author    = {Shi, Wenzhe and Caballero, Jose and Husz{\'a}r, Ferenc and Totz, Johannes and Aitken, Andrew P and Bishop, Rob and Rueckert, Daniel and Wang, Zehan},
  booktitle = {Proceedings of the IEEE Conference on Computer Vision and Pattern Recognition},
  pages     = {1874--1883},
  year      = {2016}
}

@article{shu2025restore,
  title     = {RESTORE-DiT: Reliable satellite image time series reconstruction by multimodal sequential diffusion transformer},
  author    = {Shu, Qidi and Zhu, Xiaolin and Xu, Shuai and Wang, Yan and Liu, Denghong},
  journal   = {Remote Sensing of Environment},
  volume    = {328},
  pages     = {114872},
  year      = {2025},
  publisher = {Elsevier}
}

@article{song2023vision,
  title   = {Vision Transformers for Single Image Dehazing},
  author  = {Song, Yuda and He, Zhuqing and Qian, Hui and Du, Xin},
  journal = {IEEE Transactions on Image Processing},
  year    = {2023},
  volume  = {32},
  pages   = {1927--1941}
}

@article{stucker2023u,
  title     = {U-TILISE: A Sequence-to-Sequence Model for Cloud Removal in Optical Satellite Time Series},
  author    = {Stucker, Corinne and Garnot, Vivien Sainte Fare and Schindler, Konrad},
  journal   = {IEEE Trans. Geosci. Remote Sens.},
  volume    = {61},
  pages     = {1--16},
  year      = {2023},
  publisher = {IEEE}
}

@article{sui2024diffusion,
  title     = {Diffusion Enhancement for Cloud Removal in Ultra-Resolution Remote Sensing Imagery},
  author    = {Sui, Jialu and Ma, Yiyang and Yang, Wenhan and Zhang, Xiaokang and Pun, Man-On and Liu, Jiaying},
  journal   = {IEEE Transactions on Geoscience and Remote Sensing},
  volume    = {62},
  pages     = {1--14},
  year      = {2024},
  publisher = {IEEE}
}

@article{tang2026learning,
  title     = {Learning Continuous Wasserstein Barycenter Space for Generalized All-in-One Image Restoration},
  author    = {Tang, Xiaole and He, Xiaoyi and Xu, Jiayi and Gu, Xiang and Sun, Jian},
  journal   = {IEEE Transactions on Pattern Analysis and Machine Intelligence},
  year      = {2026},
  publisher = {IEEE}
}

@inproceedings{valanarasu2022transweather,
  title     = {Transweather: Transformer-based restoration of images degraded by adverse weather conditions},
  author    = {Valanarasu, Jeya Maria Jose and Yasarla, Rajeev and Patel, Vishal M},
  booktitle = {Proceedings of the IEEE/CVF Conference on Computer Vision and Pattern Recognition},
  pages     = {2353--2363},
  year      = {2022}
}

@article{wang2026region,
  title     = {Region-based Progressive Alignment Network for Multimodal Remote Sensing Object Detection},
  author    = {Wang, Anrui and Xu, Yang and Wang, Ruiqing and Fang, Yu and Wei, Zhihui and Wu, Zebin},
  journal   = {IEEE Transactions on Geoscience and Remote Sensing},
  volume    = {64},
  year      = {2026},
  publisher = {IEEE}
}

@inproceedings{wang2026residual,
  title     = {Residual diffusion bridge model for image restoration},
  author    = {Wang, Hebaixu and Zhang, Jing and Chen, Haoyang and Guo, Haonan and Wang, Di and Ma, Jiayi and Du, Bo},
  booktitle = {Proceedings of the IEEE/CVF Conference on Computer Vision and Pattern Recognition},
  pages     = {8375--8386},
  year      = {2026}
}

@article{wang2025disasterm3,
  title   = {Disasterm3: A remote sensing vision-language dataset for disaster damage assessment and response},
  author  = {Wang, Junjue and Xuan, Weihao and Qi, Heli and Liu, Zhihao and Liu, Kunyi and Wu, Yuhan and Chen, Hongruixuan and Song, Jian and Xia, Junshi and Zheng, Zhuo and others},
  journal = {arXiv preprint arXiv:2505.21089},
  year    = {2025}
}

@article{MTGAN,
  title     = {MT\_GAN: A SAR-to-optical image translation method for cloud removal},
  author    = {Wang, Peng and Chen, Yongkang and Huang, Bo and Zhu, Daiyin and Lu, Tongwei and Dalla Mura, Mauro and Chanussot, Jocelyn},
  journal   = {ISPRS Journal of Photogrammetry and Remote Sensing},
  volume    = {225},
  pages     = {180--195},
  year      = {2025},
  publisher = {Elsevier}
}

@article{wang2025m2restore,
  title     = {M2Restore: Mixture-of-experts-based mamba-cnn fusion framework for all-in-one image restoration},
  author    = {Wang, Yongzhen and Li, Yongjun and Zheng, Zhuoran and Zhang, Xiao-Ping and Wei, Mingqiang},
  journal   = {IEEE Transactions on Image Processing},
  volume    = {34},
  pages     = {8086--8100},
  year      = {2025},
  publisher = {IEEE}
}

@article{SSIM,
  title     = {Image quality assessment: from error visibility to structural similarity},
  author    = {Wang, Zhou and Bovik, Alan C and Sheikh, Hamid R and Simoncelli, Eero P},
  journal   = {IEEE Transactions on Image Processing},
  volume    = {13},
  number    = {4},
  pages     = {600--612},
  year      = {2004},
  publisher = {IEEE}
}

@inproceedings{wei2025robust,
  title     = {Robust Single Image Sand Removal by Leveraging Uncertainty-aware SAM Priors and Prompt Learning with Refined Perceptual Loss},
  author    = {Wei, Bingcai and Liu, Hui and Qian, Chuang and Li, Zijian and Wu, Wangyu and Meng, Zijie},
  booktitle = {Proceedings of the 33rd ACM International Conference on Multimedia},
  pages     = {4932--4941},
  year      = {2025}
}

@article{wen2025cross,
  title     = {Cross-level interaction and intra-level fusion network for remote sensing image dehazing},
  author    = {Wen, Yuanbo and Gao, Tao and Chen, Ting and Li, Ziqi and Liu, Mengkun and Liu, Lidong},
  journal   = {IEEE Transactions on Geoscience and Remote Sensing},
  year      = {2025},
  publisher = {IEEE}
}

@article{wen2025all,
  title     = {All-in-one weather-degraded image restoration via adaptive degradation-aware self-prompting model},
  author    = {Wen, Yuanbo and Gao, Tao and Li, Ziqi and Zhang, Jing and Zhang, Kaihao and Chen, Ting},
  journal   = {IEEE Transactions on Multimedia},
  year      = {2025},
  publisher = {IEEE}
}

@article{wen2026structure,
  title     = {Structure-preserving frequency-regularized text-guided optimal transport for unpaired rain streaks and raindrops removal},
  author    = {Wen, Yuanbo and Gao, Tao and Li, Ziqi and Zhang, Qianxi and Zhang, Jing and Chen, Ting and Liu, Lidong},
  journal   = {IEEE Transactions on Multimedia},
  year      = {2026},
  publisher = {IEEE}
}

@article{wen2023encoder,
  title     = {Encoder-free multiaxis physics-aware fusion network for remote sensing image dehazing},
  author    = {Wen, Yuanbo and Gao, Tao and Zhang, Jing and Li, Ziqi and Chen, Ting},
  journal   = {IEEE Transactions on Geoscience and Remote Sensing},
  volume    = {61},
  pages     = {1--15},
  year      = {2023},
  publisher = {IEEE}
}

@inproceedings{wu2025mp,
  title     = {MP-HSIR: A multi-prompt framework for universal hyperspectral image restoration},
  author    = {Wu, Zhehui and Chen, Yong and Yokoya, Naoto and He, Wei},
  booktitle = {Proceedings of the IEEE/CVF International Conference on Computer Vision},
  pages     = {13009--13020},
  year      = {2025}
}

@article{yang2024depth,
  title   = {Depth anything v2},
  author  = {Yang, Lihe and Kang, Bingyi and Huang, Zilong and Zhao, Zhen and Xu, Xiaogang and Feng, Jiashi and Zhao, Hengshuang},
  journal = {Advances in Neural Information Processing Systems},
  volume  = {37},
  pages   = {21875--21911},
  year    = {2024}
}

@article{yu2024multi,
  title     = {Multi-expert adaptive selection: Task-balancing for all-in-one image restoration},
  author    = {Yu, Xiaoyan and Zhou, Shen and Li, Huafeng and Zhu, Liehuang},
  journal   = {IEEE Transactions on Circuits and Systems for Video Technology},
  volume    = {35},
  number    = {5},
  pages     = {4619--4634},
  year      = {2024},
  publisher = {IEEE}
}

@article{yuan2020deep,
  title     = {Deep learning in environmental remote sensing: Achievements and challenges},
  author    = {Yuan, Qiangqiang and Shen, Huanfeng and Li, Tongwen and Li, Zhiwei and Li, Shuwen and Jiang, Yun and Xu, Hongzhang and Tan, Weiwei and Yang, Qianqian and Wang, Jiwen and others},
  journal   = {Remote sensing of Environment},
  volume    = {241},
  pages     = {111716},
  year      = {2020},
  publisher = {Elsevier}
}

@inproceedings{zamfir2025complexity,
  title     = {Complexity experts are task-discriminative learners for any image restoration},
  author    = {Zamfir, Eduard and Wu, Zongwei and Mehta, Nancy and Tan, Yuedong and Paudel, Danda Pani and Zhang, Yulun and Timofte, Radu},
  booktitle = {Proceedings of the Computer Vision and Pattern Recognition Conference},
  pages     = {12753--12763},
  year      = {2025}
}

@inproceedings{zamir2022restormer,
  title     = {Restormer: Efficient transformer for high-resolution image restoration},
  author    = {Zamir, Syed Waqas and Arora, Aditya and Khan, Salman and Hayat, Munawar and Khan, Fahad Shahbaz and Yang, Ming-Hsuan},
  booktitle = {Proceedings of the IEEE/CVF Conference on Computer Vision and Pattern Recognition},
  pages     = {5728--5739},
  year      = {2022}
}

@inproceedings{zamir2021mprnet,
  title     = {Multi-stage progressive image restoration},
  author    = {Zamir, Syed Waqas and Arora, Aditya and Khan, Salman and Hayat, Munawar and Khan, Fahad Shahbaz and Yang, Ming-Hsuan and Shao, Ling},
  booktitle = {Proceedings of the IEEE/CVF conference on computer vision and pattern recognition},
  pages     = {14821--14831},
  year      = {2021}
}

@article{zhang2024uir,
  title   = {UIR-LoRA: achieving universal image restoration through multiple low-rank adaptation},
  author  = {Zhang, Cheng and Gong, Dong and He, Jiumei and Zhu, Yu and Sun, Jinqiu and Zhang, Yanning},
  journal = {arXiv preprint arXiv:2409.20197},
  year    = {2024}
}

@inproceedings{zhang2023idr,
  title     = {Ingredient-oriented multi-degradation learning for image restoration},
  author    = {Zhang, Jinghao and Huang, Jie and Yao, Mingde and Yang, Zizheng and Yu, Hu and Zhou, Man and Zhao, Feng},
  booktitle = {Proceedings of the IEEE/CVF Conference on Computer Vision and Pattern Recognition},
  pages     = {5825--5835},
  year      = {2023}
}

@article{zhang2022enhanced,
  title     = {Enhanced spatio-temporal interaction learning for video deraining: Faster and better},
  author    = {Zhang, Kaihao and Li, Dongxu and Luo, Wenhan and Ren, Wenqi and Liu, Wei},
  journal   = {IEEE Transactions on Pattern Analysis and Machine Intelligence},
  volume    = {45},
  number    = {1},
  pages     = {1287--1293},
  year      = {2022},
  publisher = {IEEE}
}

@article{zhang2021deep,
  title     = {Deep dense multi-scale network for snow removal using semantic and depth priors},
  author    = {Zhang, Kaihao and Li, Rongqing and Yu, Yanjiang and Luo, Wenhan and Li, Changsheng},
  journal   = {IEEE Transactions on Image Processing},
  volume    = {30},
  pages     = {7419--7431},
  year      = {2021},
  publisher = {IEEE}
}

@article{zhang2023mc,
  title     = {MC-Blur: A comprehensive benchmark for image deblurring},
  author    = {Zhang, Kaihao and Wang, Tao and Luo, Wenhan and Ren, Wenqi and Stenger, Bj{\"o}rn and Liu, Wei and Li, Hongdong and Yang, Ming-Hsuan},
  journal   = {IEEE Transactions on Circuits and Systems for Video Technology},
  volume    = {34},
  number    = {5},
  pages     = {3755--3767},
  year      = {2023},
  publisher = {IEEE}
}

@inproceedings{zhang2018unreasonable,
  title     = {The unreasonable effectiveness of deep features as a perceptual metric},
  author    = {Zhang, Richard and Isola, Phillip and Efros, Alexei A and Shechtman, Eli and Wang, Oliver},
  booktitle = {Proceedings of the IEEE conference on computer vision and pattern recognition},
  pages     = {586--595},
  year      = {2018}
}

@inproceedings{zhang2024efficient,
  title     = {Efficient Deweather Mixture-of-Experts with Uncertainty-aware Feature-wise Linear Modulation},
  author    = {Zhang, Rongyu and Luo, Yulin and Liu, Jiaming and Yang, Huanrui and Dong, Zhen and Gudovskiy, Denis and Okuno, Tomoyuki and Nakata, Yohei and Keutzer, Kurt and Du, Yuan and others},
  booktitle = {Proceedings of the AAAI Conference on Artificial Intelligence},
  volume    = {38},
  number    = {15},
  pages     = {16812--16820},
  year      = {2024}
}

@article{Perceive-IR,
  author  = {Zhang, Xu and Ma, Jiaqi and Wang, Guoli and Zhang, Qian and Zhang, Huan and Zhang, Lefei},
  journal = {IEEE Transactions on Image Processing},
  title   = {Perceive-IR: Learning to Perceive Degradation Better for All-in-One Image Restoration},
  year    = {2026},
  volume  = {35},
  number  = {},
  pages   = {2018-2033}
}

@inproceedings{ClearAIR,
  title     = {ClearAIR: A Human-Visual-Perception-Inspired All-in-One Image Restoration},
  author    = {Zhang, Xu and Zhang, Huan and Wang, Guoli and Zhang, Qian and Zhang, Lefei},
  booktitle = {Proceedings of the AAAI Conference on Artificial Intelligence},
  year      = {2026},
  pages     = {12861-12869}
}

@article{UniUIR,
  author  = {Zhang, Xu and Zhang, Huan and Wang, Guoli and Zhang, Qian and Zhang, Lefei and Du, Bo},
  journal = {IEEE Transactions on Image Processing},
  title   = {UniUIR: Considering Underwater Image Restoration as an All-in-One Learner},
  year    = {2025},
  volume  = {34},
  number  = {},
  pages   = {6963-6977}
}

@inproceedings{zhang2018image,
  title     = {Image super-resolution using very deep residual channel attention networks},
  author    = {Zhang, Yulun and Li, Kunpeng and Li, Kai and Wang, Lichen and Zhong, Bineng and Fu, Yun},
  booktitle = {Proceedings of the European conference on computer vision (ECCV)},
  pages     = {286--301},
  year      = {2018}
}

@article{ZHANG2026650,
  title    = {{STAR-IOD}: Scale-decoupled Topology Alignment with Pseudo-Label Refinement for Remote Sensing Incremental Object Detection},
  author   = {Zhang, Yaoteng and Zhou, Qing and Gao, Junyu and Wang, Qi},
  year     = {2026},
  journal  = {ISPRS Journal of Photogrammetry and Remote Sensing},
  volume   = {238},
  pages    = {650--663},
  issn     = {0924-2716},
  doi      = {10.1016/j.isprsjprs.2026.05.036}
}

@article{zhang2026ecrformer,
  title     = {{ECRformer: An Efficient Cloud Removal Transformer with Semantic-Decoupled Learning for Multimodal Satellite Imagery}},
  author    = {Zhang, Zaiyan and Li, Jie and Liang, Yuanqi and Yan, Jining and Xiao, Yi and Su, Xin and Yuan, Qiangqiang},
  journal   = {ISPRS J. Photogramm. Remote Sens.},
  volume    = {237},
  pages     = {323-338},
  year      = {2026},
  publisher = {Elsevier}
}

@article{zhang2026task,
  title   = {{Task-Driven Prompt Learning: A Joint Framework for Multi-modal Cloud Removal and Segmentation}},
  author  = {Zhang, Zaiyan and Li, Jie and Shi, Shaowei and Yuan, Qiangqiang},
  journal = {arXiv preprint arXiv:2601.12052},
  year    = {2026}
}

@article{zhang2025multi,
  author   = {Zhang, Zaiyan and Yan, Jining and Liang, Yuanqi and Feng, Jiaxin and He, Haixu and Cao, Li},
  journal  = {IEEE Transactions on Geoscience and Remote Sensing},
  title    = {Multiscale Restoration of Missing Data in Optical Time-series Images With Masked Spatial-Temporal Attention Network},
  year     = {2025},
  volume   = {63},
  pages    = {1-15}
}

@article{zhu2021detection,
  title     = {Detection and tracking meet drones challenge},
  author    = {Zhu, Pengfei and Wen, Longyin and Du, Dawei and Bian, Xiao and Fan, Heng and Hu, Qinghua and Ling, Haibin},
  journal   = {IEEE Transactions on Pattern Analysis and Machine Intelligence},
  volume    = {44},
  number    = {11},
  pages     = {7380--7399},
  year      = {2021},
  publisher = {IEEE}
}

\end{document}